\UseRawInputEncoding
\documentclass{article}
\usepackage[final]{neurips_2022}
\usepackage{amsmath,amsthm,amssymb,amsfonts,bbm,bm}
\usepackage{enumitem}
\usepackage{etoolbox}
\usepackage{graphicx}
\usepackage{hyperref}
\usepackage{subfigure}
\usepackage{xcolor}
\newcommand\EnumPrefix{}

\newlist{senenum}{enumerate}{10}
\setlist[senenum]{label=\arabic*.,ref=\EnumPrefix,leftmargin=*}

\AtBeginEnvironment{lemmalist}{\renewcommand\EnumPrefix{\thelemmalist.\arabic*}}

\newtheorem{theorem}{Theorem}[section]

\newtheorem{lemmalist}{Lemma}[section]
\newtheorem{definition}{Definition}[section]
\newtheorem{corollary}{Corollary}[section]

\newtheorem{assumption}{Assumption}[section]
\numberwithin{equation}{section}

\newcommand{\E}{\operatorname{\mathbb{E}}}

\newcommand{\bx}{\boldsymbol{x}}

\newcommand{\eqdef}{\stackrel{\text{def}}{=}}

\numberwithin{equation}{section}

\newcommand{\remove}[1]{}

\usepackage{wrapfig}

\title{Stability Analysis and Generalization Bounds of Adversarial Training}
\author{%
  Jiancong Xiao$^{1,}\thanks{This work is done when Jiancong Xiao is a research intern at Tencent AI Lab, China.}$\ \ , Yanbo Fan$^{2,\dagger}$, Ruoyu Sun$^{1,3,}\thanks{Corresponding Authors.}$\ \ , Jue Wang$^{2}$, Zhi-Quan Luo$^{1,3}$\\
   $^1$The Chinese University of Hong Kong, Shenzhen; \\ 
   $^2$Tencent AI Lab; $^3$Shenzhen Research Institute of Big Data\\
  \texttt{jiancongxiao@link.cuhk.edu.cn, fanyanbo0124@gmail.com,} \\
  \texttt{sunruoyu@cuhk.edu.cn, arphid@gmail.com, luozq@cuhk.edu.cn}
}
\date{}

\begin{document}

\maketitle

\begin{abstract}
In adversarial machine learning, deep neural networks can fit the adversarial examples on the training dataset but have poor generalization ability on the test set. This phenomenon is called robust overfitting, and it can be observed when adversarially training neural nets on common datasets, including SVHN, CIFAR-10, CIFAR-100, and ImageNet. In this paper, we study the robust overfitting issue of adversarial training by using tools from uniform stability. One major challenge is that the outer function (as a maximization of the inner function) is nonsmooth, so the standard technique (\emph{e.g.,} \citep{hardt2016train}) cannot be applied. Our approach is to consider $\eta$-approximate smoothness: we show that the outer function satisfies this modified smoothness assumption with $\eta$ being a constant related to the adversarial perturbation $\epsilon$. Based on this, we derive stability-based generalization bounds for stochastic gradient descent (SGD) on the general class of $\eta$-approximate smooth functions, which covers the adversarial loss. Our results suggest that robust test accuracy decreases in $\epsilon$ when $T$ is large, with a speed between $\Omega(\epsilon\sqrt{T})$ and $\mathcal{O}(\epsilon T)$. This phenomenon is also observed in practice. Additionally, we show that a few popular techniques for adversarial training (\emph{e.g.,} early stopping, cyclic learning rate, and stochastic weight averaging) are stability-promoting in theory.

\end{abstract}
\section{Introduction}
\begin{wrapfigure}{r}{0.3\textwidth}
\vskip -0.25in
\centering
\includegraphics[width=1\linewidth]{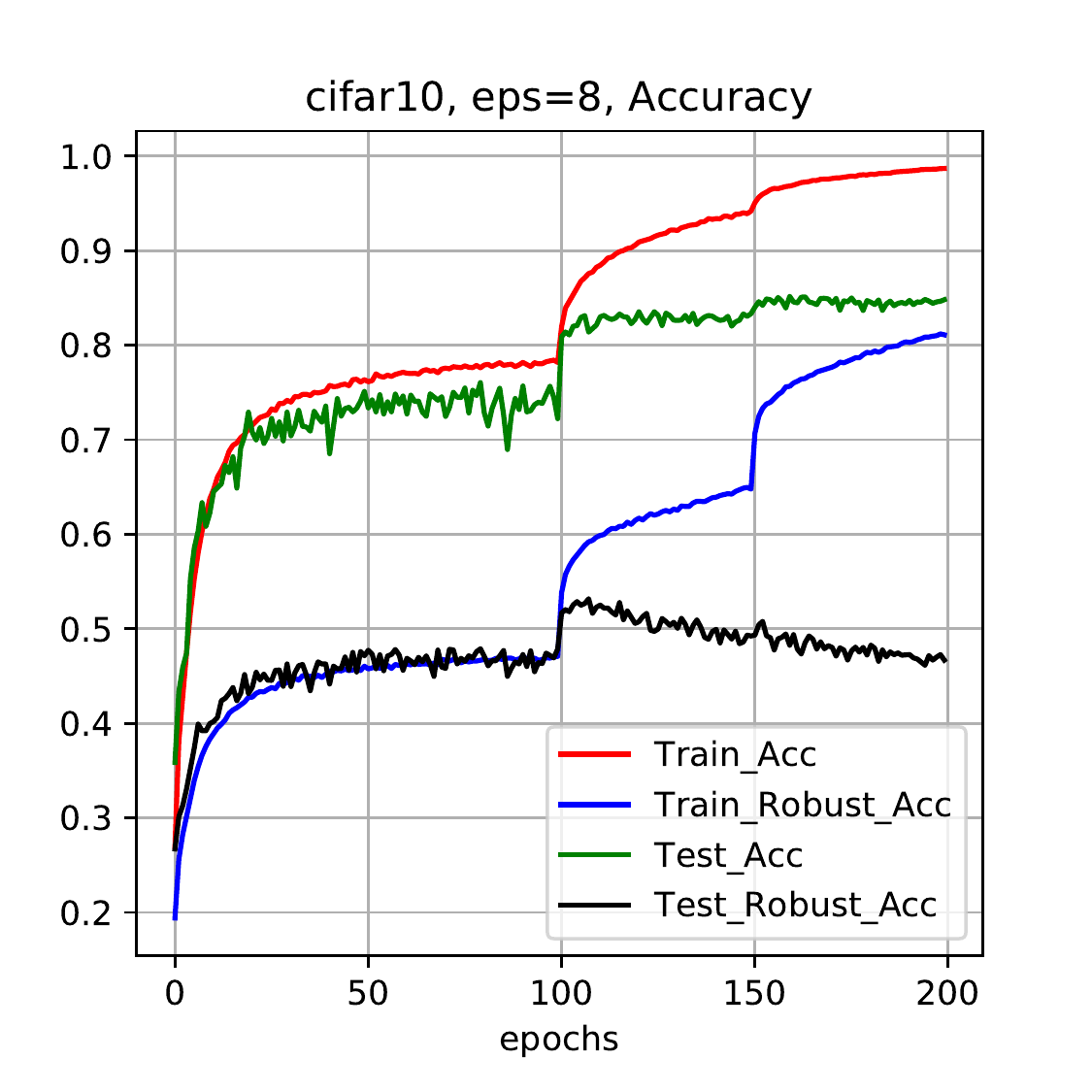}
\vskip -0.15in
\caption{Experiments of adversarial training on CIFAR-10.}
\label{fig:intro}
\vskip -0.1in
\end{wrapfigure}
Deep neural networks (DNNs) \citep{krizhevsky2012imagenet,hochreiter1997long} have become successful in many machine learning tasks and rarely suffered overfitting issues \citep{zhang2021understanding}. A neural network model can be trained to achieve zero training error and generalize well to the unseen data. While in the setting of adversarial training, robust overfitting is a dominant issue \citep{rice2020overfitting}. Specifically, robust overfitting characterizes a training procedure shown in Fig. \ref{fig:intro}. After a particular epoch, the robust test accuracy (black line) starts to decrease, but the robust training accuracy (blue line) is still increasing. This phenomenon can be observed in the experiments on common datasets, \emph{e.g.,} SVHN, CIFAR-10, CIFAR-100, and ImageNet. Recent works \citep{gowal2020uncovering,rebuffi2021fixing} mitigated the overfitting issue using regularization techniques such as stochastic weight averaging (SWA) and early stopping, but it still has a large generalization gap. Therefore, it is essential to study this issue from a theoretical perspective. In this paper, we study the robust overfitting issue of adversarial training by using tools from uniform stability.

Uniform stability analysis \citep{bousquet2002stability} in learning problems has been introduced to measure generalization gap instead of uniform convergence analysis such as classical VC-dimension \citep{vapnik2015uniform} and Rademacher complexity \citep{bartlett2002rademacher}. Generalization gap can be bounded in terms of uniform argument stability (UAS). Formally, UAS is the gap between the output parameters $\theta$ of running an algorithm $A$ on two datasets $S$ and $S'$ differ in at most one sample, denoted as $\delta(S,S')=\|\theta(S)-\theta(S')\|$. In a standard training problem with $n$ training samples, assuming that the loss function is convex, $L$-Lipschitz and $\beta$-gradient Lipschitz, running stochastic gradient descent (SGD) with step size $\alpha\leq 1/\beta$ for $T$ steps, the UAS is bounded by $\mathcal{O}(L\alpha T/n)$ \citep{hardt2016train}. Since the generalization gap is controlled by the number of samples $n$, this bound (at least partially) explains the good generalization of a standard training problem. Beyond \citep{hardt2016train}, \citet{bassily2020stability} considered the case that the loss function is non-smooth. Without the $\beta$-gradient Lipschitz assumption, they showed that the UAS is bounded by $\mathcal{O}(L\alpha\sqrt{T}+L\alpha T/n)$ and provided a lower bound to show that the additional term $\mathcal{O}(L\alpha\sqrt{T})$ is unavoidable. In adversarial settings, two works have discussed the stability of adversarial training to our knowledge.

Firstly, \citet{farnia2021train} considered the stability of minimax problems. Their work includes a discussion on an algorithm called GDmax (gradient descent on the maximization of the inner problem), which can be viewed as a general form of adversarial training. It provides the stability of GDmax when the inner problem is further assumed to be strongly concave. Under the strongly concave assumption, the outer function is smooth. Then, the generalization bound $\mathcal{O}(L\alpha T/n)$ can be applied in this case. However, the inner problem is not strongly concave in practice. The bound $\mathcal{O}(L\alpha T/n)$ does not match the poor generalization of adversarial training observed in practice.

Another stability analysis of adversarial training is the work of \citep{xing2021on}. Before they propose their algorithm, they use the stability bound $\mathcal{O}(L\alpha\sqrt{T}+L\alpha T/n)$ \citep{bassily2020stability} to characterize the issue of adversarial generalization. However, the bound is $\epsilon$-independent. Let us Consider two cases, $\epsilon\rightarrow 0$ and $\epsilon$ is large (\emph{e.g.,} 16/255). In the first case, adversarial training is very close to standard training and has good generalization ability. In the second case, adversarial generalization is very poor. The bound is the same in these two different cases. Therefore, it cannot provide an interpretation of adversarial generalization.

In summary, existing stability-based generalization bounds \citep{hardt2016train,bassily2020stability} have limited interpretation on adversarial generalization. In this paper, we provide a new stability analysis of adversarial training using the notion of $\eta$-approximate smoothness. We first show that, under the same assumptions as \citep{hardt2016train,xing2021on}, even though the outer function (adversarial loss) is non-smooth, it is $\eta$-approximately smooth (see Definition \ref{def:approx}), where $\eta$ is a constant linearly depending on the gradient Lipschitz of the inner 
function and the perturbation intensity $\epsilon$. Then, we derive stability-based generalization bounds (Thm. \ref{thm:covgen} to \ref{thm:stronggen}) for stochastic gradient descent (SGD) on this general class of $\eta$-approximate smooth functions, which covers the adversarial loss. Our main result can be summarized in the following equation. Running SGD on adversarial loss for $T$ steps with step size $\alpha\leq 1/\beta$, the excess risk, which is the sum of generalization error and optimization error, satisfies
\begin{equation}
    \text{Excess Risk}\leq\mathcal{E}_{gen}+\mathcal{E}_{opt}\leq \underbrace{\overbrace{L\eta T\alpha }^
    {\text{additional}}+\overbrace{\frac{2L^2 T\alpha}{n}+\frac{D^2}{T\alpha}+L^2\alpha}^{\text{for standard training}}}_{\text{for adversarial training}}.
\end{equation}

The excess risk of adversarial training has an additional term $L\eta T\alpha$. We also provide a lower bound of UAS of running SGD on adversarial loss. Our results suggest that robust test accuracy decreases in $\epsilon$ when $T$ is large, with a speed between $\Omega(\epsilon\sqrt{T})$ and $\mathcal{O}(\epsilon T)$. This phenomenon is also observed in practice. It provides an understanding of robust overfitting from the perspective of uniform stability. Additionally, we show that a few popular techniques for adversarial training (\emph{e.g.,} early stopping, cyclic learning rate, and stochastic weight averaging) are stability-promoting in theory and also empirically improve adversarial training. Experiments on SVHN, CIFAR-10, CIFAR-100, and ImageNet confirm our results. Our contributions are listed as follows:

\begin{itemize}
\item Main results: we derive stability-based generalization bounds for adversarial training using the notion of $\eta$-approximate smoothness. Based on this, we provide an analysis to understand robust overfitting.
\item We provide the stability analysis of a few popular techniques for adversarial training and show that they are indeed stability-promoting.
\item We provide experiments on SVHN, CIFAR-10, CIFAR-100, and ImageNet. The results verify the generalization bounds.
\item Technical contribution: we develop a set of properties of $\eta$-approximately smooth function, which might be useful in other tasks.
\end{itemize}

The paper is organized as follows. After discussing the related work in Sec. \ref{rel}, the rest of the paper contains two parts. The first part is the technical part to derive stability bounds. The second part is to analyze robust overfitting. Specifically, in the first part, Sec. \ref{updates:sec} introduces the preliminary knowledge about UAS. Sec. \ref{s4} provides the Lemma and properties of approximately smooth functions and Sec. \ref{s5} gives the stability bounds. In the second part, Sec. \ref{s6} analyzes the robust overfitting in the theoretical settings and Sec. \ref{s7} presents the experiments.

\section{Related Work} 
\label{rel}
\paragraph{Adversarial Attacks and Defense.} Starting from the work of \citep{szegedy2013intriguing}, it has been commonly realized that deep neural networks are highly susceptible to imperceptible corruptions to the input data \citep{goodfellow2014explaining,carlini2017towards,madry2017towards}. A series of work aimed at training neural networks robust to such small perturbations \citep{wu2020adversarial,gowal2020uncovering,zhang2020single} and another line of work aimed at designing more powerful adversarial attack algorithms \citep{athalye2018obfuscated,tramer2020adaptive,fan2020sparse,xiao2022understanding,qin2022boosting}. A series of work considered adversarial robustness in black-box settings \citep{chen2017zoo,qin2021random}. Semi-supervised learning has been used to improve adversarial robustness \citep{carmon2019unlabeled,li2022semi}. Fast adversarial training \citep{wong2020fast,huang2022fast} was introduced to save training time.

\paragraph{Adversarial Generalization.} A series of work tried to explain adversarial generalization in the uniform convergence framework, including VC-dimension \citep{attias2021improved,montasser2019vc} and Rademacher complexity \citep{khim2018adversarial,yin2019rademacher,awasthi2020adversarial,xiao2022adversarial}. Uniform algorithmic stability is another framework to study adversarial generalization \citep{farnia2021train,xing2021on,xiao2022adaptive}. The work of \citep{schmidt2018adversarially,raghunathan2019adversarial,zhai2019adversarially} have shown that in some scenarios achieving adversarial generalization requires more data. The work of \citep{xing2021generalization,xing2021adversarially,javanmard2020precise} studied the generalization in the setting of adversarial linear regression. \citep{sinha2017certifiable} studied the generalization of distributional robustness. The work of \citep{taheri2020asymptotic,javanmard2020precise,dan2020sharp} analyzed adversarial generalization in Gaussian mixture models.

\paragraph{Uniform Stability.} Stability is a classical approach to provide generalization bounds. It can be traced back to the work of \citep{rogers1978finite}. After a few decades, it was well developed in analyzing the generalization bounds in statistical learning problems \citep{bousquet2002stability}. These bounds have been significantly improved in a recent sequence of works \citep{feldman2018generalization,feldman2019high}. The work of
\citep{chen2018stability} derived minimax lower bounds for excess risk and discussed the optimal trade-off between stability and convergence. \citet{ozdaglar2022good} considered the generalization metric of minimax optimizer.

\section{Preliminaries of Stability} \label{updates:sec}

Consider the following setting of statistical learning. There is an
unknown distribution~${\cal D}$ over examples from some
space $\mathcal{Z}$. We receive a sample dataset $S=\{z_1,\dots,z_n\}$
of $n$ examples drawn
i.i.d.~from ${\cal D}.$
The \emph{population risk} and \emph{empirical risk} are defined as:
\[
R_{\cal D}(\theta)\eqdef \E_{z\sim{\cal D}} h(\theta,z)\quad\text{and}\quad R_S(\theta)\eqdef\frac1n\sum_{i=1}^n h(\theta,z_i),
\]
respectively, where $h(\cdot,\cdot)$ is the loss function.

\paragraph{Risk Decomposition.} Let $\theta^*$ and $\bar\theta$ be the optimal solution of $R_{\cal D}(\theta)$ and $R_S(\theta)$ respectively. Then for the algorithm output $\hat\theta=A(S)$, the excess risk can be decomposed as
\begin{equation*}
\begin{aligned}
    R_{\cal D}(\hat\theta)-R_{\cal D}(\theta^*)
    =\underbrace{R_{\cal D}(\hat\theta)-R_{S}(\hat\theta)}_{\mathcal{E}_{gen}}
    +\underbrace{R_{S}(\hat\theta)-R_{S}(\bar\theta)}_{\mathcal{E}_{opt}}
    +\underbrace{R_{S}(\bar\theta)-R_{S}(\theta^*)}_{\leq 0}+\underbrace{R_{S}(\theta^*)-R_{\cal D}(\theta^*)}_{\mathbb{E}=0}.
    \end{aligned}
\end{equation*}
To control the excess risk, we need to control the generalization gap $\mathcal{E}_{gen}$ and the optimization gap $\mathcal{E}_{opt}$. In the rest of the paper, we use $\mathcal{E}_{gen}$ and $\mathcal{E}_{opt}$ to denote the \emph{expectation} of the generalization and optimization gap. To bound the generalization gap of a model $\hat\theta=A(S)$ trained by a randomized algorithm $A$,  we employ the
following notion of \emph{uniform stability}.

\begin{definition}
A randomized algorithm $A$ is $\varepsilon$-\emph{uniformly stable} if
for all data sets $S,S'\in \mathcal{Z}^n$ such that $S$ and $S'$ differ in at most one
example, we have
\begin{equation}\label{eq:stab}
\sup_{z} \E_{A} \left[ h(A(S); z) - h(A(S'); z) \right] \le \varepsilon\,.
\end{equation}
\end{definition}
Here, the expectation is taken only over the internal randomness of $A$. We recall the important theorem that uniform stability implies
\emph{generalization in expectation} \citep{hardt2016train}.

\begin{theorem}[Generalization in expectation]
\label{thm:stab2gen}
Let $A$ be $\varepsilon$-uniformly stable. Then, the expected generalization gap satisfies
\[
|\mathcal{E}_{gen}|=\left| \E_{S,A}[R_{\cal D}[A(S)] - R_S[A(S)]]\right| \le \varepsilon\,.
\]
\end{theorem}
Therefore, we turn to the properties of iterative algorithms that control their uniform stability.
\section{Stability of Adversarial Training}

\paragraph{Adversarial Surrogate Loss.} 
\label{s4}
In adversarial training, we consider the following surrogate loss
\begin{equation}
\label{advloss}
h(\theta;z)=\max_{\|z-z'\|_p\leq\epsilon}g(\theta;z'),
\end{equation}
where $g(\theta;z)$ is the loss function of the standard counterpart, $\|\cdot\|_p$ is the $\ell_p$-norm, $p\geq 1$. Usually, $g$ can also be written in the form of $\ell(f_{\theta}(\bx);y)$, where $f_\theta$ is the neural network to be trained and $(\bx,y)$ is the input-label pair. We assume the loss function $g$ satisfies the following smoothness assumption.
\begin{assumption}
	\label{ass1}
	The function $g$ satisfies the following Lipschitzian smoothness conditions:
	\begin{equation*}
	\begin{aligned}
	&\|g(\theta_1,z)-g(\theta_2,z)\|\leq L\|\theta_1-\theta_2\|,\\
	&\|\nabla_\theta g(\theta_1,z)-\nabla_\theta g(\theta_2,z)\|\leq L_{\theta}\|\theta_1-\theta_2\|,\\
	&\|\nabla_\theta g(\theta,z_1)-\nabla_\theta g(\theta,z_2)\|\leq L_{z}\|z_1-z_2\|_p,\\
	\end{aligned}
	\end{equation*}
	where $\|\cdot\|$ is Euclidean norm.
\end{assumption}
Assumption \ref{ass1} assumes that the loss function is smooth, which are also used in the stability literature \citep{farnia2021train,xing2021on}, as well as the convergence analysis literature \citep{wang2019convergence,liu2020loss}. While ReLU activation function is non-smooth, recent works \citep{allen2019convergence,du2019gradient} showed that the loss function of over-parameterized DNNs is semi-smooth. It helps justify Assumption \ref{ass1}. Under Assumption \ref{ass1}, the loss function of adversarial training satisfies the following Lemma \citep{liu2020loss}.

\begin{lemmalist}
\label{lem:nonsmooth}
	Let $h$ be the adversarial loss defined in Eq. (\ref{advloss}) and $g$ satisfies Assumption \ref{ass1}. $\forall \theta_1, \theta_2$ and $\forall z\in\mathcal{Z}$, the following properties hold.
\begin{senenum}
\item \label{lem:1}
(Lipschitz function.) $\|h(\theta_1,z)-h(\theta_2,z)\|\leq L\|\theta_1-\theta_2\|$.

\item \label{lem:2} For all subgradient $d(\theta,z)\in\partial_\theta h(\theta,z)$, we have $\|d(\theta_1,z)-d(\theta_2,z)\|\leq L_{\theta}\|\theta_1-\theta_2\|+2L_{z}\epsilon$.  
\end{senenum}
\end{lemmalist}

If we further assume that $g(\theta,z)$ is $\mu$-strongly concave in $z$, the adversarial surrogate loss $h(\theta,z)$ is also smooth \citep{sinha2017certifiable}. Therefore, the uniform stability of adversarial training follows \citep{hardt2016train,farnia2021train}, see Appendix \ref{b3}. However, $g(\theta,z)$ is non-strongly concave in practice. The above results provide limited explanations of the poor generalization of adversarial training. We discuss the generalization properties of adversarial training under Lemma \ref{lem:2}.

\subsection{Basic Properties of Approximate Smoothness}
To simplify the notation, we use $h(\theta)$ as a shorthand notation of $h(\theta,z)$. To simplify the argument, we consider differentiable function $h$. The results can be extended to non-differentiable cases. Lemma \ref{lem:nonsmooth} motivates us to analyze a function with the following modified smoothness assumption, which we call approximate smoothness assumption.

\begin{definition}
\label{def:approx}
Let $\beta>0$ and $\eta>0$. We say a differentiable function $h(\theta)$ is $\eta$-approximately $\beta$-gradient Lipschitz, if $\forall \theta_1$ and $\theta_2$, we have
\begin{equation*}
    \|\nabla h(\theta_1)-\nabla h(\theta_2)\|\leq \beta\|\theta_1-\theta_2\|+\eta.
\end{equation*}
\end{definition}
In definition \ref{def:approx}, $\eta$ controls the smoothness of the loss function $h(\cdot)$. When $\eta=0$, the function $h$ is gradient Lipschitz. When $\eta\rightarrow +\infty$, $h$ is a general non-smooth function. As our discussion before, the adversarial surrogate loss is $2L_z\epsilon$-approximately smooth. As far as we know, this assumption is rarely discussed in the optimization literature since it cannot improve the convergence rate from a general non-smooth assumption. But it affects uniform stability, as we will discuss later. We need to develop the basic properties of approximate smoothness first.

\begin{lemmalist}\label{lem:cocoercive}
	Assume that the function $h$ is $\eta$-approximately $\beta$-gradient Lipschitz. $\forall \theta_1, \theta_2$ and $\forall z\in\mathcal{Z}$, the following properties hold.
	\vspace{-0.1in}
\begin{senenum}
\item \label{lem:3} ($\eta$-approximate descent lemma.) 
\vspace{-0.1in}
\begin{equation*}
\begin{aligned}
    h(\theta_1)- h(\theta_2)
    \leq\nabla h(\theta_2)^T(\theta_1-\theta_2)+\frac{\beta}{2}\|\theta_1-\theta_2\|^2+\eta \|\theta_1-\theta_2\|.
    \end{aligned}
\end{equation*}
\vspace{-0.15in}
\item \label{lem:4} ($\eta$-approximately co-coercive.) Assume in addition that $h(\theta,z)$ is convex in $\theta$ for all $z\in\mathcal{Z}$. Let $[\cdot]_+=\max(0,\cdot)$. We have
\vspace{-0.1in}
\begin{equation*}
\begin{aligned}
    \langle\nabla h(\theta_1)-\nabla h(\theta_2),\theta_1-\theta_2\rangle\geq\frac{1}{\beta}\bigg[[\|\nabla h(\theta_1)-\nabla h(\theta_2)\|-\eta]_+\bigg]^2.
\end{aligned}
\end{equation*}
\vspace{-0.1in}
\end{senenum}
\end{lemmalist}

We defer the proof to Appendix \ref{a}. Note that the loss function is
$L$-Lipschitz for every example $z$, we have $\mathbb{E}|h(\theta_1,z)-h(\theta_2,z)|\leq L \mathbb{E}\|\theta_1-\theta_2\|$, for all $z\in\mathcal{Z}.$ To obtain the stability generalization bounds, we need to analysis the difference $\|\theta_1^T-\theta_2^T\|$, where $\theta_1^T$ and $\theta_2^T$ are the outputs of running SGD on adversarial surrogate loss for $T$ iterations on two datasets with only one different sample. Next we provide the recursive bounds under the approximate smoothness assumption.

\paragraph{Algorithms.}
We consider the stochastic gradient descent on the adversarial surrogate loss. i.e.,
\begin{equation}
    \theta^{t+1}=\theta^t-\alpha_t\nabla_\theta h(\theta^t,z_{i_t}),
\end{equation}
where $\alpha_t$ is the step size in iteration $t$, $z_{i_t}$ is the sample chosen in iteration $t$. We consider two popular schemes for
choosing the examples’ indices $i_t$. \emph{Sampling with replacement}: One is to pick $i_t\sim \text{Uniform}\{1,\cdots,n\}$ at each step. \emph{Fixed permutation}: The other is to choose a random permutation over $\{1,\cdots,n\}$ and cycle through the examples
repeatedly in the order determined by the permutation. Our results hold for both variants.

\paragraph{Properties of Update Rules.} We define $G_{\alpha,z}(\theta)=\theta-\alpha\nabla h(\theta,z)$ be the update rule of SGD. The following lemma holds.

\begin{lemmalist}\label{updaterules}
	Assume that the function $h$ is $\eta$-approximately $\beta$-gradient Lipschitz. $\forall \theta_1, \theta_2$ and $\forall z\in\mathcal{Z}$,we have
\begin{senenum}
\item \label{lem:5} ($\alpha\eta$-approximately $(1+\alpha\beta)$-expansive.) $\|G_{\alpha,z}(\theta_1)-G_{\alpha,z}(\theta_2)\|\leq (1+\alpha\beta)\|\theta_1-\theta_2\|+\alpha\eta.$
\item \label{lem:6} ($\alpha\eta$-approximately non-expansive.) Assume in addition that $h(\theta,z)$ is convex in $\theta$ for all $z\in\mathcal{Z}$, for $\alpha\leq 1/\beta$, we have $\|G_{\alpha,z}(\theta_1)-G_{\alpha,z}(\theta_2)\|\leq \|\theta_1-\theta_2\|+\alpha\eta.$
\item \label{lem:7} ($\alpha\eta$-approximately $(1-\alpha\gamma)$-contractive.) Assume in addition that $h(\theta,z)$ is $\gamma$-strongly convex in $\theta$ for all $z\in\mathcal{Z}$, for $\alpha\leq 1/\beta$, we have
$\|G_{\alpha,z}(\theta_1)-G_{\alpha,z}(\theta_2)\|\leq (1-\alpha\gamma)\|\theta_1-\theta_2\|+\alpha\eta.$
\end{senenum}
\end{lemmalist}

The proof of Lemma \ref{updaterules} is based on Lemma \ref{lem:cocoercive} and is deferred to Appendix \ref{a}. Lemma \ref{updaterules} provides the recursive distance $\|\theta_1-\theta_2\|$ from iteration $t$ to $t+1$. Based on this, we can recursively derive the distance $\|\theta_1^T-\theta_2^T\|$. Then, we can obtain the stability generalization bounds.
\section{Stability Generalization Bounds}
\label{s5}
In the previous section, we have discussed the properties of approximate smoothness and developed tools we need to use. In this section, we discuss the stability bounds. 
\subsection{Convex Optimization}
We first consider the case that $h(\theta,z)$ is convex in $\theta$ for all $z\in \mathcal{Z}$.

\begin{theorem}
\label{thm:covgen}
	Assume that $h(\theta,z)$ is convex, $L$-Lipschitz, and $\eta$-approximately $\beta$-gradient Lipschitz in $\theta$ for all given $z\in\mathcal{Z}$. Suppose that we run SGD with step sizes $\alpha_t\leq 1/\beta$ for $T$ steps. Then, 
\begin{equation}\small
\label{eq:gen}
    \mathcal{E}_{gen}=\mathbb{E}[R_\mathcal{D}(\theta^T)-R_S(\theta^T)]\leq L\bigg(\eta +\frac{2L}{n}\bigg)\sum_{t=1}^T\alpha_t.
\end{equation}
\end{theorem}

The proof is mainly based on Lemma \ref{updaterules} that the update rule is approximately non-expansive in this case. We defer it to appendix \ref{a}. We have shown that the adversarial surrogate loss is $2L_z\epsilon$-approximately $L_\theta$-gradient Lipschitz. Let $\eta=2L_z\epsilon$ in Eq. (\ref{eq:gen}), we directly obtain the stability bounds for adversarial training. In practice, the solution of the inner problem is sub-optimal. Let $\Delta\epsilon$ be the maximum error between the optimal and sub-optimal attacks in each iteration. We have the following Corollary.

\begin{corollary}[Uniform stability for sub-optimal attacks adversarial training.]
\label{thm:corogen}
	Under Assumption \ref{ass1}, assume in addition that $g(\theta,z)$ is convex in $\theta$ for all given $z\in\mathcal{Z}$. Suppose that we run adversarial training with step sizes $\alpha_t\leq 1/L_\theta$ for $T$ steps. Then, adversarial training satisfies
uniform stability with
\begin{equation}\label{ATbound}
\begin{aligned}
        \mathcal{E}_{gen}  \leq \bigg(2 LL_{z}(\epsilon+\Delta\epsilon) +\frac{2L^2}{n}\bigg)\sum_{t=1}^T\alpha_t\leq \mathcal{O}\bigg( L(L_{z}\epsilon +\frac{L}{n})\sum_{t=1}^T\alpha_t\bigg).
\end{aligned}
\end{equation}
\end{corollary}
\textbf{Remark:} Corollary \ref{thm:corogen} shows that adversarial training with weak attacks (large $\Delta\epsilon$) have worse adversarial generalization than that  with strong attacks. This is also observed in practice. However, $\Delta\epsilon$ is at most $2\epsilon$, the upper bound of AT with different attacks have the same order. It might be due to the weakness of Assumption \ref{ass1} or uniform stabilty framework.

\paragraph{Interpreting Robust Generalization.} If we let $\epsilon=0$ in Eq. (\ref{ATbound}) , it reduced to the generalization bound in \citep{hardt2016train} for standard training. Therefore, the additional generalization error of adversarial training comes from the additional term $L_z\epsilon$. The global gradient Lipschitz $L_z$ with respect to $z$ plays an important role in generalization. Even though the perturbation $\epsilon$ is small, it is amplified by the Lipschitz $L_z$ and finally hurts the robust generalization.

\subsection{Further Discussion on the Generalization Bounds}

We first provide a lower bound. Then we compare our bounds with the existing bounds in Table \ref{tab:compare}.
\begin{theorem}[Lower Bound]
\label{thm:lower}
	There exists functions $h(\theta,z)$, s.t. $h$ is convex, $L$-Lipschitz, and $\eta$-approximately $\beta$-gradient Lipschitz in $\theta$ for all given $z\in\mathcal{Z}$. Exists $S$ and $S'$ differ in one sample. Suppose that we run SGD with step fixed step sizes $\alpha\leq 1/\beta$ for $T$ steps. Then, 
\begin{equation}\small
    \mathbb{E}[\delta(S,S')]\geq \Omega\bigg(\eta\alpha\sqrt{T} +\frac{L\alpha T}{n}\bigg).
\end{equation}
\end{theorem}

\begin{table}[htbp]
    \centering
        \caption{Comparison of the upper and lower bounds of $\mathbb{E}[\delta(S,S')]$. Comparing with the previous results, we replace $L$ by $\eta$ and provide the matching lower bound in $\eta$.}
        \label{tab:compare}
    \resizebox{\linewidth}{!}{%
    \begin{tabular}{cccc}
    \hline
        & Assumption &  Upper bounds & Lower bounds\\
          \hline
          \citet{farnia2021train} & convex-strongly concave &$\mathcal{O}(LT\alpha/n)$ &$\Omega(LT\alpha/n)$\\
          \citet{xing2021on} & convex-nonconcave&  $\mathcal{O}(L\sqrt{T}\alpha+LT\alpha/n)$ &$\Omega(\sqrt{T}\alpha+LT\alpha/n)$ \\
        Ours & convex-nonconcave & $\mathcal{O}(\eta T\alpha+LT\alpha/n)$ &$\Omega(\eta\sqrt{T}\alpha+LT\alpha/n)$ \\
          \hline
    \end{tabular}}
\end{table}

\textbf{Comparison with the Existing UAS Bounds.} Compared with the work of \citep{farnia2021train}, they assume that the inner problem is strongly concave. Thus the bounds are not comparable. Strongly concave is a strong assumption in practice. Therefore, the work of \citep{xing2021on} and our analysis focus on the nonconcave cases. Comparing with the bound $\mathcal{O}(L\sqrt{T}\alpha+LT\alpha/n)$, our bound captures a critical aspect of adversarial generalization bound: $\epsilon$-dependent. As observed in practice, the adversarial generalization gap reduces to the standard generalization gap as $\epsilon\rightarrow 0$. Our bound consists with this observation. On the contrary, the bound $\mathcal{O}(L\sqrt{T}\alpha+LT\alpha/n)$ is very large when $\epsilon\rightarrow 0$. Additionally, we provide a matching lower bound w.r.t $\eta$. The comparison is provided in Table \ref{tab:compare}.

 \textbf{Comparison with the Work of \citep{xing2021on}.} The work of \citep{xing2021on} argued that the max function is not smooth even though the standard counterpart is smooth. Therefore, they followed the bound in non-smooth cases \citep{bassily2020stability}. Then, they aimed to solve the non-smooth issue. They design a noise-injected algorithm and show its effectiveness in tackling the non-smooth issue. Our work focus on providing better bounds to interpret robust overfitting.

\subsection{Non-convex Optimization and Strongly Convex Optimization}
Next, we consider the case that the loss function $h$ is general non-convex and strongly convex. By Lemma \ref{updaterules}, we have
\begin{theorem}
\label{thm:noncovgen}
	Assume that $h(\theta,z)$ is $L$-Lipschitz, and $\eta$-approximately $\beta$-gradient Lipschitz in $\theta$ for all given $z\in\mathcal{Z}$. Assume in addition that $0\leq h(\theta,z)\leq B$ for all $\theta$ and $z$. Suppose that we run SGD with diminishing step sizes $\alpha_t\leq 1/(\beta t)$ for $T$ steps. Then
\begin{equation}
    \mathcal{E}_{gen}\leq \frac{BL_\theta+(2L^2+L\eta  n)T}{\beta(n-1)}.
\end{equation}
\end{theorem}

\begin{theorem}
	\label{thm:stronggen}
	Assume that $h(\theta,z)$ is $\gamma$-strongly convex, $L$-Lipschitz, and $\eta$-approximately $\beta$-gradient Lipschitz in $\theta$ for all given $z\in\mathcal{Z}$. Suppose that we run SGD with step sizes $\alpha_t\leq 1/\beta$ for $T$ steps. Then
	\begin{equation}
	\mathcal{E}_{gen}=\mathbb{E}[R_\mathcal{D}(\theta^T)-R_S(\theta^T)]\leq \frac{L\eta}{\gamma}+\frac{2L^2}{\gamma n}.
	\end{equation}
\end{theorem}
\paragraph{Remark:} The bound in non-convex cases provides a similar interpretation of robust generalization to the analysis in the convex case. In uniform stability analysis, whether the loss function is convex or non-convex does not give a major difference. Therefore, we provide the analysis of the non-convex case in Appendix \ref{b}. We also provide our convergence analysis of running SGD on a $\eta$-approximate smoothness, non-convex function in Theorem \ref{converge2}. Strongly convex is a strong assumption. We leave the analysis of the bound in strongly convex cases in Appendix \ref{b2}.

\section{Excess Risk Minimization}
\label{s6}
Based on the risk decomposition, we have $\text{Excess Risk}\leq \mathcal{E}_{gen}+\mathcal{E}_{opt},$
we need to minimize $\mathcal{E}_{gen}+\mathcal{E}_{opt}$ to achieve better performance. Per our previous discussion, whether the loss function is convex or non-convex does not give a major difference in stability analysis. We study the convex case in this section. We leave the discussion on the non-convex and strongly convex cases in Appendix \ref{b1} and \ref{b2}, respectively. We first introduce the optimization error. 
\paragraph{Optimization Analysis.}
The convergence analysis of SGD on a $L$-Lipschitz, convex function is discussed in \citep{nemirovski2009robust}. The convergence rate cannot be improved if we further assume that the function $h$ is gradient Lipschitz. Therefore, a weaker condition, approximately gradient Lipschitz, cannot improve the convergence rate. We use the following convergence error bound (adopted from \citep{nemirovski2009robust}) for the optimization error of both adversarial training and standard training.
\begin{theorem}
\label{thm:covopt}
	Assume that $h(\theta,z)$ is L-Lipschitz and convex in $\theta$ for all given $z\in\mathcal{Z}$. Let $D=\|\theta^0-\theta^*\|$, where $\theta^0$ is the initialization of SGD. Suppose that we run SGD with step sizes $\alpha_t$ for $T$ steps. Then, $\exists k\leq T$, s.t.
\begin{equation}\label{eq:opt}
    \mathcal{E}_{opt}(\theta^k)\leq \frac{D^2+L^2\sum_{t=1}^T\alpha_t^2}{\sum_{t=1}^T\alpha_t}.
\end{equation}
\end{theorem}
If we let $\alpha_t=1/\sqrt{T}$, we have $\mathcal{E}_{opt}\leq \mathcal{O}(1/\sqrt{T})$, which is the convergence rate of SGD on convex function. Since we need to consider the generalization and optimization errors simultaneously, we keep the $\alpha_t$ in Theorem \ref{thm:covopt}.

\paragraph{Generalization-Optimization Trade-off.} 
We have now discussed the optimization and generalization errors of adversarial training. We aim to find the optimal trade-off between generalization and optimization in terms of $\alpha_t$ and $T$. However, finding the optimal $\alpha_t$ and $T$ simultaneously is challenging. We consider the settings we use in practice. 

\paragraph{Fixed Step Size.}
We first consider the simplest case, the step size $\alpha$ is fixed. Then, combining Eq. (\ref{eq:gen}) and Eq. (\ref{eq:opt}), we have
\begin{equation}
\label{eq:tradeoff}
    \mathcal{E}_{gen}+\mathcal{E}_{opt}\leq \underbrace{\overbrace{L\eta T\alpha }^
    {additional}+\overbrace{\frac{2L^2 T\alpha}{n}+\frac{D^2}{T\alpha}+L^2\alpha}^{\text{for standard training}}}_{\text{for adversarial training}}.
\end{equation}

\paragraph{Interpretation of Robust Overfitting.} In standard training, overfitting is rarely observed in practice. The optimization error and generalization error are both small. In Eq. (\ref{eq:tradeoff}), the second to the fourth terms are for standard training. The second term is controlled by the number of samples $n$, which is small if we have sufficient training samples. The third term is controlled by $T$, and the last term is fixed given a small $\alpha$. This bound partially explains the good performance of standard training. However, we have an additional term $L\eta T\alpha$ for excess risk in adversarial training. Then, after a particular iteration that $L\eta T\alpha$ dominates Eq. (\ref{eq:tradeoff}), robust overfitting appears. This is consistent with the training procedure in practice. Therefore, the $\eta=2L_z\epsilon$ approximate smoothness of adversarial loss provides a possible explanation of robust overfitting. To achieve better performance, we need to stop training the model earlier.

\paragraph{Early Stopping.} It is shown that early stopping is an important training technique for adversarial training \citep{rice2020overfitting}. In Eq. (\ref{eq:tradeoff}), if we optimize the right-hand-side with respect to $T$, we have
\begin{equation*}
\begin{aligned}
    T^*=\frac{\|\theta^0-\theta^*\|}{\alpha\sqrt{L\eta +2L^2/n}},\quad
    \mathcal{E}_{gen}+\mathcal{E}_{opt}\leq 2\sqrt{L\eta +\frac{2L^2}{n}}D+L^2\alpha.
    \end{aligned}
\end{equation*}
Therefore, it is the best to stop training at $T^*$. However, $T^*$ is unknown in practice. It is important to select stopping criteria. For example, we can use a validation set to determine when to stop.

\paragraph{Varying Step Size.}
We discuss one popular varying step size schedule, cyclic learning rate, which is also called super-converge learning rate for adversarial training. It is shown that it can speed up adversarial training with fewer epochs \citep{wong2020fast}. The Super-converge learning rate follows the following rules. In the first phase (warm-up), the step size increase from 0 to $\alpha'$ linearly. In the second phase (cold down), the step size decreases back to 0 linearly. It is unclear (to our knowledge) why this schedule can speed up convergence in optimization theory. But the generalization part can partially be explained by UAS. If we set $\alpha'=2\alpha$, it is easy to check that $\mathcal{E}_{gen}\leq (L\eta +2L^2/n)T\alpha$ in this case, which is the same as the bound in the fixed learning rate case. Notice that the cyclic learning rate usually requires fewer steps $T$ to converge. Then, the generalization gap is smaller. Cyclic learning rate can be viewed as another form of early stopping from the perspective of UAS.

\paragraph{Stochastic Weight Averaging.} SWA is also a useful training technique for adversarial training \citep{hwang2021adversarial}. Instead of using the last checkpoint, SWA suggests using the average of the checkpoints for inference. It is shown that SWA can find a model with better generalization since it leads to wider minima \citep{izmailov2018averaging}. Below we study SWA from the perspective of UAS.
\begin{theorem}
\label{thm:swagen}
	Assume that $h(\theta,z)$ is convex, $L$-Lipschitz, and $\eta$-approximately $\beta$-gradient Lipschitz in $\theta$ for all given $z\in\mathcal{Z}$. Suppose that we run SGD with step sizes $\alpha_t\leq 1/\beta$ for $T$ steps. Let $\bar{\theta}$ be the average of the trajectory. Then, 
\begin{equation}
\label{eq:swagen}
    \mathcal{E}_{gen}(\bar{\theta})\leq \bigg(\frac{L\eta}{2} +\frac{L^2}{n}\bigg)\sum_{t=1}^T\alpha_t,\quad 
    \mathcal{E}_{opt}(\bar{\theta})\leq \frac{\|\theta^0-\theta^*\|^2+L^2\sum_{t=1}^T\alpha_t^2}{\sum_{t=1}^T\alpha_t}.
\end{equation}
\end{theorem}
In words, SWA reduces the generalization error bound to one-half of the one without SWA. But the training error bound remains unchanged. 

\section{Experiments}
\label{s7}
\paragraph{Training Settings.} We mainly consider the experiments on CIFAR-10 \citep{krizhevsky2009learning}, CIFAR-100, and SVHN \citep{netzer2011reading}. We also provide one experiment on ImageNet \citep{deng2009imagenet}. For the first three datasets, we conduct the experiments on training PreActResNet-18, which follows \citep{rice2020overfitting}, For the experiment on ImageNet, we use ResNet-50 \citep{he2016deep}, following the experiment of \citep{madry2017towards}. For the inner problems, we adopt the $\ell_\infty$ PGD adversarial training in \citep{madry2017towards}, the step size in the inner maximization is set to be $\epsilon/4$ on CIFAR-10 and CIFAR100 and is set to be $\epsilon/8$ on SVHN. Weight decay is set to be $5\times10^{-4}$. Additional experiments are provided in Appendix \ref{a3}. \footnote{\url{https://github.com/JiancongXiao/Stability-of-Adversarial-Training}}

\begin{figure*}[htbp]
	\centering
	\hspace{-0.4in}\scalebox{0.9}{
		\subfigure[]{
			\begin{minipage}[htp]{0.26\linewidth}
				\centering
				\includegraphics[width=1.6in]{./figures/cifar_eps8.pdf}
			\end{minipage}%
		}
		\subfigure[]{
			\begin{minipage}[htp]{0.26\linewidth}
				\centering
				\includegraphics[width=1.6in]{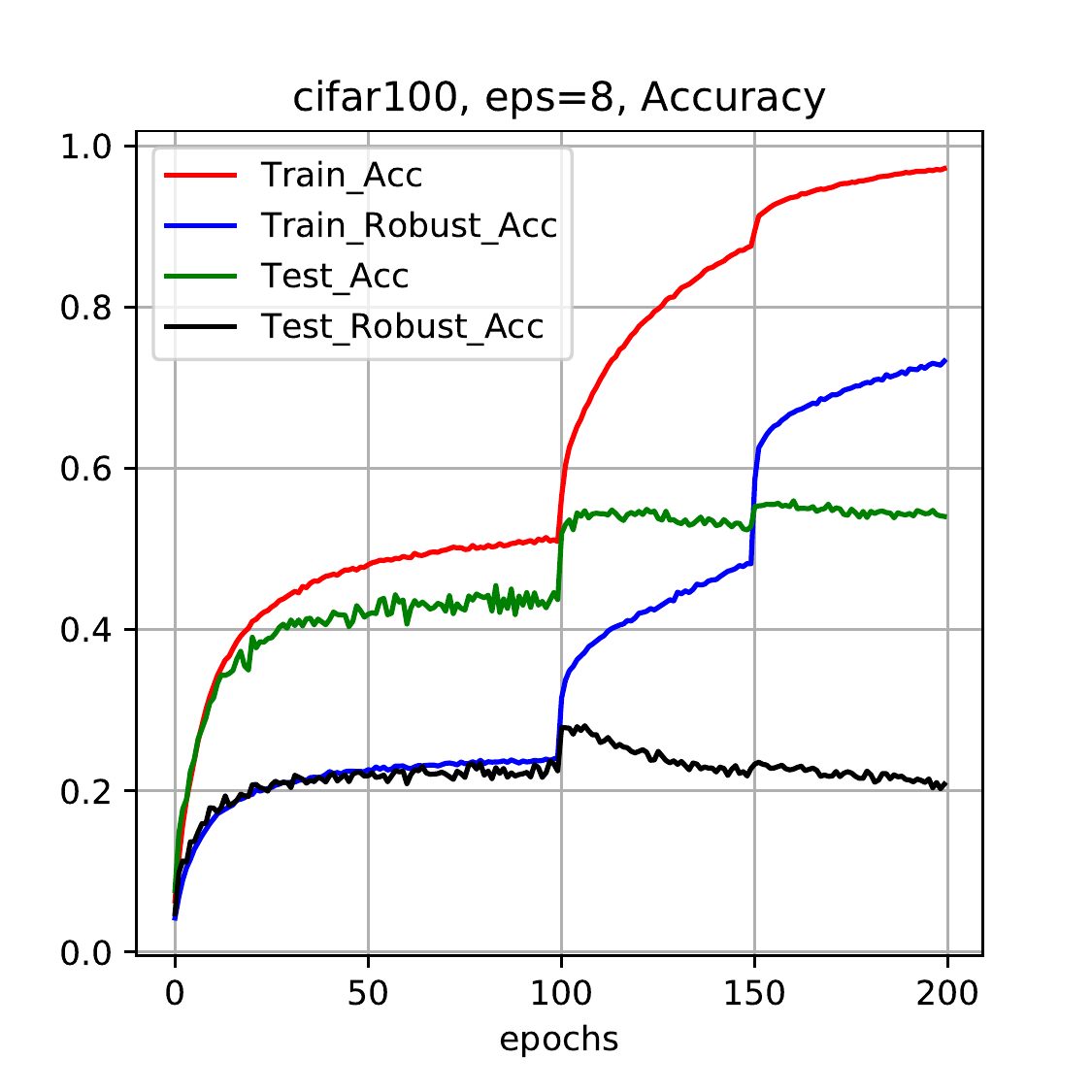}
			\end{minipage}
		}
		\subfigure[]{
			\begin{minipage}[htp]{0.26\linewidth}
				\centering
				\includegraphics[width=1.6in]{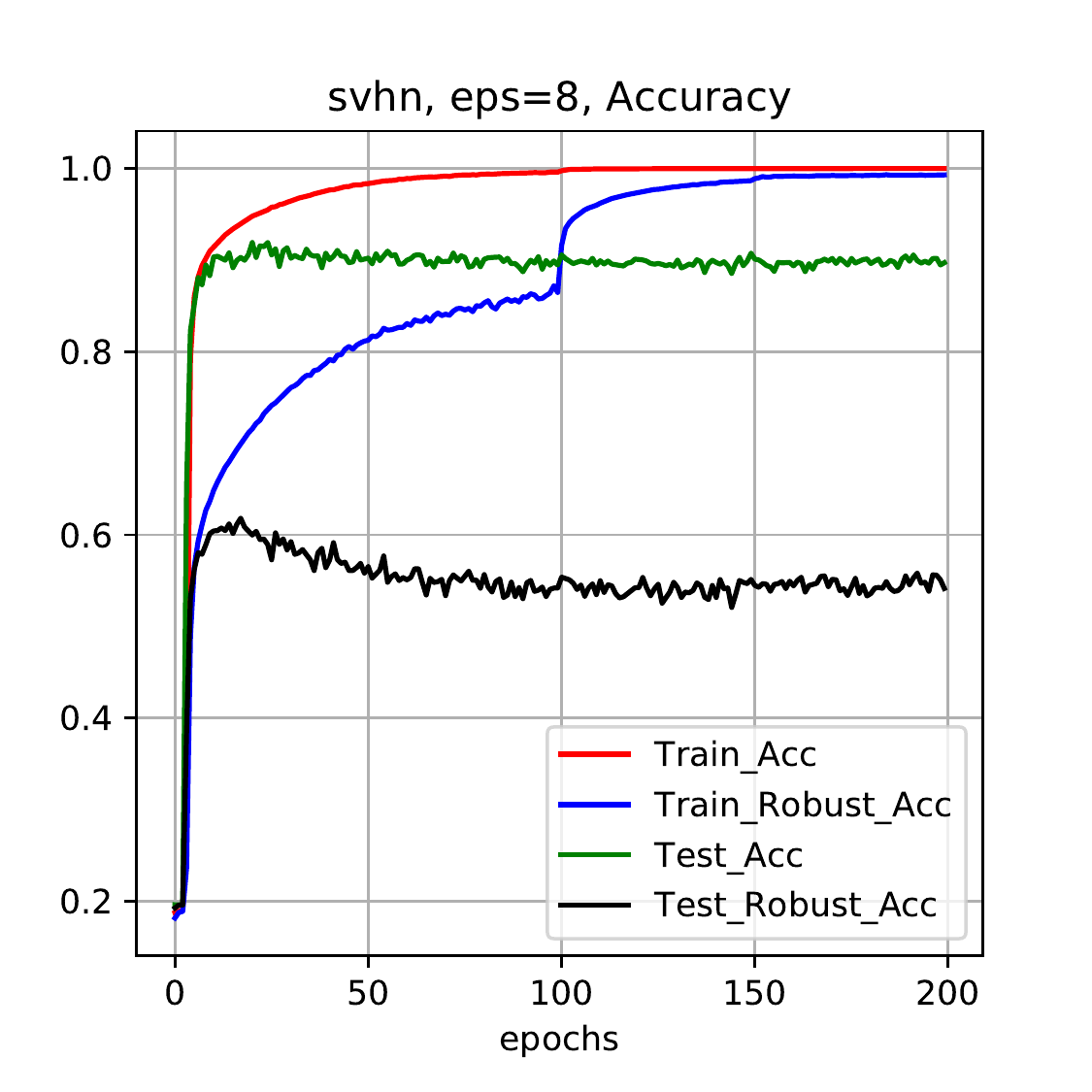}
			\end{minipage}
		}
		\subfigure[]{
			\begin{minipage}[htp]{0.26\linewidth}
				\centering
				\includegraphics[width=1.6in]{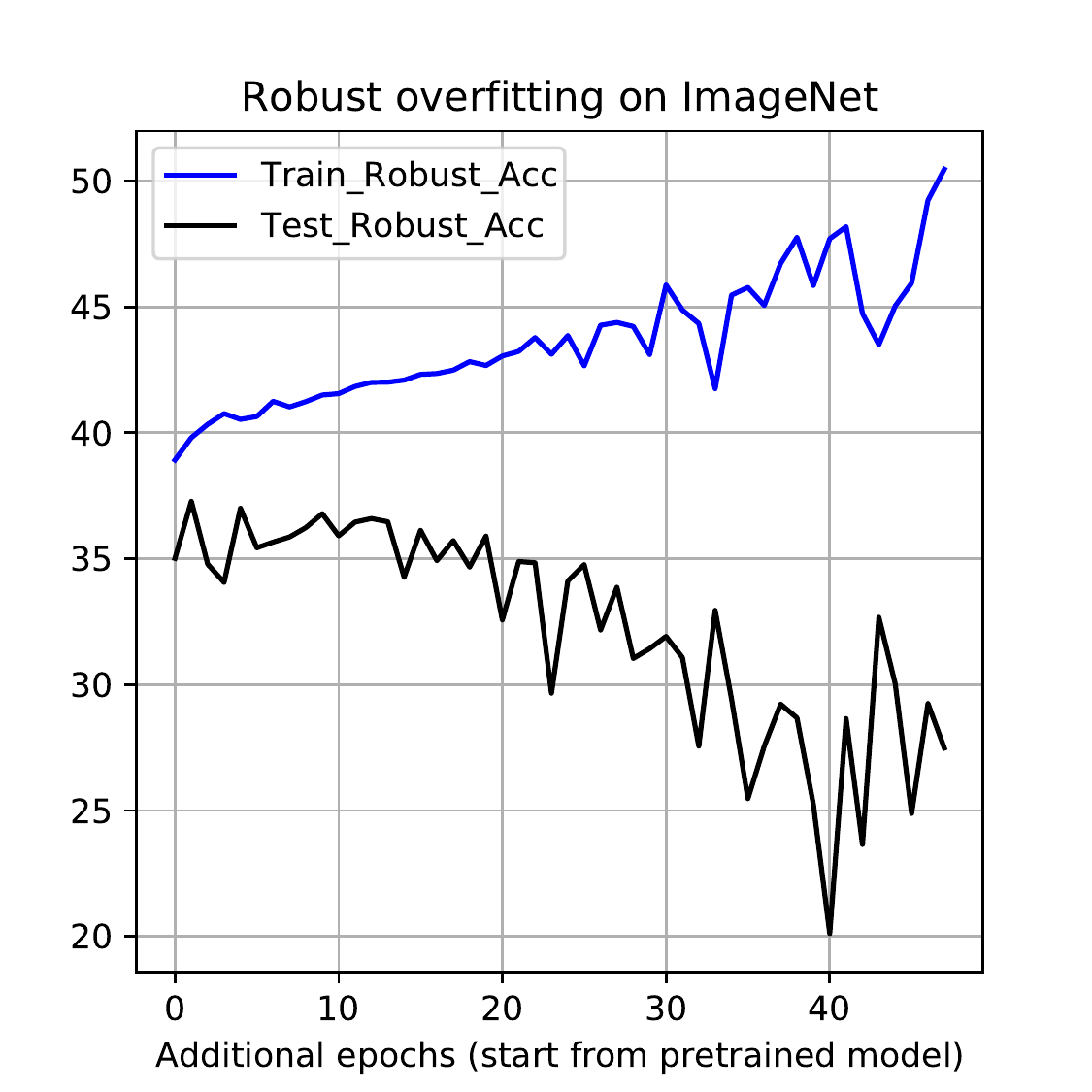}
			\end{minipage}
	}}
	\centering

	\caption{Robust overfitting in the experiments on (a) CIFAR-10, (b) CIFAR-10, (c) SVHN and (d) ImageNet.}
	\label{fig:over}
\end{figure*}

\paragraph{Robust Overfitting on Common Dataset.} In Fig. \ref{fig:over} (a), (b), and (c), we show the experiments on the piece-wise learning rate schedule, which is 0.1 over the first 100 epochs, down to 0.01 over the following 50 epochs, and finally be 0.001 in
the last 50 epochs, on CIFAR-10, CIFAR-100, and SVHN. Experiments on different $\epsilon$ are shown in Appendix \ref{c}. \emph{Robust Overfitting on ImageNet.} We provide one experiment on ImageNet in Fig. \ref{fig:over} (d). We start from a pre-trained model from Madry's Lab and keep running 50 more epochs. \emph{Robust overfitting} can be observed in these experiments. After a particular epoch (around the 100$^{th}$ epoch), the robust training accuracy is still increasing, but the robust test accuracy starts to decrease.

\begin{figure*}[htbp]
	\centering
	\hspace{-0.4in}\scalebox{0.9}{
		\subfigure[]{
			\begin{minipage}[htp]{0.26\linewidth}
				\centering
				\includegraphics[width=1.6in]{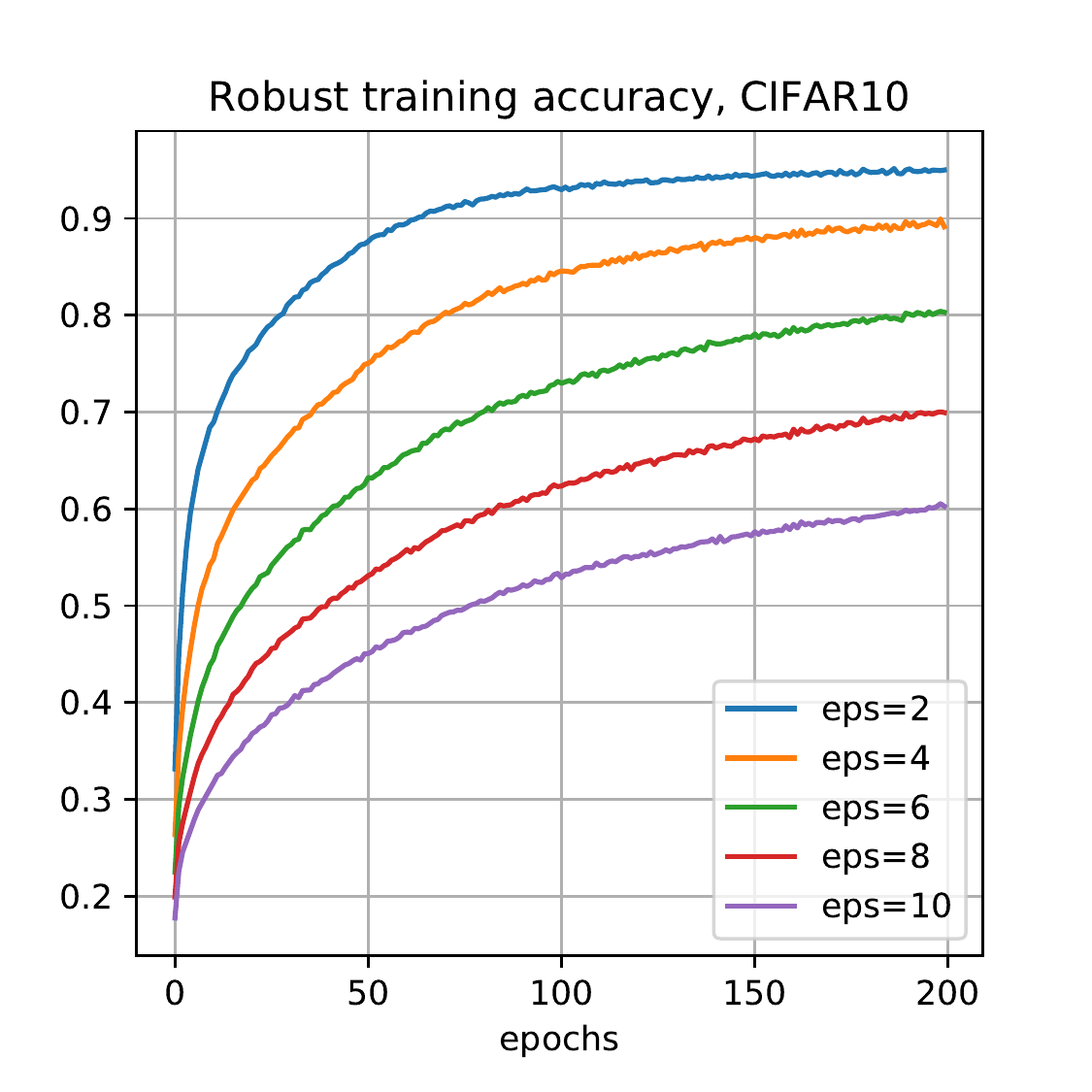}
			\end{minipage}%
		}
		\subfigure[]{
			\begin{minipage}[htp]{0.26\linewidth}
				\centering
				\includegraphics[width=1.6in]{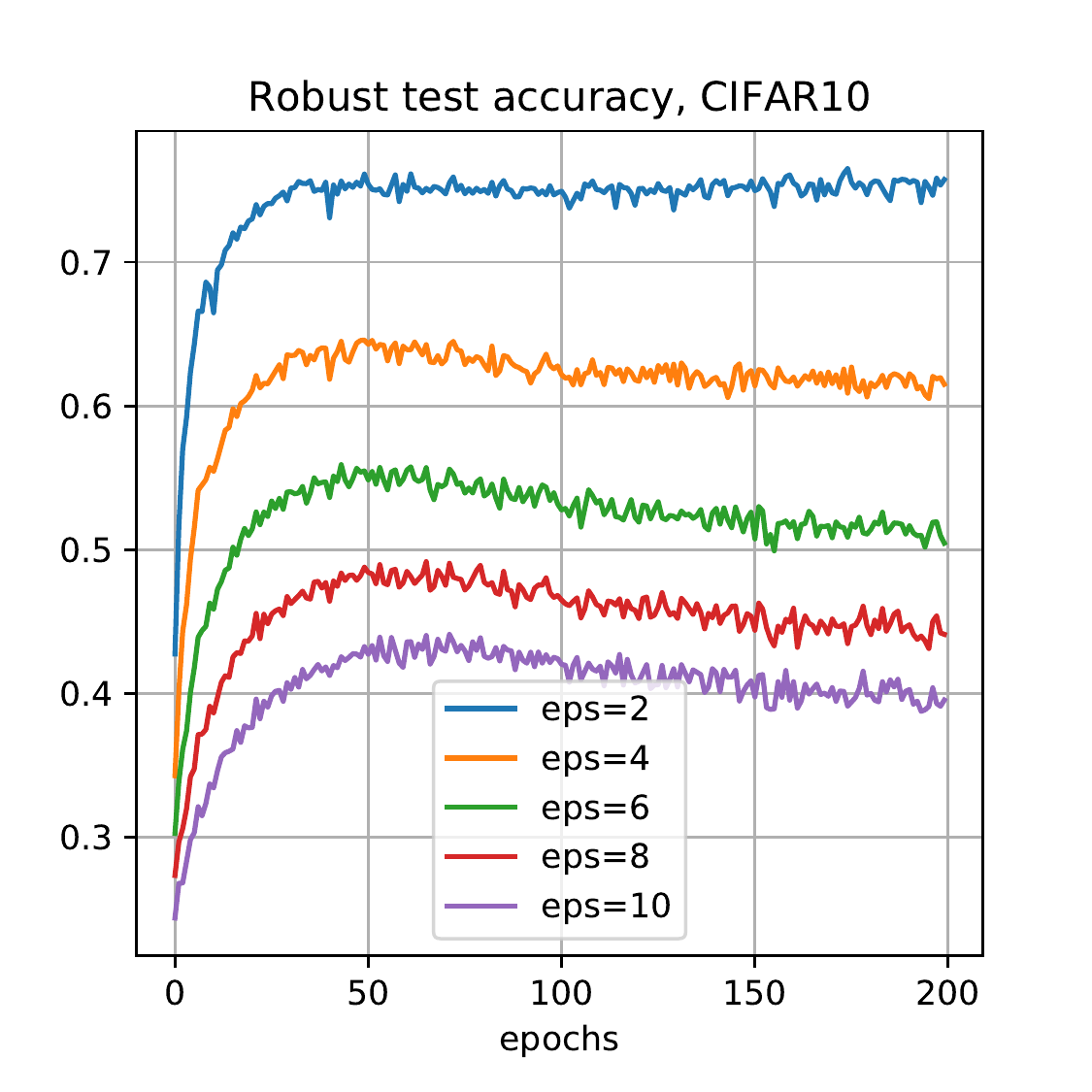}
			\end{minipage}
		}
		\subfigure[]{
			\begin{minipage}[htp]{0.26\linewidth}
				\centering
				\includegraphics[width=1.6in]{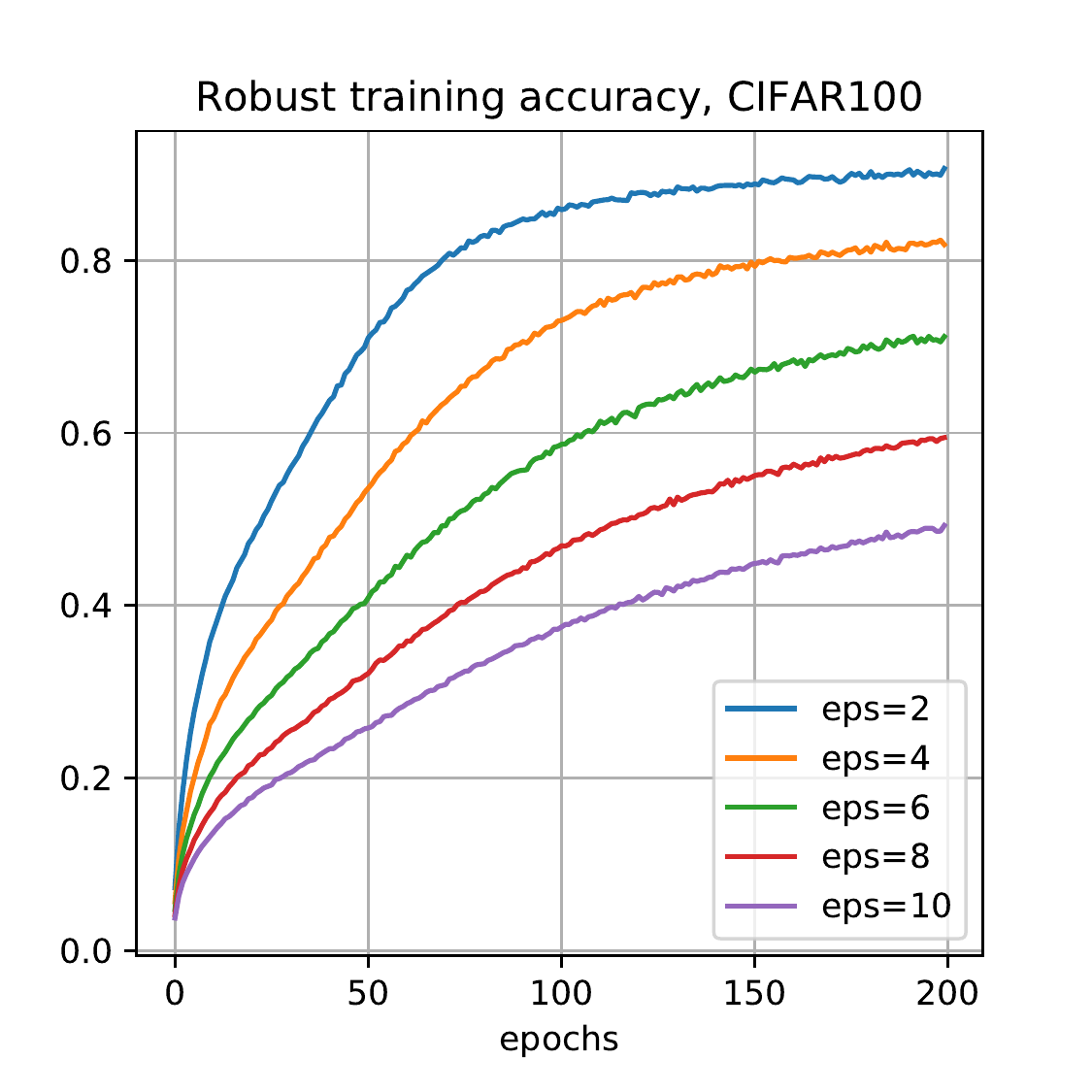}
			\end{minipage}
		}
		\subfigure[]{
			\begin{minipage}[htp]{0.26\linewidth}
				\centering
				\includegraphics[width=1.6in]{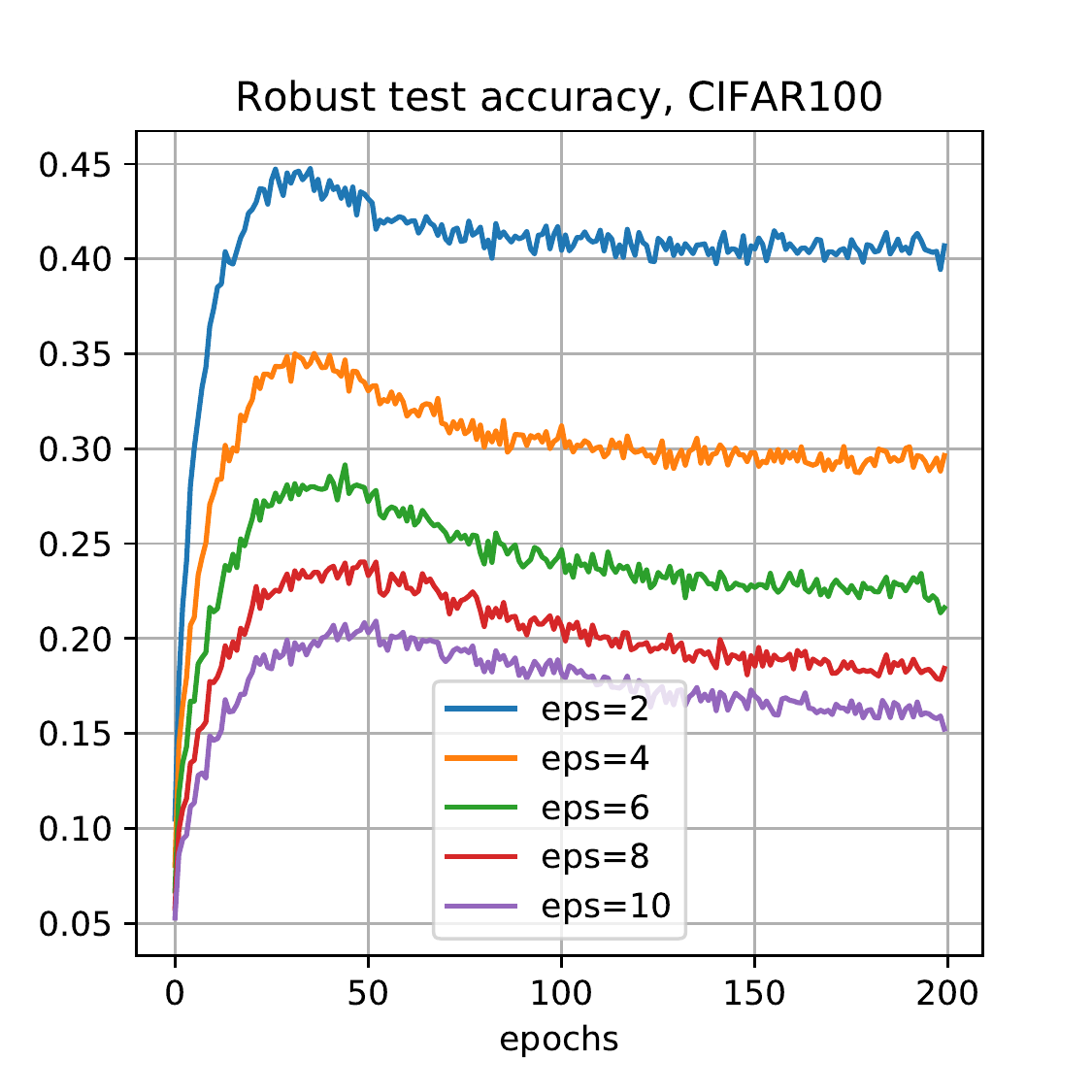}
			\end{minipage}
	}}
	\centering

	\caption{Experiments of adversarial training on CIFAR-10 and CIFAR-100 with a fixed learning rate. (a) Robust training accuracy on CIFAR-10. (b) Robust test accuracy on CIFAR-10. (c) Robust training accuracy on CIFAR-100. (d) Robust test accuracy on CIFAR-100. $\epsilon$ are set to be 2, 4, 6, 8, and 10.}
	\label{fig:fix1}
\end{figure*}

\begin{wrapfigure}{r}{0.5\textwidth}
	\centering
	\scalebox{0.9}{
		\subfigure[]{
			\begin{minipage}[htp]{0.5\linewidth}
				\centering
				\includegraphics[width=1.6in]{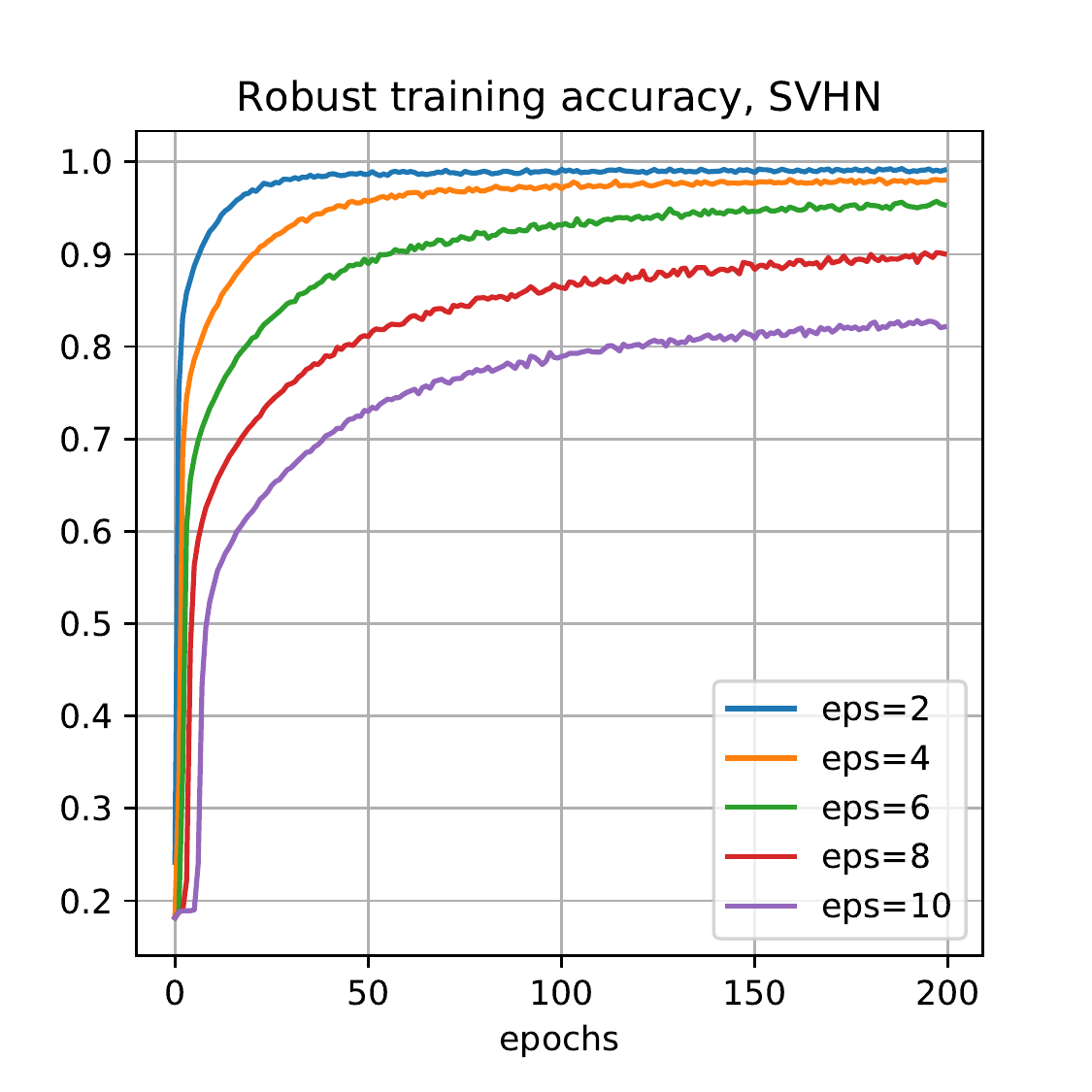}
			\end{minipage}%
		}
		\subfigure[]{
			\begin{minipage}[htp]{0.5\linewidth}
				\centering
				\includegraphics[width=1.6in]{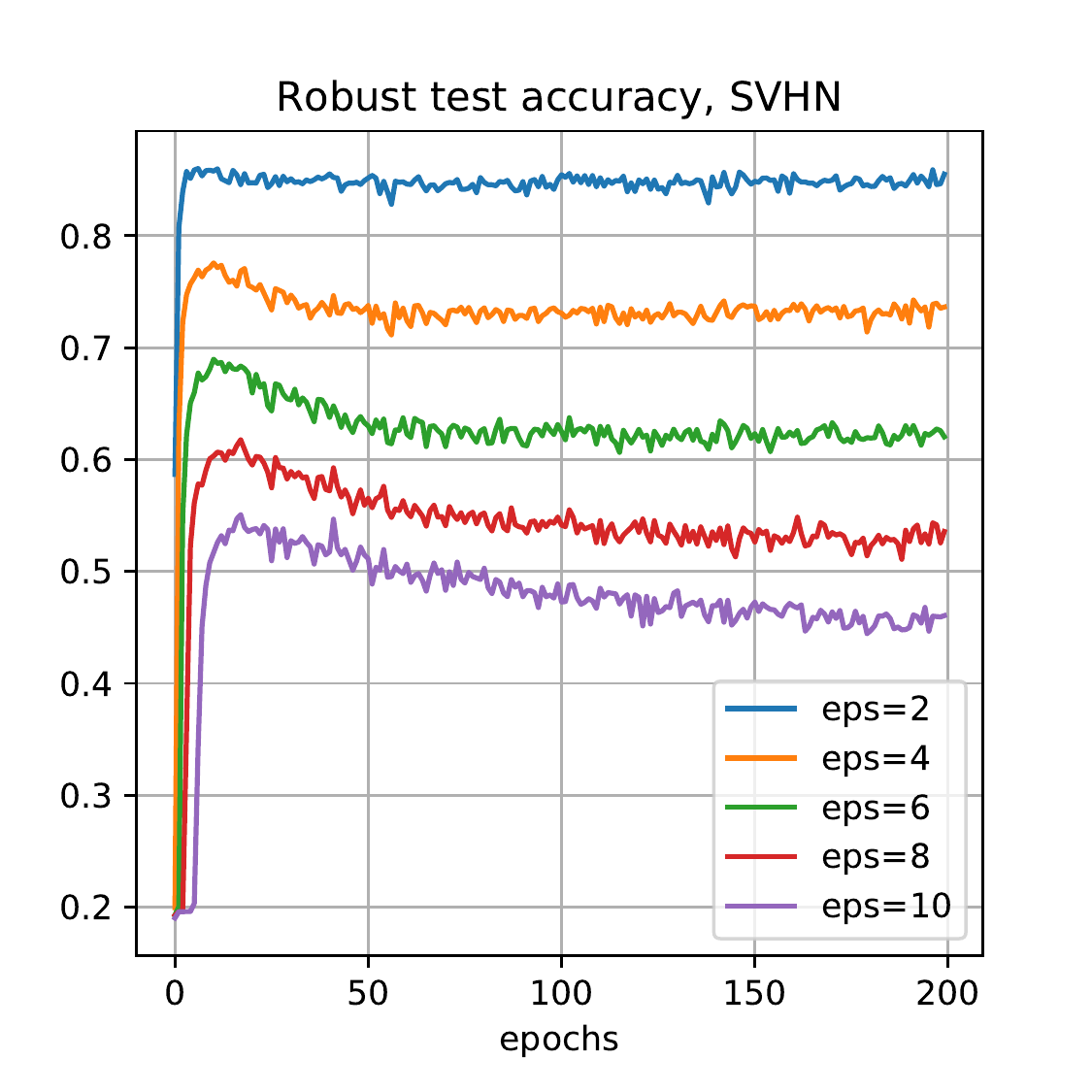}
			\end{minipage}
	}}
\vskip -0.15in
	\caption{Experiments of adversarial training on SVHN with a fixed learning rate. (a) Robust training accuracy. (b) Robust test accuracy.}
	\label{svhn}
\end{wrapfigure}

\paragraph{Fixed Step Size.} To better understand robust overfitting and match the theoretical settings (in Eq. (\ref{eq:tradeoff})), we consider the fixed learning rate schedule. In Fig. \ref{fig:fix1}, we show the experiments of adversarial training using a fixed learning rate $0.01$. The perturbation intensity $\epsilon$ is set to be 2, 4, 6, 8, and 10. respectively. Fig. \ref{fig:fix1} (a) and (b) show the experiments on CIFAR-10. Fig. \ref{fig:fix1} (c) and (d) show the experiments on CIFAR-100, respectively. Fig. \ref{svhn} shows the experiments on SVHN. 

\paragraph{Generalization Error Dominates Training Error.} In the robust overfitting phase, the robust generalization error dominates the robust training error. This phenomenon corresponds to the Eq. (\ref{eq:tradeoff}) that the first term dominates the other terms with large $T$.

\paragraph{Robust Test Accuracy Decreases in $\epsilon$.} Comparing different $\epsilon$ in Fig. \ref{fig:fix1}, we can see that the robust test accuracy decreases faster when $\epsilon$ is larger. This corresponds to the robust overfitting rule in the theoretical setting that the test performance decreases in $\epsilon$. These phenomena are similar in theoretical and practical settings. Therefore, the stability analysis provides a different perspective on understanding robust overfitting.

\paragraph{Robust Test Accuracy Decreases in a Rate between $\Omega(\epsilon \sqrt{T})$ and $\mathcal{O}({\epsilon T})$.} If $\epsilon$ is small, \emph{e.g.,} $\epsilon=2$, the decrease rate is close to $\mathcal{O}({\epsilon T})$. If $\epsilon$ is large, the decrease rate is more likely to be $\Omega(\epsilon \sqrt{T})$. This is also the gap between the upper bound and lower bound in Sec. \ref{s5}.

\section{Conclusion}
\paragraph{Limitations and Future Work.} Firstly, in Fig. \ref{fig:fix1}, we can see that the decrease rate of robust overfitting is close to $\mathcal{O}(T)$ when $\epsilon$ is small and is close to $\Omega(\sqrt{T})$ when $\epsilon$ is large. One possible direction is to figure out the relation. Secondly, one might improve adversarial training by controlling $L_z$. $L_z$ depends on the loss, the network architecture, and the dataset. One possible direction is to design a smoother loss or smooth activation function (e.g., SiLU) for a lower $L_z$. Notice that $L_z$ is uniform for all $\theta$. If we view $L_z(\theta)$ locally with respect to $\theta$, we might use an (approximated) second-order penalty term on its magnitude to control it.

In this paper, we show that the adversarial loss satisfies $\eta$-approximate smoothness, and we derive stability-based generalization bounds on this general class of $\eta$-approximate smooth functions. Our bounds give a different perspective on understanding robust overfitting. The robust test accuracy decreases in $\eta$, and experimental results confirm this phenomenon. We think our work will inspire more theoretical and empirical research to improve adversarial training.

\section*{Acknowledgement}
We would like to thank Jiawei Zhang, Congliang Chen, and Zeyu Qin for the helpful discussions. We thank all the anonymous reviewers for their comments and suggestions. The work of Zhi-Quan Luo is supported by the National Natural Science Foundation of China (No. 61731018) and the Guangdong Provincial Key Laboratory of Big Data Computing.

\clearpage
\bibliography{main.bib}

\begin{thebibliography}{}

\bibitem[Allen-Zhu et~al., 2019]{allen2019convergence}
Allen-Zhu, Z., Li, Y., and Song, Z. (2019).
\newblock A convergence theory for deep learning via over-parameterization.
\newblock In {\em International Conference on Machine Learning}, pages
  242--252. PMLR.

\bibitem[Athalye et~al., 2018]{athalye2018obfuscated}
Athalye, A., Carlini, N., and Wagner, D. (2018).
\newblock Obfuscated gradients give a false sense of security: Circumventing
  defenses to adversarial examples.
\newblock {\em arXiv preprint arXiv:1802.00420}.

\bibitem[Attias et~al., 2021]{attias2021improved}
Attias, I., Kontorovich, A., and Mansour, Y. (2021).
\newblock Improved generalization bounds for adversarially robust learning.

\bibitem[Awasthi et~al., 2020]{awasthi2020adversarial}
Awasthi, P., Frank, N., and Mohri, M. (2020).
\newblock Adversarial learning guarantees for linear hypotheses and neural
  networks.
\newblock In {\em International Conference on Machine Learning}, pages
  431--441. PMLR.

\bibitem[Bartlett and Mendelson, 2002]{bartlett2002rademacher}
Bartlett, P.~L. and Mendelson, S. (2002).
\newblock Rademacher and gaussian complexities: Risk bounds and structural
  results.
\newblock {\em Journal of Machine Learning Research}, 3(Nov):463--482.

\bibitem[Bassily et~al., 2020]{bassily2020stability}
Bassily, R., Feldman, V., Guzm{\'a}n, C., and Talwar, K. (2020).
\newblock Stability of stochastic gradient descent on nonsmooth convex losses.
\newblock {\em arXiv preprint arXiv:2006.06914}.

\bibitem[Bousquet and Elisseeff, 2002]{bousquet2002stability}
Bousquet, O. and Elisseeff, A. (2002).
\newblock Stability and generalization.
\newblock {\em The Journal of Machine Learning Research}, 2:499--526.

\bibitem[Carlini and Wagner, 2017]{carlini2017towards}
Carlini, N. and Wagner, D. (2017).
\newblock Towards evaluating the robustness of neural networks.
\newblock In {\em 2017 ieee symposium on security and privacy (sp)}, pages
  39--57. IEEE.

\bibitem[Carmon et~al., 2019]{carmon2019unlabeled}
Carmon, Y., Raghunathan, A., Schmidt, L., Duchi, J.~C., and Liang, P.~S.
  (2019).
\newblock Unlabeled data improves adversarial robustness.
\newblock In {\em Advances in Neural Information Processing Systems}, pages
  11190--11201.

\bibitem[Chen et~al., 2017]{chen2017zoo}
Chen, P.-Y., Zhang, H., Sharma, Y., Yi, J., and Hsieh, C.-J. (2017).
\newblock Zoo: Zeroth order optimization based black-box attacks to deep neural
  networks without training substitute models.
\newblock In {\em Proceedings of the 10th ACM Workshop on Artificial
  Intelligence and Security}, pages 15--26.

\bibitem[Chen et~al., 2018]{chen2018stability}
Chen, Y., Jin, C., and Yu, B. (2018).
\newblock Stability and convergence trade-off of iterative optimization
  algorithms.
\newblock {\em arXiv preprint arXiv:1804.01619}.

\bibitem[Dan et~al., 2020]{dan2020sharp}
Dan, C., Wei, Y., and Ravikumar, P. (2020).
\newblock Sharp statistical guaratees for adversarially robust gaussian
  classification.
\newblock In {\em International Conference on Machine Learning}, pages
  2345--2355. PMLR.

\bibitem[Deng et~al., 2009]{deng2009imagenet}
Deng, J., Dong, W., Socher, R., Li, L.-J., Li, K., and Fei-Fei, L. (2009).
\newblock Imagenet: A large-scale hierarchical image database.
\newblock In {\em 2009 IEEE conference on computer vision and pattern
  recognition}, pages 248--255. Ieee.

\bibitem[Du et~al., 2019]{du2019gradient}
Du, S., Lee, J., Li, H., Wang, L., and Zhai, X. (2019).
\newblock Gradient descent finds global minima of deep neural networks.
\newblock In {\em International Conference on Machine Learning}, pages
  1675--1685. PMLR.

\bibitem[Fan et~al., 2020]{fan2020sparse}
Fan, Y., Wu, B., Li, T., Zhang, Y., Li, M., Li, Z., and Yang, Y. (2020).
\newblock Sparse adversarial attack via perturbation factorization.
\newblock In {\em European conference on computer vision}, pages 35--50.
  Springer.

\bibitem[Farnia and Ozdaglar, 2021]{farnia2021train}
Farnia, F. and Ozdaglar, A. (2021).
\newblock Train simultaneously, generalize better: Stability of gradient-based
  minimax learners.
\newblock In {\em International Conference on Machine Learning}, pages
  3174--3185. PMLR.

\bibitem[Feldman and Vondrak, 2018]{feldman2018generalization}
Feldman, V. and Vondrak, J. (2018).
\newblock Generalization bounds for uniformly stable algorithms.
\newblock {\em arXiv preprint arXiv:1812.09859}.

\bibitem[Feldman and Vondrak, 2019]{feldman2019high}
Feldman, V. and Vondrak, J. (2019).
\newblock High probability generalization bounds for uniformly stable
  algorithms with nearly optimal rate.
\newblock In {\em Conference on Learning Theory}, pages 1270--1279. PMLR.

\bibitem[Goodfellow et~al., 2014]{goodfellow2014explaining}
Goodfellow, I.~J., Shlens, J., and Szegedy, C. (2014).
\newblock Explaining and harnessing adversarial examples.
\newblock {\em arXiv preprint arXiv:1412.6572}.

\bibitem[Gowal et~al., 2020]{gowal2020uncovering}
Gowal, S., Qin, C., Uesato, J., Mann, T., and Kohli, P. (2020).
\newblock Uncovering the limits of adversarial training against norm-bounded
  adversarial examples.
\newblock {\em arXiv preprint arXiv:2010.03593}.

\bibitem[Hardt et~al., 2016]{hardt2016train}
Hardt, M., Recht, B., and Singer, Y. (2016).
\newblock Train faster, generalize better: Stability of stochastic gradient
  descent.
\newblock In {\em International Conference on Machine Learning}, pages
  1225--1234. PMLR.

\bibitem[He et~al., 2016]{he2016deep}
He, K., Zhang, X., Ren, S., and Sun, J. (2016).
\newblock Deep residual learning for image recognition.
\newblock In {\em Proceedings of the IEEE conference on computer vision and
  pattern recognition}, pages 770--778.

\bibitem[Hochreiter and Schmidhuber, 1997]{hochreiter1997long}
Hochreiter, S. and Schmidhuber, J. (1997).
\newblock Long short-term memory.
\newblock {\em Neural computation}, 9(8):1735--1780.

\bibitem[Huang et~al., 2022]{huang2022fast}
Huang, Z., Fan, Y., Liu, C., Zhang, W., Zhang, Y., Salzmann, M., S{\"u}sstrunk,
  S., and Wang, J. (2022).
\newblock Fast adversarial training with adaptive step size.
\newblock {\em arXiv preprint arXiv:2206.02417}.

\bibitem[Hwang et~al., 2021]{hwang2021adversarial}
Hwang, J.-W., Lee, Y., Oh, S., and Bae, Y. (2021).
\newblock Adversarial training with stochastic weight average.
\newblock In {\em 2021 IEEE International Conference on Image Processing
  (ICIP)}, pages 814--818. IEEE.

\bibitem[Izmailov et~al., 2018]{izmailov2018averaging}
Izmailov, P., Podoprikhin, D., Garipov, T., Vetrov, D., and Wilson, A.~G.
  (2018).
\newblock Averaging weights leads to wider optima and better generalization.
\newblock {\em arXiv preprint arXiv:1803.05407}.

\bibitem[Javanmard et~al., 2020]{javanmard2020precise}
Javanmard, A., Soltanolkotabi, M., and Hassani, H. (2020).
\newblock Precise tradeoffs in adversarial training for linear regression.
\newblock In {\em Conference on Learning Theory}, pages 2034--2078. PMLR.

\bibitem[Khim and Loh, 2018]{khim2018adversarial}
Khim, J. and Loh, P.-L. (2018).
\newblock Adversarial risk bounds via function transformation.
\newblock {\em arXiv preprint arXiv:1810.09519}.

\bibitem[Krizhevsky et~al., 2009]{krizhevsky2009learning}
Krizhevsky, A., Hinton, G., et~al. (2009).
\newblock Learning multiple layers of features from tiny images.

\bibitem[Krizhevsky et~al., 2012]{krizhevsky2012imagenet}
Krizhevsky, A., Sutskever, I., and Hinton, G.~E. (2012).
\newblock Imagenet classification with deep convolutional neural networks.
\newblock In {\em Advances in neural information processing systems}, pages
  1097--1105.

\bibitem[Li et~al., 2022]{li2022semi}
Li, Y., Wu, B., Feng, Y., Fan, Y., Jiang, Y., Li, Z., and Xia, S.-T. (2022).
\newblock Semi-supervised robust training with generalized perturbed
  neighborhood.
\newblock {\em Pattern Recognition}, 124:108472.

\bibitem[Liu et~al., 2020]{liu2020loss}
Liu, C., Salzmann, M., Lin, T., Tomioka, R., and S{\"u}sstrunk, S. (2020).
\newblock On the loss landscape of adversarial training: Identifying challenges
  and how to overcome them.
\newblock {\em arXiv preprint arXiv:2006.08403}.

\bibitem[Madry et~al., 2017]{madry2017towards}
Madry, A., Makelov, A., Schmidt, L., Tsipras, D., and Vladu, A. (2017).
\newblock Towards deep learning models resistant to adversarial attacks.
\newblock {\em arXiv preprint arXiv:1706.06083}.

\bibitem[Montasser et~al., 2019]{montasser2019vc}
Montasser, O., Hanneke, S., and Srebro, N. (2019).
\newblock Vc classes are adversarially robustly learnable, but only improperly.
\newblock In {\em Conference on Learning Theory}, pages 2512--2530. PMLR.

\bibitem[Nemirovski et~al., 2009]{nemirovski2009robust}
Nemirovski, A., Juditsky, A., Lan, G., and Shapiro, A. (2009).
\newblock Robust stochastic approximation approach to stochastic programming.
\newblock {\em SIAM Journal on optimization}, 19(4):1574--1609.

\bibitem[Netzer et~al., 2011]{netzer2011reading}
Netzer, Y., Wang, T., Coates, A., Bissacco, A., Wu, B., and Ng, A.~Y. (2011).
\newblock Reading digits in natural images with unsupervised feature learning.

\bibitem[Ozdaglar et~al., 2022]{ozdaglar2022good}
Ozdaglar, A., Pattathil, S., Zhang, J., and Zhang, K. (2022).
\newblock What is a good metric to study generalization of minimax learners?
\newblock {\em arXiv preprint arXiv:2206.04502}.

\bibitem[Qin et~al., 2022]{qin2022boosting}
Qin, Z., Fan, Y., Liu, Y., Zhang, Y., Wang, J., and Wu, B. (2022).
\newblock Boosting the transferability of adversarial attacks with reverse
  adversarial perturbation.

\bibitem[Qin et~al., 2021]{qin2021random}
Qin, Z., Fan, Y., Zha, H., and Wu, B. (2021).
\newblock Random noise defense against query-based black-box attacks.
\newblock {\em Advances in Neural Information Processing Systems},
  34:7650--7663.

\bibitem[Raghunathan et~al., 2019]{raghunathan2019adversarial}
Raghunathan, A., Xie, S.~M., Yang, F., Duchi, J.~C., and Liang, P. (2019).
\newblock Adversarial training can hurt generalization.
\newblock {\em arXiv preprint arXiv:1906.06032}.

\bibitem[Rebuffi et~al., 2021]{rebuffi2021fixing}
Rebuffi, S.-A., Gowal, S., Calian, D.~A., Stimberg, F., Wiles, O., and Mann, T.
  (2021).
\newblock Fixing data augmentation to improve adversarial robustness.
\newblock {\em arXiv preprint arXiv:2103.01946}.

\bibitem[Rice et~al., 2020]{rice2020overfitting}
Rice, L., Wong, E., and Kolter, Z. (2020).
\newblock Overfitting in adversarially robust deep learning.
\newblock In {\em International Conference on Machine Learning}, pages
  8093--8104. PMLR.

\bibitem[Rogers and Wagner, 1978]{rogers1978finite}
Rogers, W.~H. and Wagner, T.~J. (1978).
\newblock A finite sample distribution-free performance bound for local
  discrimination rules.
\newblock {\em The Annals of Statistics}, pages 506--514.

\bibitem[Schmidt et~al., 2018]{schmidt2018adversarially}
Schmidt, L., Santurkar, S., Tsipras, D., Talwar, K., and Madry, A. (2018).
\newblock Adversarially robust generalization requires more data.
\newblock In {\em Advances in Neural Information Processing Systems}, pages
  5014--5026.

\bibitem[Sinha et~al., 2017]{sinha2017certifiable}
Sinha, A., Namkoong, H., and Duchi, J. (2017).
\newblock Certifiable distributional robustness with principled adversarial
  training.
\newblock {\em arXiv preprint arXiv:1710.10571}, 2.

\bibitem[Szegedy et~al., 2013]{szegedy2013intriguing}
Szegedy, C., Zaremba, W., Sutskever, I., Bruna, J., Erhan, D., Goodfellow, I.,
  and Fergus, R. (2013).
\newblock Intriguing properties of neural networks.
\newblock {\em arXiv preprint arXiv:1312.6199}.

\bibitem[Taheri et~al., 2020]{taheri2020asymptotic}
Taheri, H., Pedarsani, R., and Thrampoulidis, C. (2020).
\newblock Asymptotic behavior of adversarial training in binary classification.
\newblock {\em arXiv preprint arXiv:2010.13275}.

\bibitem[Tramer et~al., 2020]{tramer2020adaptive}
Tramer, F., Carlini, N., Brendel, W., and Madry, A. (2020).
\newblock On adaptive attacks to adversarial example defenses.
\newblock {\em arXiv preprint arXiv:2002.08347}.

\bibitem[Vapnik and Chervonenkis, 2015]{vapnik2015uniform}
Vapnik, V.~N. and Chervonenkis, A.~Y. (2015).
\newblock On the uniform convergence of relative frequencies of events to their
  probabilities.
\newblock In {\em Measures of complexity}, pages 11--30. Springer.

\bibitem[Wang et~al., 2019]{wang2019convergence}
Wang, Y., Ma, X., Bailey, J., Yi, J., Zhou, B., and Gu, Q. (2019).
\newblock On the convergence and robustness of adversarial training.
\newblock In {\em ICML}, volume~1, page~2.

\bibitem[Wong et~al., 2020]{wong2020fast}
Wong, E., Rice, L., and Kolter, J.~Z. (2020).
\newblock Fast is better than free: Revisiting adversarial training.
\newblock {\em arXiv preprint arXiv:2001.03994}.

\bibitem[Wu et~al., 2020]{wu2020adversarial}
Wu, D., Xia, S.-T., and Wang, Y. (2020).
\newblock Adversarial weight perturbation helps robust generalization.
\newblock {\em arXiv preprint arXiv:2004.05884}.

\bibitem[Xiao et~al., 2022a]{xiao2022adversarial}
Xiao, J., Fan, Y., Sun, R., and Luo, Z.-Q. (2022a).
\newblock Adversarial rademacher complexity of deep neural networks.

\bibitem[Xiao et~al., 2022b]{xiao2022adaptive}
Xiao, J., Qin, Z., Fan, Y., Wu, B., Wang, J., and Luo, Z.-Q. (2022b).
\newblock Adaptive smoothness-weighted adversarial training for multiple
  perturbations with its stability analysis.
\newblock {\em arXiv preprint arXiv:2210.00557}.

\bibitem[Xiao et~al., 2022c]{xiao2022understanding}
Xiao, J., Yang, L., Fan, Y., Wang, J., and Luo, Z.-Q. (2022c).
\newblock Understanding adversarial robustness against on-manifold adversarial
  examples.
\newblock {\em arXiv preprint arXiv:2210.00430}.

\bibitem[Xing et~al., 2021a]{xing2021on}
Xing, Y., Song, Q., and Cheng, G. (2021a).
\newblock On the algorithmic stability of adversarial training.
\newblock In {\em Thirty-Fifth Conference on Neural Information Processing
  Systems}.

\bibitem[Xing et~al., 2021b]{xing2021generalization}
Xing, Y., Song, Q., and Cheng, G. (2021b).
\newblock On the generalization properties of adversarial training.
\newblock In {\em International Conference on Artificial Intelligence and
  Statistics}, pages 505--513. PMLR.

\bibitem[Xing et~al., 2021c]{xing2021adversarially}
Xing, Y., Zhang, R., and Cheng, G. (2021c).
\newblock Adversarially robust estimate and risk analysis in linear regression.
\newblock In {\em International Conference on Artificial Intelligence and
  Statistics}, pages 514--522. PMLR.

\bibitem[Yin et~al., 2019]{yin2019rademacher}
Yin, D., Kannan, R., and Bartlett, P. (2019).
\newblock Rademacher complexity for adversarially robust generalization.
\newblock In {\em International Conference on Machine Learning}, pages
  7085--7094. PMLR.

\bibitem[Zhai et~al., 2019]{zhai2019adversarially}
Zhai, R., Cai, T., He, D., Dan, C., He, K., Hopcroft, J., and Wang, L. (2019).
\newblock Adversarially robust generalization just requires more unlabeled
  data.
\newblock {\em arXiv preprint arXiv:1906.00555}.

\bibitem[Zhang et~al., 2021]{zhang2021understanding}
Zhang, C., Bengio, S., Hardt, M., Recht, B., and Vinyals, O. (2021).
\newblock Understanding deep learning (still) requires rethinking
  generalization.
\newblock {\em Communications of the ACM}, 64(3):107--115.

\bibitem[Zhang et~al., 2020]{zhang2020single}
Zhang, J., Xiao, P., Sun, R., and Luo, Z. (2020).
\newblock A single-loop smoothed gradient descent-ascent algorithm for
  nonconvex-concave min-max problems.
\newblock {\em Advances in Neural Information Processing Systems},
  33:7377--7389.

\end{thebibliography}
\bibliographystyle{apalike}
\clearpage

\begin{enumerate}
	\item For all authors...
	\begin{enumerate}
		\item Do the main claims made in the abstract and introduction accurately reflect the paper's contributions and scope?
		\answerYes{}
		\item Did you describe the limitations of your work?
		\answerYes{}
		\item Did you discuss any potential negative societal impacts of your work?
		\answerYes{}
		\item Have you read the ethics review guidelines and ensured that your paper conforms to them?
		\answerYes{}
	\end{enumerate}

	\item If you are including theoretical results...
	\begin{enumerate}
		\item Did you state the full set of assumptions of all theoretical results?
		\answerYes{}
		\item Did you include complete proofs of all theoretical results?
		\answerYes{}
	\end{enumerate}

	\item If you ran experiments...
	\begin{enumerate}
		\item Did you include the code, data, and instructions needed to reproduce the main experimental results (either in the supplemental material or as a URL)?
		\answerNA{}
		\item Did you specify all the training details (e.g., data splits, hyperparameters, how they were chosen)?
		\answerYes{}
		\item Did you report error bars (e.g., with respect to the random seed after running experiments multiple times)?
		\answerNo{}
		\item Did you include the total amount of compute and the type of resources used (e.g., type of GPUs, internal cluster, or cloud provider)?
		\answerYes{}
	\end{enumerate}

	\item If you are using existing assets (e.g., code, data, models) or curating/releasing new assets...
	\begin{enumerate}
		\item If your work uses existing assets, did you cite the creators?
		\answerNA{}
		\item Did you mention the license of the assets?
		\answerNA{}
		\item Did you include any new assets either in the supplemental material or as a URL?
		\answerNA{}
		\item Did you discuss whether and how consent was obtained from people whose data you're using/curating?
		\answerNA{}
		\item Did you discuss whether the data you are using/curating contains personally identifiable information or offensive content?
		\answerNA{}
	\end{enumerate}

	\item If you used crowdsourcing or conducted research with human subjects...
	\begin{enumerate}
		\item Did you include the full text of instructions given to participants and screenshots, if applicable?
		\answerNA{}
		\item Did you describe any potential participant risks, with links to Institutional Review Board (IRB) approvals, if applicable?
		\answerNA{}
		\item Did you include the estimated hourly wage paid to participants and the total amount spent on participant compensation?
		\answerNA{}
	\end{enumerate}
\end{enumerate}
\clearpage
\appendix
\section{Proof of the Theorem}
\label{a}
\subsection{Proof of Theorem \ref{thm:stab2gen}}
The proof can be found in \citep{hardt2016train}. We provide the proof in Appendix for reference. Denote by $S=(z_1,\dots,z_n)$ and $S'=(z_1',\dots,z_n')$ two independent
random samples and let $S^{(i)}=(z_1,\dots,z_{i-1},z_i',z_{i+1},\dots,z_n)$ be
the sample that is identical to $S$ except in the $i$'th example where we
replace $z_i$ with $z_i'$. With this notation, we get that
\begin{align*}
\E_S\E_A \left[ R_S[A(S)]\right]
& = \E_S\E_A \left[ \frac 1n\sum_{i=1}^n h(A(S);z_i) \right]\\
& = \E_S\E_{S'}\E_A\left[\frac 1n\sum_{i=1}^n h(A(S^{(i)});z_i')\right] \\
& = \E_S\E_{S'}\E_A\left[\frac 1n\sum_{i=1}^n h(A(S);z_i')\right] + \delta \\
& = \E_S\E_A\left[ R_{\mathcal{D}}[A(S)]\right] + \delta,
\end{align*}
where we can express $\delta$ as
\[
\delta
=
\E_S\E_{S'}\E_A\left[\frac 1n\sum_{i=1}^n h(A(S^{(i)});z_i')
- \frac 1n\sum_{i=1}^n h(A(S);z_i')\right]\,.
\]
Furthermore, taking the supremum over any two data sets $S,S'$ differing in only
one sample, we can bound the difference as
\[
\left|\delta\right| \le \sup_{S,S',z}\E_A\left[ h(A(S);z) - h(A(S');z)\right] \le \epsilon,
\]
by our assumption on the uniform stability of $A.$ The claim follows.\qed
\subsection{Proof of Lemma \ref{lem:nonsmooth}}
Let the adversarial examples for parameter $\theta_1$ and $\theta_2$ be
\[z_1\in\arg\max_{\|z-z'\|_p\leq\epsilon} g(\theta_1,z')\] \[z_2\in\arg\max_{\|z-z'\|_p\leq\epsilon} g(\theta_2,z'),\]
then we have
\begin{equation*}
\begin{aligned}
&\|h(\theta_1,z)-h(\theta_2,z)\|\\
=&|g(\theta_1,z_1)-g(\theta_2,z_2)|\\
\leq &\max\{|g(\theta_1,z_1)-g(\theta_2,z_1)|,|g(\theta_1,z_2)-g(\theta_2,z_2)|\}\\
\leq &L\|\theta_1-\theta_2\|,
\end{aligned}
\end{equation*}
where the first inequality is based on the fact that $g(\theta_1,z_1)\geq g(\theta_1,z_2)$ and $g(\theta_2,z_2)\geq g(\theta_2,z_1)$, the second inequality is based on Assumption \ref{ass1}. This proves Lemma \ref{lem:1}. 

For all subgradient $d(\theta,z)\in\partial_\theta h(\theta,z)$, we have
\begin{equation*}
\begin{aligned}
&\|d(\theta_1,z)-d(\theta_2,z)\|\\
=&\|\nabla_\theta g(\theta_1,z_1)-\nabla_\theta g(\theta_2,z_2)\|\\
\leq&\|\nabla_\theta g(\theta_1,z_1)-\nabla_\theta g(\theta_2,z_1)\|+\|\nabla_\theta g(\theta_2,z_1)-\nabla_\theta g(\theta_2,z_2)\|\\
\leq &  L_{\theta}\|\theta_1-\theta_2\|+L_z\|z_1-z_2\|_p\\
\leq &  L_{\theta}\|\theta_1-\theta_2\|+L_z[\|z_1-z\|_p+\|z-z_2\|_p]\\
\leq &  L_\theta\|\theta_1-\theta_2\|+2L_z\epsilon
\end{aligned}
\end{equation*}
where the first and the third inequality is due to triangle inequality, the second inequality is based on Assumption \ref{ass1}. This proves the second inequality (non-gradient Lipschitz) in Lemma \ref{lem:nonsmooth}.\qed

\subsection{Proof of Lemma \ref{lem:cocoercive}}
Proof of Lemma \ref{lem:3} ($\eta$-approximate descent Lemma).\\
Let $\tilde{\theta}$ be a point in the line segment of $\theta_1$ and $\theta_2$, $\tilde{\theta}(u)= \theta_2+u(\theta_1-\theta_2)$, then
\begin{equation*}
\begin{aligned}
   &h(\theta_1)-h(\theta_2)\\
   =&\int_0^1 \langle \theta_1-\theta_2, \nabla_\theta h(\tilde{\theta}(u))\rangle d u\\
      =&\int_0^1 \langle \theta_1-\theta_2, \nabla_\theta h(\theta_2)+\nabla_\theta h(\tilde{\theta}(u))-\nabla_\theta h(\theta_2)\rangle d u\\
      =&\langle \nabla_\theta h(\theta_2),\theta_1-\theta_2\rangle+\int_0^1 \langle \theta_1-\theta_2,\nabla_\theta h(\tilde{\theta}(u))-\nabla_\theta h(\theta_2)\rangle d u\\
      \leq&\langle \nabla_\theta h(\theta_2),\theta_1-\theta_2\rangle+\int_0^1 \|\theta_1-\theta_2\|\|\nabla_\theta h(\tilde{\theta}(u))-\nabla_\theta h(\theta_2)\| d u\\
      \leq&\langle \nabla_\theta h(\theta_2),\theta_1-\theta_2\rangle+\int_0^1 \|\theta_1-\theta_2\|[\beta\|\tilde{\theta}(u)-\theta_2\|+\eta] d u\\
      =&\langle \nabla_\theta h(\theta_2),\theta_1-\theta_2\rangle+\int_0^1 \|\theta_1-\theta_2\|[\beta u \|\theta_1-\theta_2\|+\eta] d u\\
      =&\langle \nabla_\theta h(\theta_2),\theta_1-\theta_2\rangle+\beta\|\theta_1-\theta_2\|^2\int_0^1 u d u+\eta\|\theta_1-\theta_2\|]\\
      =&\langle \nabla_\theta h(\theta_2),\theta_1-\theta_2\rangle +\frac{\beta}{2}\|\theta_1-\theta_2\|^2+\eta\|\theta_1-\theta_2\|.
\end{aligned}
\end{equation*}\qed\\
Proof of Lemma \ref{lem:4} ($\eta$-approximate co-coercive).\\
By Lemma \ref{lem:3} ($\eta$-approximate descent Lemma), we have

\begin{equation*}
\begin{aligned}
   h(\theta_1)\leq h(\theta_2)
      +\langle \nabla_\theta h(\theta_2),\theta_1-\theta_2\rangle +\frac{\beta}{2}\|\theta_1-\theta_2\|^2+\eta\|\theta_1-\theta_2\|.
\end{aligned}
\end{equation*}
Let $\theta^*$ be a minimizer of $h$, then
\begin{equation*}
\begin{aligned}
   h(\theta^*)=\inf_{\theta_1}h(\theta_1)&\leq \inf_{\theta_1}\bigg(h(\theta_2)
      +\langle \nabla_\theta h(\theta_2),\theta_1-\theta_2\rangle +\frac{\beta}{2}\|\theta_1-\theta_2\|^2+\eta\|\theta_1-\theta_2\|\bigg)\\
      &=\inf_{\|v\|=1}\inf_{t\geq 0}\bigg(h(\theta_2)
      +t\nabla_\theta h(\theta_2)^Tv +\frac{\beta t^2}{2}+\eta t\bigg),
\end{aligned}
\end{equation*}
where $t=\|\theta_1-\theta_2\|$ and $v=(\theta_1-\theta_2)/\|\theta_1-\theta_2\|$. Then
\begin{equation*}
\begin{aligned}
&\inf_{\|v\|=1}\inf_{t\geq 0}\bigg(h(\theta_2)
      +t\nabla_\theta h(\theta_2)^Tv +\frac{\beta t^2}{2}+\eta t\bigg)\\
      =&\inf_{t\geq 0}\bigg(h(\theta_2)
      -t\|\nabla_\theta h(\theta_2)\| +\frac{\beta t^2}{2}+\eta t\bigg)\\
            =&h(\theta_2)+\inf_{t\geq 0}\bigg(
      -t(\|\nabla_\theta h(\theta_2)\|-\eta) +\frac{\beta t^2}{2}\bigg).
\end{aligned}
\end{equation*}
If $\|\nabla_\theta h(\theta_2)\|-\eta\leq 0$, the quaduatic function is optimized when $t=0$. Then
\begin{equation*}
\begin{aligned}
            &h(\theta_2)+\inf_{t\geq 0}\bigg(
      -t(\|\nabla_\theta h(\theta_2)\|-\eta) +\frac{\beta t^2}{2}\bigg)\\
      =& h(\theta_2).
\end{aligned}
\end{equation*}
If $\|\nabla_\theta h(\theta_2)\|-\eta\geq 0$, then
\begin{equation*}
\begin{aligned}
            &h(\theta_2)+\inf_{t\geq 0}\bigg(
      -t(\|\nabla_\theta h(\theta_2)\|-\eta) +\frac{\beta t^2}{2}\bigg)\\
      =& h(\theta_2)-\frac{1}{2\beta}[\|\nabla_\theta h(\theta_2)\|-\eta]^2.
\end{aligned}
\end{equation*}
Therefore, we obtain that
\begin{equation}
\label{eq:temp1}
\begin{aligned}
    h(\theta^*)-h(\theta) \leq -\frac{1}{2 \beta}\bigg[[\|\nabla h(\theta)\|-\eta]_+\bigg]^2.
\end{aligned}
\end{equation}

Define 
$$h_1(w)=h(w)-\nabla h(\theta_1)^T w$$
and 
$$h_2(w)=h(w)-\nabla h(\theta_2)^T w.$$ 
Firstly, it is easy to see that $h_1(w)$ and $h_2(w)$ are both $\eta$-approximate $\beta$-gradient Lipschitz, which satisfies inequatily in Eq. (\ref{eq:temp1}). Secondly, $w=\theta_1$ minimizes $h_1(w)$. Then 
\begin{equation}
\label{eq:temp2}
\begin{aligned}
&h(\theta_2)-h(\theta_1)-\nabla h(\theta_1)^T(\theta_2-\theta_1)\\=&h_1(\theta_2)-h_1 (\theta_1)\\
\geq &\frac{1}{2\beta}\bigg[[\|\nabla h_1(\theta_2)\|-\eta]_+\bigg]^2\\
 =&\frac{1}{2\beta}\bigg[[\|\nabla h(\theta_1)-\nabla h(\theta_2)\|-\eta]_+\bigg]^2.
\end{aligned}
\end{equation}
Similarly, we have
\begin{equation}
\label{eq:temp3}
\begin{aligned}
h(\theta_1)-h(\theta_2)-\nabla h(\theta_2)^T(\theta_1-\theta_2)\geq\frac{1}{2\beta}\bigg[[\|\nabla h(\theta_1)-\nabla h(\theta_2)\|-\eta]_+\bigg]^2.
\end{aligned}
\end{equation}

Take the summation of Eq. (\ref{eq:temp2}) and Eq. (\ref{eq:temp3}), we have
\begin{equation*}
\begin{aligned}
    \langle\nabla h(\theta_1)-\nabla h(\theta_2),\theta_1-\theta_2\rangle\geq  \frac{1}{\beta}\bigg[[\|\nabla h(\theta_1)-\nabla h(\theta_2)\|-\eta]_+\bigg]^2.
\end{aligned}
\end{equation*}\qed
\subsection{Proof of Lemma \ref{updaterules}}
Proof of Lemma \ref{lem:5} ($\alpha\eta$-approximately $(1+\alpha\beta)$-expansive).\\
\begin{equation*}
\begin{aligned}
    &\|G_{\alpha,z}(\theta_1)-G_{\alpha,z}(\theta_2)\|\\
    =&\|\theta_1-\theta_2-\alpha(\nabla h(\theta_1)-\nabla h(\theta_2))\|\\
    \leq&\|\theta_1-\theta_2\|+\|\alpha(\nabla h(\theta_1)-\nabla h(\theta_2))\|\\
        \leq&\|\theta_1-\theta_2\|+\alpha(\beta\|\theta_1-\theta_2\|+\eta)\\
    \leq& (1+\alpha\beta)\|\theta_1-\theta_2\|+\alpha\eta.\\
\end{aligned}
\end{equation*}\qed\\
Proof of Lemma \ref{lem:6} ($\alpha\eta$-approximately non-expansive.) Let t = $\|\nabla h(\theta_1)-\nabla h(\theta_2)\|$.\\
 If $t\leq\eta$, we have\\
\begin{equation*}
\begin{aligned}
    &\|G_{\alpha,z}(\theta_1)-G_{\alpha,z}(\theta_2)\|\\
    =&\|\theta_1-\theta_2-\alpha(\nabla h(\theta_1)-\nabla h(\theta_2))\|\\
    \leq&\|\theta_1-\theta_2\|+\alpha \|\nabla h(\theta_1)-\nabla h(\theta_2)\|\\
   \leq&\|\theta_1-\theta_2\|+\alpha \eta.\\
\end{aligned}
\end{equation*}
If $t\geq\eta$, we have\\
\begin{equation*}
\begin{aligned}
    &\|G_{\alpha,z}(\theta_1)-G_{\alpha,z}(\theta_2)\|^2\\
    =&\|\theta_1-\theta_2-\alpha(\nabla h(\theta_1)-\nabla h(\theta_2))\|^2\\
    =&\|\theta_1-\theta_2\|^2-2\alpha (\nabla h(\theta_1)-\nabla h(\theta_2)^T(\theta_1-\theta_2) +\alpha^2 t^2\\
    \leq&\|\theta_1-\theta_2\|^2-\frac{2\alpha}{\beta}(t-\eta)^2+\alpha^2 t^2\\
    =&\|\theta_1-\theta_2\|^2-\frac{2\alpha}{\beta}(t-\eta)^2-\frac{2\alpha\eta}{\beta}(t-\eta)+\alpha^2 t^2+\frac{2\alpha\eta}{\beta}(t-\eta)\\
     =&\|\theta_1-\theta_2\|^2-\frac{2\alpha t}{\beta}(t-\eta)+\alpha^2 t^2+\frac{2\alpha\eta}{\beta}(t-\eta).\\
\end{aligned}
\end{equation*}
Let $\alpha\leq 1/\beta$, then
\begin{equation*}
\begin{aligned}
     &\|\theta_1-\theta_2\|^2-\frac{2\alpha t}{\beta}(t-\eta)+\alpha^2 t^2+\frac{2\alpha\eta}{\beta}(t-\eta)\\
    \leq&\|\theta_1-\theta_2\|^2-2\alpha^2 t(t-\eta)+\alpha^2 t^2+\frac{2\alpha\eta}{\beta}(t-\eta)\\
    \leq&\|\theta_1-\theta_2\|^2-\alpha^2 (t+\eta)(t-\eta)+\alpha^2 t^2+\frac{2\alpha\eta}{\beta}(t-\eta)\\
    \leq&\|\theta_1-\theta_2\|^2+\alpha^2 \eta^2+\frac{2\alpha\eta}{\beta}(t-\eta).\\
\end{aligned}
\end{equation*}
By the definition of $\eta$-approximate smoothness, 
\[
\frac{1}{\beta}(t-\eta)\leq\|\theta_1-\theta_2\|.
\]
Then
\begin{equation*}
\begin{aligned}
    &\|\theta_1-\theta_2\|^2+\alpha^2 \eta^2+\frac{2\alpha\eta}{\beta}(t-\eta)\\
    \leq&\|\theta_1-\theta_2\|^2+\alpha^2 \eta^2+2\alpha\eta\|\theta_1-\theta_2\|\\
    =&(\|\theta_1-\theta_2\|+\alpha\eta)^2.
\end{aligned}
\end{equation*}
Therefore, we obtain that
\begin{equation*}
    \|G_{\alpha,z}(\theta_1)-G_{\alpha,z}(\theta_2)\|\leq \|\theta_1-\theta_2\|+\alpha\eta.
\end{equation*}\qed\\
Proof of Lemma \ref{lem:7} ($\alpha\eta$-approximately $(1-\alpha\gamma)$-contraction.).\\
Firstly, if $h(\theta)$ is a $\gamma$-strongly convex, $\eta$-approximately $\beta$-gradient Lipschitz function, $\phi(\theta)=h(\theta)-\frac{\gamma}{2}\|\theta\|^2$ is a convex, $\eta$-approximate $(\beta-\gamma)$-gradient Lipschitz function. The proof of convexity follows the definition. To see the second claim, since
\begin{equation*}
\begin{aligned}
   &\phi(\theta_1)-\phi(\theta_2)\\
    =&h(\theta_1)-h(\theta_2)-(\frac{\gamma}{2}\|\theta_1\|^2-\frac{\gamma}{2}\|\theta_2\|^2)\\
    \leq & \langle \nabla_\theta h(\theta_2),\theta_1-\theta_2\rangle +\frac{\beta}{2}\|\theta_1-\theta_2\|^2+\eta\|\theta_1-\theta_2\|-(\frac{\gamma}{2}\|\theta_1\|^2-\frac{\gamma}{2}\|\theta_2\|^2)\\
    \leq & \langle \nabla_\theta \phi(\theta_2),\theta_1-\theta_2\rangle +\frac{\beta}{2}\|\theta_1-\theta_2\|^2+\eta\|\theta_1-\theta_2\|-(\frac{\gamma}{2}\|\theta_1\|^2-\frac{\gamma}{2}\|\theta_2\|^2)+\gamma\theta_2^T(\theta_1-\theta_2)\\
    \leq & \langle \nabla_\theta \phi(\theta_2),\theta_1-\theta_2\rangle +\frac{\beta}{2}\|\theta_1-\theta_2\|^2+\eta\|\theta_1-\theta_2\|-\frac{\gamma}{2}\|\theta_1-\theta_2\|^2\\
    \leq & \langle \nabla_\theta \phi(\theta_2),\theta_1-\theta_2\rangle +\frac{\beta-\gamma}{2}\|\theta_1-\theta_2\|^2+\eta\|\theta_1-\theta_2\|.\\
\end{aligned}
\end{equation*}
Therefore, $\phi(\theta)$ satisfies the $\eta$-approximate $(\beta-\gamma)$-descent Lemma. Let $t=\|\nabla \phi(\theta_1)-\nabla \phi(\theta_2)\|$.\\
If $t\leq\eta$, we have
\begin{equation*}
\begin{aligned}
    &\|G_{\alpha,z}(\theta_1)-G_{\alpha,z}(\theta_2)\|\\
    =&\|\theta_1-\theta_2-\alpha(\nabla h(\theta_1)-\nabla h(\theta_2))\|\\
    =&\|\theta_1-\theta_2-\alpha(\nabla \phi(\theta_1)+\gamma\theta_1-\nabla \phi(\theta_2)-\gamma\theta_2)\|\\
    \leq&\|(1-\alpha\gamma)(\theta_1-\theta_2)\|+\alpha\|\nabla \phi(\theta_1)-\nabla \phi(\theta_2)\|\\
    \leq&(1-\alpha\gamma)\|\theta_1-\theta_2\|+\alpha\eta.\\
\end{aligned}
\end{equation*}
If $t\geq\eta$, we have
\begin{equation*}\small
\begin{aligned}
    &\|G_{\alpha,z}(\theta_1)-G_{\alpha,z}(\theta_2)\|^2\\
    =&\|\theta_1-\theta_2-\alpha(\nabla h(\theta_1)-\nabla h(\theta_2))\|^2\\
    =&\|\theta_1-\theta_2-\alpha(\nabla \phi(\theta_1)+\gamma\theta_1-\nabla \phi(\theta_2)-\gamma\theta_2)\|^2\\
    \leq&(1-\alpha\gamma)^2\|\theta_1-\theta_2\|^2-2\alpha(1-\alpha\gamma)(\nabla \phi(\theta_1)-\nabla \phi(\theta_2)^T(\theta_1-\theta_2) +\alpha^2 t^2\\
    \leq&(1-\alpha\gamma)^2\|\theta_1-\theta_2\|^2-\frac{2\alpha(1-\alpha\gamma)}{\beta-\gamma}(t-\eta)^2 +\alpha^2 t^2\\
        \leq&(1-\alpha\gamma)^2\|\theta_1-\theta_2\|^2-\frac{2\alpha(1-\alpha\gamma)}{\beta-\gamma}(t-\eta)^2-\frac{2\alpha(1-\alpha\gamma)\eta}{\beta-\gamma}(t-\eta) +\alpha^2 t^2+\frac{2\alpha(1-\alpha\gamma)\eta}{\beta-\gamma}(t-\eta).\\
\end{aligned}
\end{equation*}
Since $\alpha\leq1/\beta$, we have $(1-\alpha\gamma)/(\beta-\gamma)\geq\alpha$, then
\begin{equation*}\small
\begin{aligned}
&(1-\alpha\gamma)^2\|\theta_1-\theta_2\|^2-\frac{2\alpha(1-\alpha\gamma)}{\beta-\gamma}(t-\eta)^2-\frac{2\alpha(1-\alpha\gamma)\eta}{\beta-\gamma}(t-\eta) +\alpha^2 t^2+\frac{2\alpha(1-\alpha\gamma)\eta}{\beta-\gamma}(t-\eta)\\
\leq&(1-\alpha\gamma)^2\|\theta_1-\theta_2\|^2-\alpha^2 t (t-\eta)+\alpha^2 t^2+\frac{2\alpha(1-\alpha\gamma)\eta}{\beta-\gamma}(t-\eta)\\
\leq&(1-\alpha\gamma)^2\|\theta_1-\theta_2\|^2+\alpha^2 \eta^2+2\alpha(1-\alpha\gamma)\eta\|\theta_1-\theta_2\|\\
\leq&\bigg((1-\alpha\gamma)\|\theta_1-\theta_2\|+\alpha\eta\bigg)^2.\\
\end{aligned}
\end{equation*}
Therefore, we obtain that
\begin{equation*}
    \|G_{\alpha,z}(\theta_1)-G_{\alpha,z}(\theta_2)\|\leq (1-\alpha\gamma)\|\theta_1-\theta_2\|+\alpha\eta.
\end{equation*}\qed
\subsection{Proof of Theorem \ref{thm:covgen}}
The proof follows the standard techniques for uniform stability. We need to replace the non-expansive property used in standard analysis by the approximately non-expansive property. Let $S$ and $S'$ be two samples of size $n$ differing in only a single
example. Consider two trajectories $\theta_1^1,\dots,\theta_1^T$ and $\theta_2^1,\dots,\theta_2^T$
induced by running SGD on sample $S$ and $S',$ respectively. Let $\delta_t=\|\theta_1^t-\theta_2^t\|$.

Fixing an example $z\in Z$ and apply
the Lipschitz condition on $h(\cdot\,;z)$, we have
\begin{equation}\label{eq:convex-diff}
\E\left|h(\theta_1^T;z)-h(\theta_2^T;z)\right| \le
L\E\left[\delta_T\right]\,.
\end{equation}

Observe that at step $t,$ with probability $1-1/n,$ the
example selected by SGD is the same in both $S$ and $S'.$ 
With probability $1/n$ the selected example is
different. Therefore, by the $\alpha\eta$-approximate non-expansive property, we have

\begin{align}\label{eq:convex-recursion}
\E\left[\delta_{t+1}\right]
 \le \left(1-\frac1n\right)\bigg(\E\left[\delta_t\right]+\alpha_t\eta\bigg) +
 \frac1n\E\left[\delta_t\right]  +
\frac{2\alpha_t L}n  \le \E\left[\delta_t\right] + \bigg(\eta+\frac{2L}{n}\bigg)\alpha_t\,.
\end{align}

Unraveling the recursion gives
\[
\E\left[\delta_T\right] \le\bigg(\eta+\frac{2L}{n}\bigg)\sum_{t=1}^T\alpha_t\,.
\]
Plugging this back into Eq.~\eqref{eq:convex-diff}, 
we obtain
\[
\mathcal{E}_{gen} \le L \bigg(\eta+\frac{2L}{n}\bigg)\sum_{t=1}^T\alpha_t\,.
\]
Since this bounds holds for all $S,S'$ and $z,$ we obtain the desired bound on
the uniform stability.\qed

\subsection{Proof of Theorem \ref{thm:lower}}
Proof: The construction of function $h$ is adopted from the construction in the work of \citep{bassily2020stability}.\\
Let $T\leq d$, and $v, K\geq 0$. Considering $\mathcal{Z}=\{0,1\}$, and the objective function
\begin{equation}
h(\theta,z)=\begin{cases}
\eta\max\{0,x_1-v,\cdots,x_T-v\}\ \text{if}\ z=0\\
\langle r,x\rangle/K\quad \quad\quad\quad\quad\quad\quad\quad\quad\ \text{if}\ z=1,
\end{cases}
\end{equation}
where $r=(-1,\cdots,-1,0,\cdots,0)$ (i.e., equals to -1 for the first $T$ components). Function $h$ is $\eta$-approximately smooth since the first case is a piece-wise linear function. For the dataset $S$ and $S'$ differ in at most one sample, the empirical objective functions are
\[
R_S(\theta)=\frac{1}{nK}\langle r,x\rangle+\frac{n-1}{n}\eta\max\{0,x_1-v,\cdots,x_T-v\},
\]
and
\[
R_{S'}(\theta)=\eta\max\{0,x_1-v,\cdots,x_T-v\}.
\]
Let $\theta_1^T$ and $\theta_2^T$ be the trajectories running algorithm on dataset $S$ and $S'$, initialized on $\theta_1^0=\theta_2^0=0$. Clearly, $\theta_2^t=0$ for all $t$. It is easy to obtain $\theta_1^t=-\frac{t\alpha r}{nk}-\alpha\eta\frac{n-1}{n}\sum_{s=1}^{t-1}e_s$ recursively. By the orthogonality of the subgradients, we have
\begin{eqnarray}\label{lw1}
	\delta(S,S')=\|\theta_1^T-\theta_2^T\|=\|\theta_1^T\|\geq\Omega\bigg(\alpha\eta\bigg\|\sum_{s=1}^{T}e_s\bigg\|\bigg)=\Omega(\alpha\eta\sqrt{T}).
\end{eqnarray}
On the other hand, the work of \citep{hardt2016train} provided a lower bound for general non-smooth function
\begin{eqnarray}\label{lw2}
\delta(S,S')\geq\Omega\bigg(\frac{L\alpha T}{n}\bigg).
\end{eqnarray}
Combining Eq. (\ref{lw1}) and Eq. (\ref{lw2}), we have
\begin{eqnarray}\label{lw3}
\delta(S,S')\geq\Omega\bigg(\alpha\eta\sqrt{T}+\frac{L\alpha T}{n}\bigg).
\end{eqnarray}\qed
\subsection{Proof of Theorem \ref{thm:noncovgen}}
We consider a general form of Theorem \ref{thm:noncovgen}.
\begin{theorem}[Non-convex]
	Assume that $h(\theta,z)$ is $L$-Lipschitz, and $\eta$-approximately $\beta$-gradient Lipschitz in $\theta$ for all given $z\in\mathcal{Z}$. Assume in addition that $0\leq g(\theta,z)\leq B$ for all $\theta$ and $z$. Suppose that we run SGD on the adversarial surrogate loss with step sizes $\alpha_t\leq c/t$ for $T$ steps, where $c>0$. Then, for all $t_0\in\{1,2,\cdots,n\}$, adversarial training satisfies
uniform stability with
\begin{equation}
    \mathcal{E}_{gen}=\mathbb{E}[R_\mathcal{D}(\theta^T)-R_S(\theta^T)]\leq \frac{B t_0}{n-1}+\frac{1}{\beta (n-1)}\bigg(2L^2+L\eta n\bigg)\bigg(\frac{T}{t_0}\bigg)^{\beta c}.
\end{equation}
Let $q=\beta c$. For small $T$ (s.t. $t_0\leq n$ when we set the first term equals to the second term). we select $t_0$ to optimize the right hand side, then
\begin{equation}
    \mathcal{E}_{gen}\leq \frac{2}{n-1}B^{\frac{q}{q+1}}\bigg(\frac{2L^2+L\eta n}{\beta}\bigg)^{\frac{1}{q+1}}T^{\frac{q}{q+1}}.
\end{equation}
For arbitrary $T$, the optimal $t_0>n$. We simply let $t_0=1$, then
\begin{equation}
    \mathcal{E}_{gen}\leq \frac{BL_\theta+(2L^2+L\eta  n)T^{q}}{\beta(n-1)}.
\end{equation}
\end{theorem}
Proof: Let $S$ and $S'$ be two samples of size $n$ differing in only a single
example. Consider two trajectories $\theta_1^1,\dots,\theta_1^T$ and $\theta_2^1,\dots,\theta_2^T$
induced by running SGD on sample $S$ and $S',$ respectively. Let $\delta_t=\|\theta_1^t-\theta_2^t\|$. Let $t_0\in\{0,1,\dots,n\},$ be the iteration that $\delta_{t_0}=0$, but SGD picks two different samples form $S$ and $S'$ in iteration $t_0+1$, then
\begin{equation}
\label{eq:nonconvex-diff}
    \mathcal{E}_{gen}\le \frac{t_0}n B
+ L\E\left[\delta_T\mid\delta_{t_0}=0\right]\,.
\end{equation}
Let $\Delta_t=\E\left[\delta_t\mid\delta_{t_0}=0\right]$. Observe that at step $t,$ with probability $1-1/n,$ the
example selected by SGD is the same in both $S$ and $S'.$ 
With probability $1/n$ the selected example is
different. Therefore, by the $\alpha\eta$-approximate ($1+\alpha\beta$)-expansive property, for every $t\ge t_0,$
\begin{align*}
\Delta_{t+1}
& \le \left(1-\frac1n\right)(1+\alpha_t\beta)\Delta_t + \frac1n\Delta_t  +
\bigg(\eta+\frac{2 L}n\bigg)\alpha_t\\
& \le \left(\frac1n +  (1-1/n)(1+c\beta/t)\right)\Delta_t + \bigg(\eta+\frac{2 L}n\bigg)\frac{c}{t}\\
& = \left(1 + (1-1/n)\frac{c\beta}t\right)\Delta_t + \bigg(\eta+\frac{2 L}n\bigg)\frac{c}{t}\\
& \le \exp\left((1-1/n)\frac{c\beta}t\right)\Delta_t +\bigg(\eta+\frac{2 L}n\bigg)\frac{c}{t}\,.
\end{align*}
Here we used the fact that $1+x\le\exp(x)$ for all $x.$

Using the fact that $\Delta_{t_0}=0,$ we can unwind this recurrence relation
from $T$ down to $t_0+1.$ This gives
\begin{align*}
\Delta_T &\leq \sum_{t=t_0+1}^T \left\{\prod_{k=t+1}^T
\exp\left((1-\tfrac{1}{n}) \tfrac{\beta c}{k}\right) \right\} \bigg(\eta+\frac{2 L}n\bigg)\frac{c}{t}\\
&= \sum_{t=t_0+1}^T \exp\left((1-\tfrac{1}{n}) \beta c \sum_{k={t+1}}^T
\tfrac{1}{k} \right) \bigg(\eta+\frac{2 L}n\bigg)\frac{c}{t}\\
&\leq \sum_{t=t_0+1}^T \exp\left((1-\tfrac{1}{n}) \beta c \log(\tfrac{T}{t})
\right) \bigg(\eta+\frac{2 L}n\bigg)\frac{c}{t} \\
& =  \bigg(\eta+\frac{2 L}n\bigg)c T^{\beta c (1-1/n)} \sum_{t=t_0+1}^T t^{-\beta c (1-1/n) - 1}\\
&\leq  \bigg(\eta+\frac{2 L}n\bigg)\frac1{(1-1/n)\beta c} c \left(\frac{T}{t_0}\right)^{\beta c (1-1/n)} \\
&\le \frac{\eta n+2L}{\beta(n-1)} \left(\frac{T}{t_0}\right)^{\beta c}\,,
\end{align*}
Plugging this bound into~\eqref{eq:nonconvex-diff},
we get
\[
\mathcal{E}_{gen} \le \frac{Bt_0}{n-1} +
\frac{L\eta n +2L^2}{\beta(n-1)} \left(\frac{T}{t_0}\right)^{\beta c}\,.
\]
Let $q=\beta c$. For small $T$ (s.t. $t_0\leq n$ when we set the first term equals to the second term). we select $t_0$ to optimize the right hand side, then
\begin{equation*}
    \mathcal{E}_{gen}\leq \frac{2}{n-1}B^{\frac{q}{q+1}}\bigg(\frac{2L^2+L\eta n}{\beta}\bigg)^{\frac{1}{q+1}}T^{\frac{q}{q+1}}.
\end{equation*}
For arbitrary $T$, the optimal $t_0>n$. We simply let $t_0=1$, then
\begin{equation*}
    \mathcal{E}_{gen}\leq \frac{BL_\theta+(2L^2+L\eta  n)T^{q}}{\beta(n-1)}.
\end{equation*}

Since the bound we just derived holds for all $S,S'$ and $z,$ we immediately get
the claimed upper bound on the uniform stability. Let $q=1$, we obtain the result of Theorem \ref{thm:noncovgen}.\qed
\subsection{Proof of Theorem \ref{thm:stronggen}}
The proof follows the idea in convex case. By the $\alpha\eta$-approximately $(1-\alpha\gamma)$-contraction, for every $t,$
\begin{align}\label{eq:sconvex-recursion}
\E\delta_{t+1}
& \le \left(1-\frac1n\right)(1-\alpha \gamma)\E\delta_t + \frac1n(1-\alpha \gamma)\E\delta_t  +
\bigg(\eta+\frac{ 2L}n\bigg )\alpha \\
\nonumber & = \left(1- \alpha \gamma \right)\E\delta_t    +
\bigg(\eta+\frac{2 L}n\bigg )\alpha \,.
\end{align}
Unraveling the recursion gives
\[
\E\delta_T \le \bigg(\eta+\frac{2 L}n\bigg )\alpha\sum_{t=0}^T\left(1- \alpha \gamma \right)^t
\le\frac{\eta}{\gamma}+\frac{2L}{\gamma n}
\,.
\]
Plugging the above inequality into Eq. ~\eqref{eq:convex-diff}, we obtain
\[
\mathcal{E}_{gen} \le \frac{L\eta}{\gamma}+\frac{2 L^2}{\gamma n}\,.
\]
Since this bounds holds for all $S,S'$ and $z,$ the Theorem follows.\qed
\subsection{Proof of Theorem \ref{thm:swagen}}
Let $\bar{\theta} = \frac{1}{T}\sum_{t=1}^T \theta^t$ denote the
average of the stochastic gradient iterates. Since
\[
\theta^t = \sum_{k=1}^t \alpha_k \nabla h(\theta^k; z_k)\,,
\]
we have
\[
 \bar{\theta} =  \sum_{t=1}^T  \alpha_t\frac{T-t+1}{T}  \nabla h(\theta^k; z_k)
\]
Using the $\alpha\eta$-approximate non-expansive, we have
\[
	\delta_t \leq (1-1/n) \delta_{t-1} +
		\frac{1}{n}\left( \delta_{t-1} + (\eta n+2  L) \alpha_t \frac{T-t+1}{T}\right)\,.
\]
which implies
\[
	\delta_T \leq  \bigg(\eta+\frac{2 L}{n}\bigg) \sum_{t=1}^T \alpha_t \frac{T-t+1}{T} =
	\bigg(\frac{\eta}{2} +\frac{L}{n}\bigg)\sum_{t=1}^T\alpha_t\,.
\]
Since $f$ is $L$-Lipschitz, we have
\begin{equation}
    \mathcal{E}_{gen}(\bar{\theta})\leq \bigg(\frac{L\eta}{2} +\frac{L^2}{n}\bigg)\sum_{t=1}^T\alpha_t.
\end{equation}

Here the expectation is taken over the algorithm and hence the claim follows by
our definition of uniform stability. $\mathcal{E}_{opt}$ follows \citep{nemirovski2009robust}.\qed

\section{Discussion on Non-convex and Strongly Convex Case}
\label{b}
\subsection{Discussion on Non-convex Case}
\label{b1}
To discuss the generalization-optimization trade-off in the non-convex case. We first give the optimization error bound.

\begin{theorem}
\label{converge2}
Assume that $h$ is $\eta$-approximate $\beta$-gradient Lipschitz and given $0<\tau<1$. Without loss of generality, assume the stochastic gradient $\nabla \tilde{h}(\theta)$ be unbiased and have a bounded variance $\sigma^2$. Let the stochastic gradient descent (SGD) update be $\theta_{t+1}=\theta_t-\alpha\nabla \tilde{h}(\theta_t)$ with a constant step size $\alpha =1/\sqrt{T}$ for number of iterations $T\geq (\beta/2(1-\tau))^2$. $\exists t\leq T$, s.t.
\begin{equation}
\mathbb{E}\|\nabla h(\theta_t)\|^2\leq\frac{\eta^2}{\tau^2}+\frac{2\eta\sigma}{\tau}+\mathcal{O}(\frac{1}{\sqrt{T}}).
\end{equation}
\end{theorem}
Proof:

Assume that the stochastic gradient $\nabla h(\theta)$ be unbiased and have a bounded variance $\sigma^2$.
\[
\mathbb{E}[\nabla \tilde{h}(\theta)]=\nabla h(\theta),
\]
\[
\mathbb{E}\|\nabla \tilde{h}(\theta)\|^2\leq\|\nabla h(\theta)\|^2+\sigma^2.
\]
Notice that when $\theta$ is a random vector, the above expectation is condition on $\theta$. Let the stochastic gradient descent (SGD) update be $\theta_{t+1}=\theta_t-\alpha\tilde{h}(\theta_t)$ with a constant step size $\alpha=1/\sqrt{T}$. By $\eta$-approximately Descent Lemma, we have
\begin{equation*}
\begin{aligned}
   &h(\theta_{t+1})-h(\theta_{t})\\
      =&-\alpha\langle \nabla h(\theta_t),\theta_{t+1}-\theta_t\rangle +\frac{\beta}{2}\|\theta_{t+1}-\theta_t\|^2+\eta\|\theta_{t+1}-\theta_t\|\\
      =&-\alpha\langle \nabla h(\theta_t),\nabla\tilde{h}(\theta_t)\rangle +\frac{\beta\alpha^2}{2}\|\nabla  \tilde{h}(\theta_t)\|^2+\eta\alpha\|\nabla \tilde{h}(\theta_t)\|.\\
\end{aligned}
\end{equation*}
Given $\theta_t$, take the conditional expectation over the noised introduced by SGD, we have
\begin{equation*}
\begin{aligned}
   &\mathbb{E}[h(\theta_{t+1})]-h(\theta_{t})\\
      =&-\alpha\langle \nabla h(\theta_t),\mathbb{E}[\nabla  \tilde{h}(\theta_t)]\rangle +\frac{\beta\alpha^2}{2}\mathbb{E}\|\nabla \tilde{h}(\theta_t)\|^2+\eta\alpha\mathbb{E}\|\nabla \tilde{h}(\theta_t)\|\\
      \leq &-\alpha\|\nabla h(\theta_t)\|^2+\frac{\beta\alpha^2}{2}\big[\|\nabla h(\theta_t)\|^2+\sigma^2\big]+\eta\alpha\sqrt{\big[\mathbb{E}\|\nabla\tilde{h}(\theta_t)\|\big]^2}\\
      \leq &-\alpha\|\nabla h(\theta_t)\|^2+\frac{\beta\alpha^2}{2}\big[\|\nabla h(\theta_t)\|^2+\sigma^2\big]+\eta\alpha\sqrt{\|\nabla h(\theta_t)\|^2+\sigma^2}\\
      \leq &-\alpha\|\nabla h(\theta_t)\|^2+\frac{\beta\alpha^2}{2}\big[\|\nabla h(\theta_t)\|^2+\sigma^2\big]+\eta\alpha\big[\|\nabla h(\theta_t)\|+\sigma\big]\\
      = &-\alpha\|\nabla h(\theta_t)\|^2+\frac{\beta\alpha^2}{2}\|\nabla h(\theta_t)\|^2+\eta\alpha\|\nabla_\theta  h(\theta_t)\|+\frac{\beta\alpha^2\sigma^2}{2}+\eta\alpha\sigma\\
      \leq&-\tau\alpha\|\nabla h(\theta_t)\|^2+\eta\alpha\|\nabla h(\theta_t)\|+\frac{\beta\alpha^2\sigma^2}{2}+\eta\alpha\sigma,
\end{aligned}
\end{equation*}
where the first inequality is the assumption of SGD, the second inequality is the Jensen's inequality, the third one is the  assumption of SGD, the fourth one is the Cauchy-Schwartz inequality, and the last one is because of the size of step size $\alpha$. Take the expectation over the trajectory $\theta_0,\theta_1,\cdots,\theta_T$, and take the average aver $t=0,1,\cdots,T$, we have
\begin{equation*}
\begin{aligned}
   &\frac{1}{T}\sum_{t=0}^T\big[\tau\alpha\mathbb{E}\|\nabla h(\theta_t)\|^2-\eta\alpha\mathbb{E}\|\nabla h(\theta_t)\|\big]\\
      \leq&\frac{1}{T}\sum_{t=0}^T\big[\mathbb{E}[h(\theta_{t})]-\mathbb{E}h(\theta_{t+1})\big]+\frac{\beta\alpha^2\sigma^2}{2}+\eta\alpha\sigma\\
      \leq&\frac{1}{T}\big[\mathbb{E}[h(\theta_{0})]-h(\theta_{*})\big]+\frac{\eta\alpha^2\sigma^2}{2}+\eta\alpha\sigma.\\
\end{aligned}
\end{equation*}
Let $\mathbb{E}[h(\theta_{0})]-h(\theta_{*})=D$ and divide $\alpha$ on both side. $\exists t\leq T$, $s.t.$
\begin{equation*}
\begin{aligned}
   \tau\mathbb{E}\|\nabla h(\theta_t)\|^2-\eta\mathbb{E}\|\nabla h(\theta_t)\|
      \leq\frac{D}{T\alpha}+\frac{\beta\alpha\sigma^2}{2}+\eta\sigma.\\
\end{aligned}
\end{equation*}
Since $\alpha=1/\sqrt{T}$, we have 
\begin{equation*}
\begin{aligned}
   &\tau\mathbb{E}\|\nabla h(\theta_t)\|^2-\eta\mathbb{E}\|\nabla h(\theta_t)\|
      \leq\frac{D}{\sqrt{T}}+\frac{\beta\sigma^2}{2\sqrt{T}}+\eta\sigma\\
      \Leftrightarrow&\mathbb{E}\|\nabla h(\theta_t)\|^2-\frac{\eta}{\tau}\mathbb{E}\|\nabla h(\theta_t)\|+(\frac{\eta}{2\tau})^2
      \leq\frac{1}{\sqrt{T}}\bigg[\frac{D}{\tau}+\frac{\beta\sigma^2}{2\tau}\bigg]+\frac{\eta\sigma}{\tau}+(\frac{\eta}{2\tau})^2\\
       \Leftrightarrow&\bigg|\mathbb{E}\|\nabla h(\theta_t)\|-\frac{\eta}{2\tau}\bigg|
      \leq\sqrt{\frac{1}{\sqrt{T}}\bigg[\frac{D}{\tau}+\frac{\beta\sigma^2}{2\tau}\bigg]+\frac{\eta\sigma}{\tau}+(\frac{\eta}{2\tau})^2}.\\
\end{aligned}
\end{equation*}
Then we remove the absolute value, and obtain
\begin{equation*}
\begin{aligned}
&\mathbb{E}\|\nabla h(\theta_t)\|-\frac{\eta}{2\tau}
      \leq\sqrt{\frac{1}{\sqrt{T}}\bigg[\frac{D}{\tau}+\frac{\beta\sigma^2}{2\tau}\bigg]+\frac{\eta\sigma}{\tau}+(\frac{\eta}{2\tau})^2}\\
      \Leftrightarrow& \mathbb{E}\|\nabla h(\theta_t)\|
      \leq\frac{\eta}{2\tau}+\sqrt{\frac{1}{\sqrt{T}}\bigg[\frac{D}{\tau}+\frac{\beta\sigma^2}{2\tau}\bigg]+\frac{\eta\sigma}{\tau}+(\frac{\eta}{2\tau})^2}\\
            \Leftrightarrow& \mathbb{E}\|\nabla h(\theta_t)\|^2
      \leq\bigg[\frac{\eta}{2\tau}+\sqrt{\frac{1}{\sqrt{T}}\bigg[\frac{D}{\tau}+\frac{\beta\sigma^2}{2\tau}\bigg]+\frac{\beta\sigma}{\tau}+(\frac{\eta}{2\tau})^2}\bigg]^2.\\
\end{aligned}
\end{equation*}
By Cauchy-Schwartz inequality, we have
\begin{equation*}
\begin{aligned}
 \mathbb{E}\|\nabla h (\theta_t)\|^2
      \leq&\bigg[\frac{\eta}{2\tau}+\sqrt{\frac{1}{\sqrt{T}}\bigg[\frac{D}{\tau}+\frac{\beta\sigma^2}{2\tau}\bigg]+\frac{\eta\sigma}{\tau}+(\frac{\eta}{2\tau})^2}\bigg]^2\\
      \leq&2\bigg[(\frac{\eta}{2\tau})^2+\frac{1}{\sqrt{T}}\bigg[\frac{D}{\tau}+\frac{\beta\sigma^2}{2\tau}\bigg]+\frac{\beta\sigma}{\tau}+(\frac{\eta}{2\tau})^2\bigg]\\
      =&\frac{\eta^2}{\tau^2}+\frac{2\eta\sigma}{\tau}+\frac{2}{\sqrt{T}}\bigg[\frac{D}{\tau}+\frac{\beta\sigma^2}{2\tau}\bigg]\\
            =&\frac{\eta^2}{\tau^2}+\frac{2\eta\sigma}{\tau}+\mathcal{O}(\frac{1}{\sqrt{T}}).\\
\end{aligned}
\end{equation*}\qed\\
In words, running SGD on an approximately smooth non-convex function, the algorithm cannot converge to a stationary point but with an additional constant term. Notice that this is an error bound for gradient norm. If we need an error bound for the optimality gap, we need an additional PL condition.
Combining the optimization error and generalization error, we have
\[
\mathcal{E}_{opt}+\mathcal{E}_{gen}\leq \mathcal{O}\bigg(\eta T+\frac{T}{n}+\frac{1}{\sqrt{T}}\bigg)+\text{constant},
\]
where the first term is an additional term for adversarial training, which induces robust overfitting. Therefore, we can see that the analysis of the convex and non-convex cases do not have a major difference. To simplify the argument, we only discuss the convex case in the main paper.
\subsection{Discussion on Strongly Convex Case}
\label{b2}
By \citep{nemirovski2009robust},  $$\mathcal{E}_{opt}\leq\frac{LD^2}{T}=\mathcal{O}(\frac{1}{T})$$ 
in the strongly convex case. Whether the function is smooth does not affect the convergence rate.
Therefore,
\[
\mathcal{E}_{opt}+\mathcal{E}_{gen}\leq \mathcal{O}\bigg(\eta +\frac{1}{n}+\frac{1}{T}\bigg).
\]
This result shows that robust overfitting will disappear if the loss function is strongly convex. But the performance of adversarial training is still worse than the performance of standard training in $\mathcal{O}(\eta )$ in this strong assumption. 

\subsection{Discussion on Strongly Concave Assumption on the Inner Problem}
\label{b3}
In this subsection, we discuss the case that $g(\theta,z)$ is $\mu$-strongly concave in $z$.
\begin{assumption}
	\label{ass2}
	The function $g$ satisfies the following Lipschitzian smoothness conditions:
	\begin{equation*}
	\begin{aligned}
	&\|g(\theta_1,z)-g(\theta_2,z)\|\leq L\|\theta_1-\theta_2\|,\\
	&\|\nabla_\theta g(\theta_1,z)-\nabla_\theta g(\theta_2,z)\|\leq L_{\theta}\|\theta_1-\theta_2\|,\\
	&\|\nabla_\theta g(\theta,z_1)-\nabla_\theta g(\theta,z_2)\|\leq L_{z}\|z_1-z_2\|,\\
	&\|\nabla_z g(\theta_1,z)-\nabla_\theta g(\theta_2,z)\|\leq L_{z\theta}\|\theta_1-\theta_2\|.\\
	\end{aligned}
	\end{equation*}
\end{assumption}
Assumption \ref{ass2} assumes that the loss function is smooth (in zeroth-order and first-order), which are also used in the stability literature \citep{farnia2021train,xing2021on}, as well as the convergence analysis literature \citep{wang2019convergence,liu2020loss}. Comparing with Assumption \ref{ass1}, Assumption \ref{ass2} requires one more gradient Lipschitz $\|\nabla_z g(\theta_1,z)-\nabla_\theta g(\theta_2,z)\|\leq L_{z\theta}\|\theta_1-\theta_2\|$.

\begin{lemmalist}
	Under Assumption \ref{ass2}, assume in addition that $g(\theta,z)$ is $\mu$-strongly concave in $z$. $\forall \theta_1, \theta_2$ and $\forall z\in\mathcal{Z}$, the following properties hold.
\begin{senenum}
\item 
(Lipschitz function.) $\|h(\theta_1,z)-h(\theta_2,z)\|\leq L\|\theta_1-\theta_2\|$.

\item  (gradient Lipschitz.) $\|\nabla_\theta h(\theta_1,z)-\nabla_\theta h(\theta_2,z)\|\leq \beta_2\|\theta_1-\theta_2\|$, where
\[
\beta_2=\frac{L_{z}L_{z\theta}}{\mu}+L_\theta.
\]
\end{senenum}
\end{lemmalist}
The proof can be found in \citep{sinha2017certifiable,wang2019convergence}. Therefore, the adversarial surrogate loss is $\beta_2$-gradient Lipschitz. The stability generalization bounds follows \citep{hardt2016train} by replacing $\beta$ by $\beta_2$ (for the choice of step size $\alpha$).

\section{Additional Experiments}
\label{c}
\begin{figure*}[htbp]
	\centering
	\hspace{-0.4in}\scalebox{0.9}{
		\subfigure[]{
			\begin{minipage}[htp]{0.2\linewidth}
				\centering
				\includegraphics[width=1.3in]{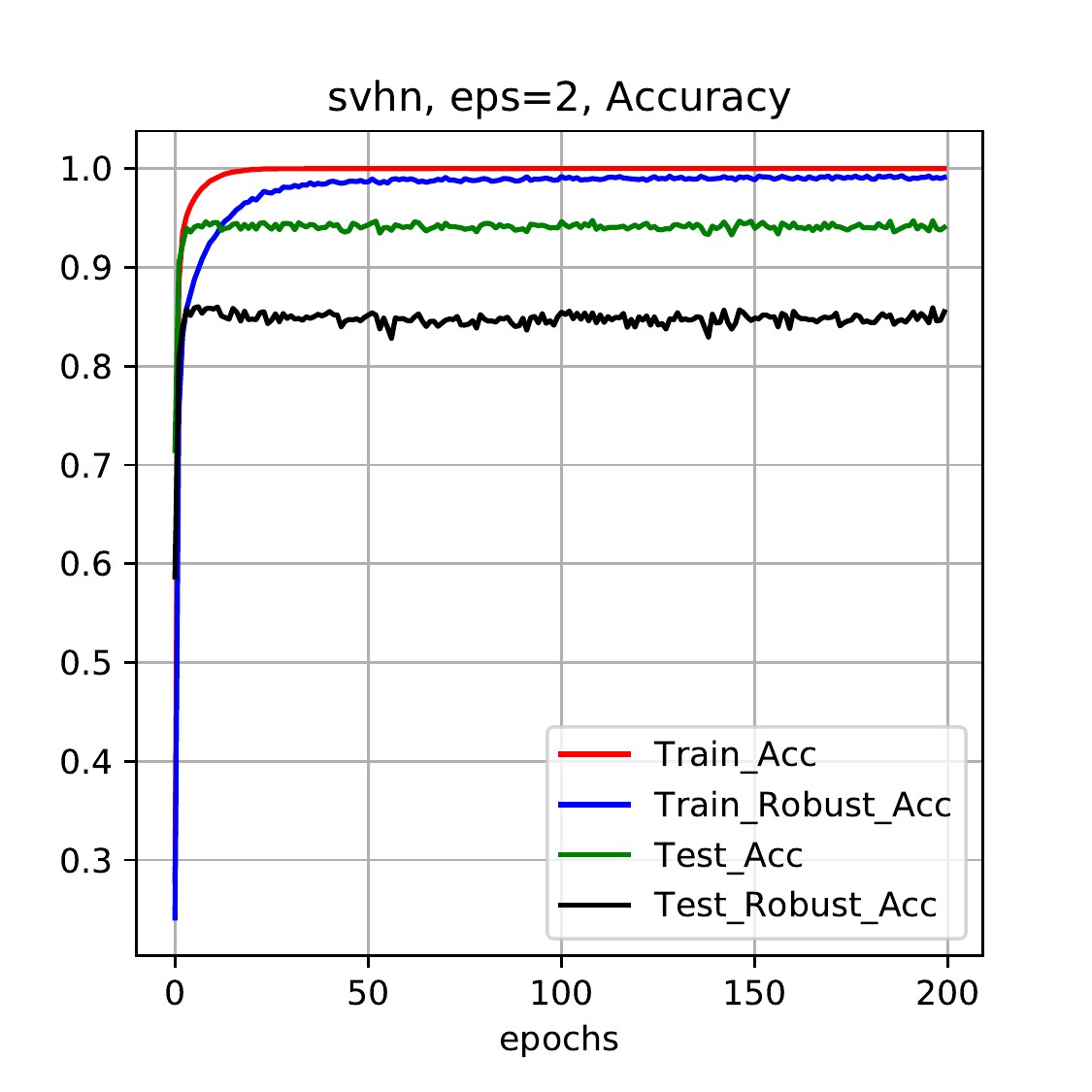}
			\end{minipage}%
		}
		\subfigure[]{
			\begin{minipage}[htp]{0.2\linewidth}
				\centering
				\includegraphics[width=1.3in]{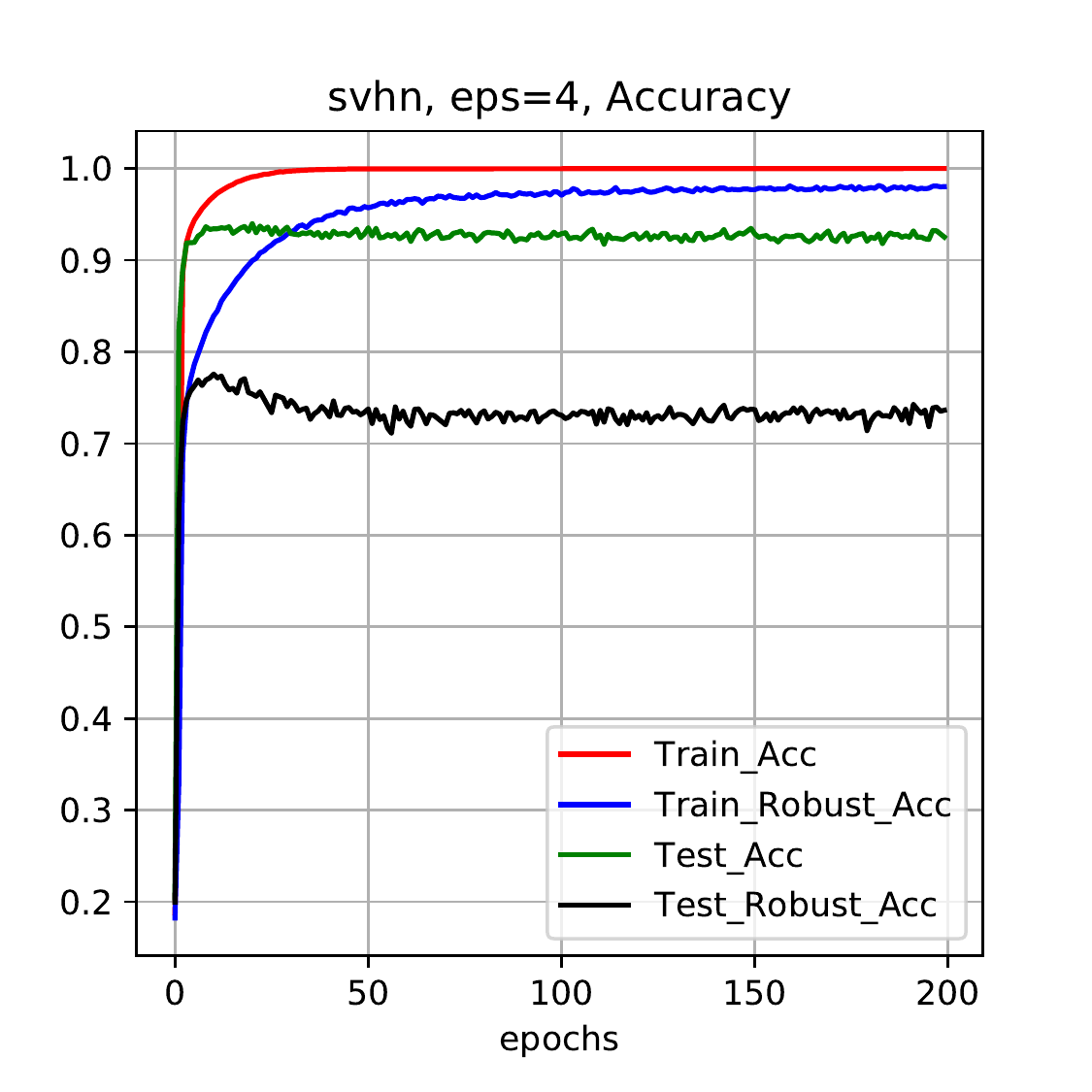}
			\end{minipage}
		}
		\subfigure[]{
			\begin{minipage}[htp]{0.2\linewidth}
				\centering
				\includegraphics[width=1.3in]{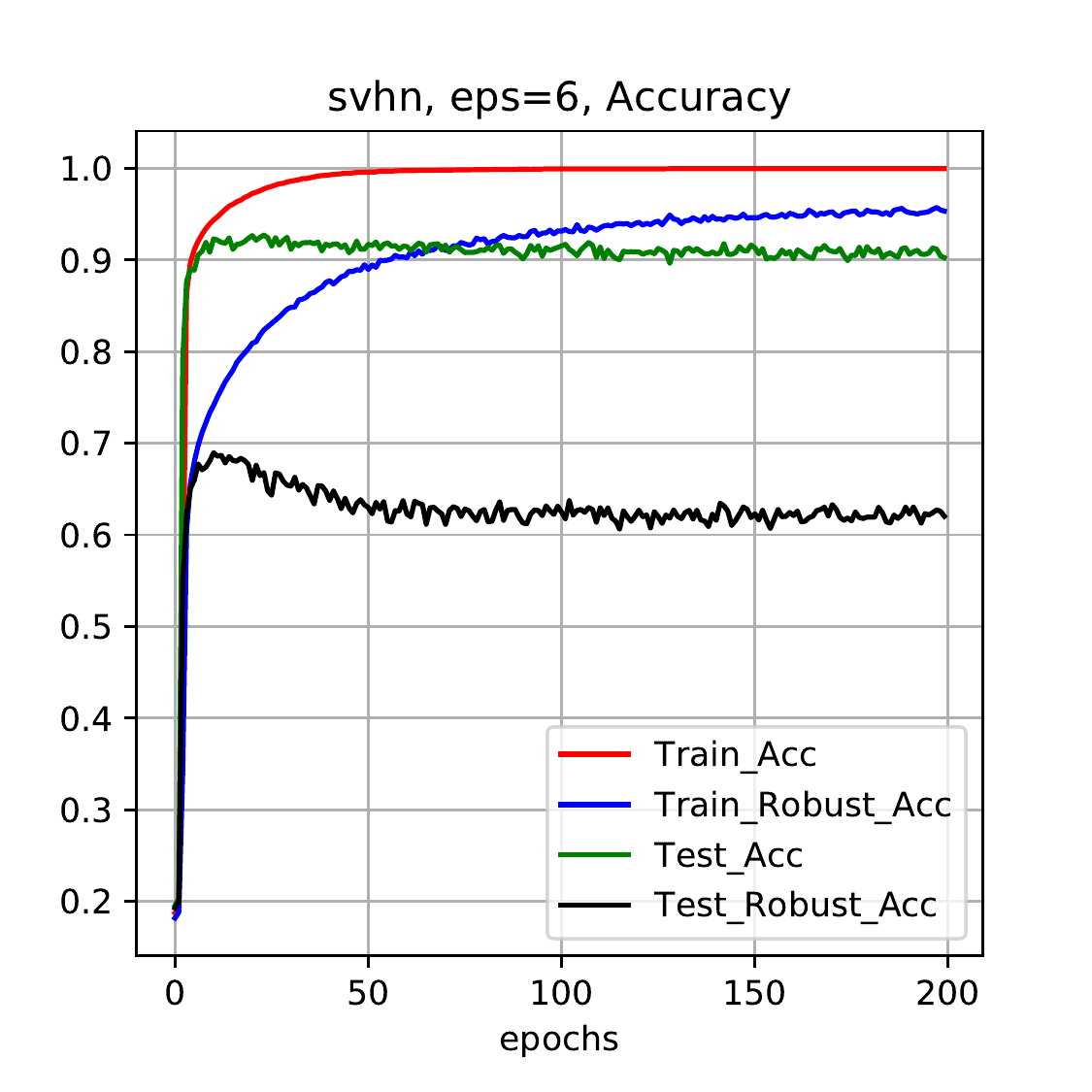}
			\end{minipage}
		}
		\subfigure[]{
			\begin{minipage}[htp]{0.2\linewidth}
				\centering
				\includegraphics[width=1.3in]{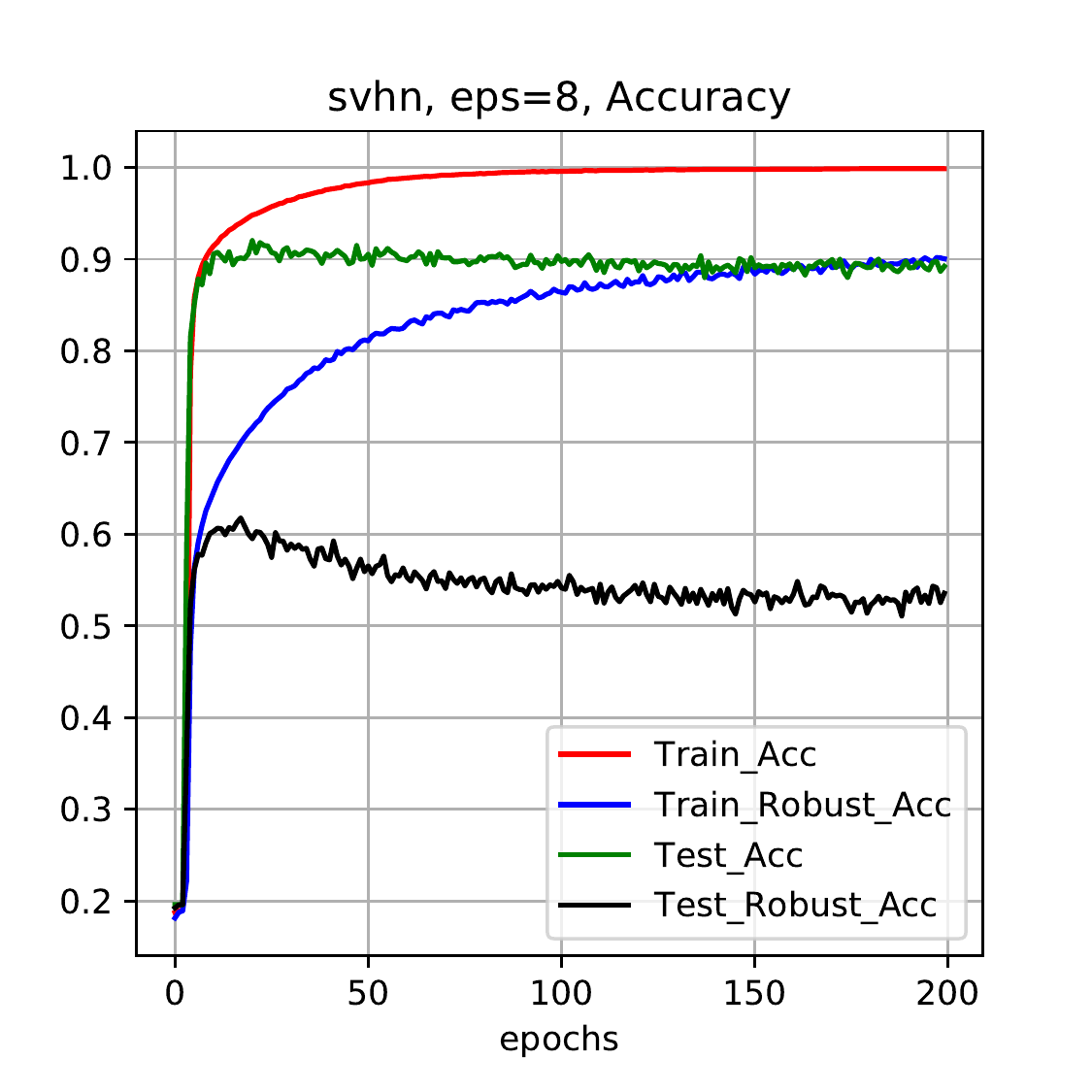}
			\end{minipage}
		}
		\subfigure[]{
			\begin{minipage}[htp]{0.2\linewidth}
				\centering
				\includegraphics[width=1.3in]{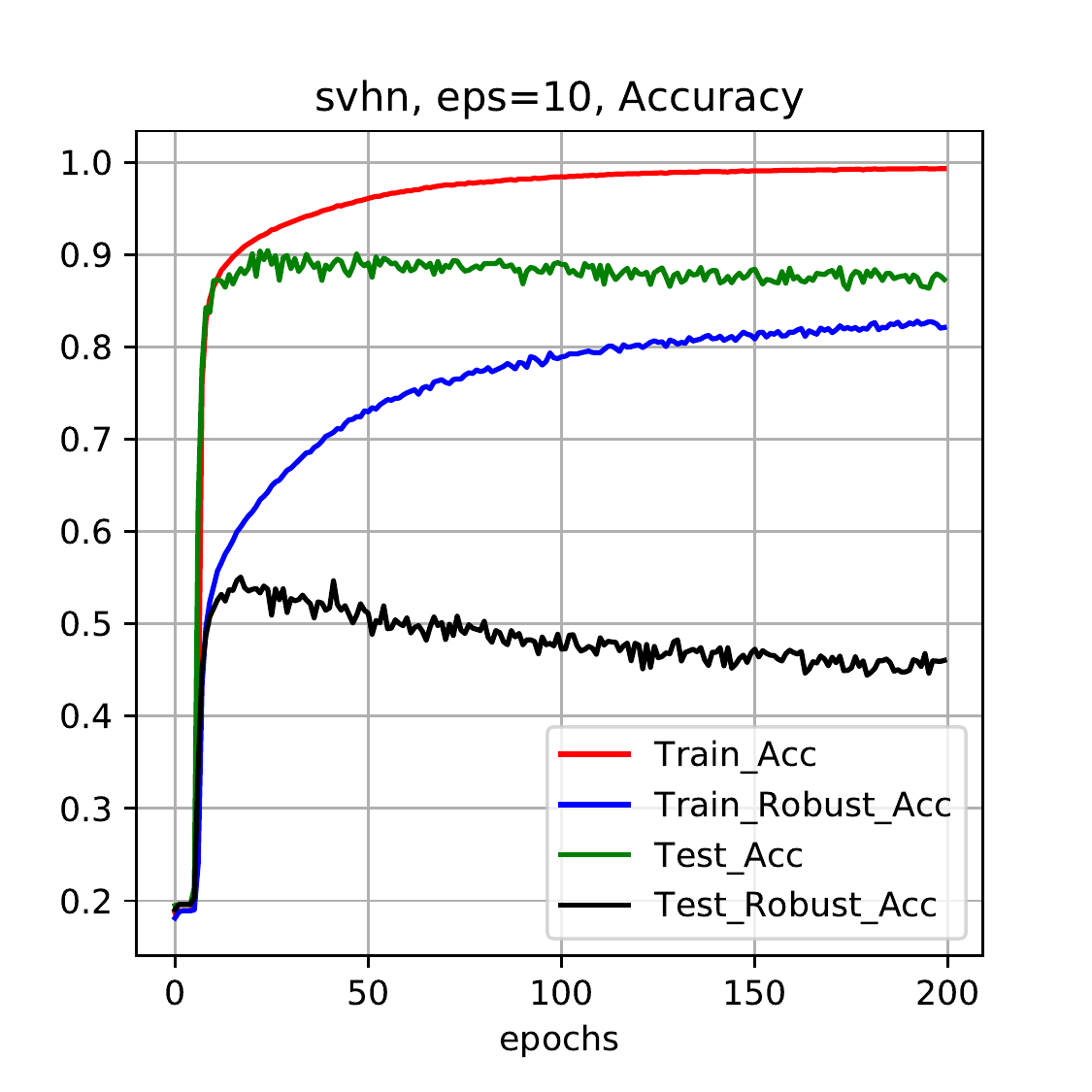}
			\end{minipage}
	}}
	
	\hspace{-0.4in}\scalebox{0.9}{
		\subfigure[]{
			\begin{minipage}[htp]{0.2\linewidth}
				\centering
				\includegraphics[width=1.3in]{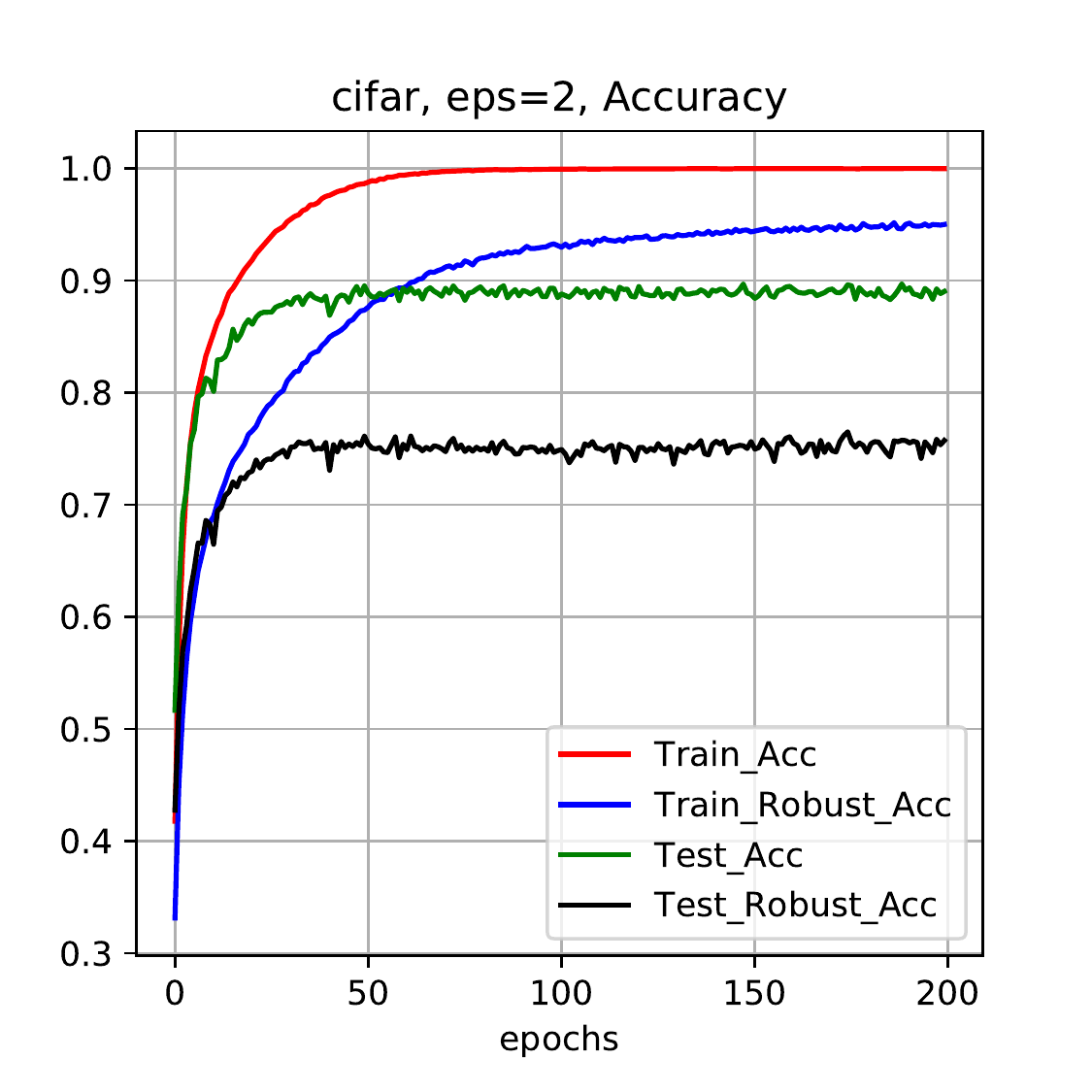}
			\end{minipage}%
		}
		\subfigure[]{
			\begin{minipage}[htp]{0.2\linewidth}
				\centering
				\includegraphics[width=1.3in]{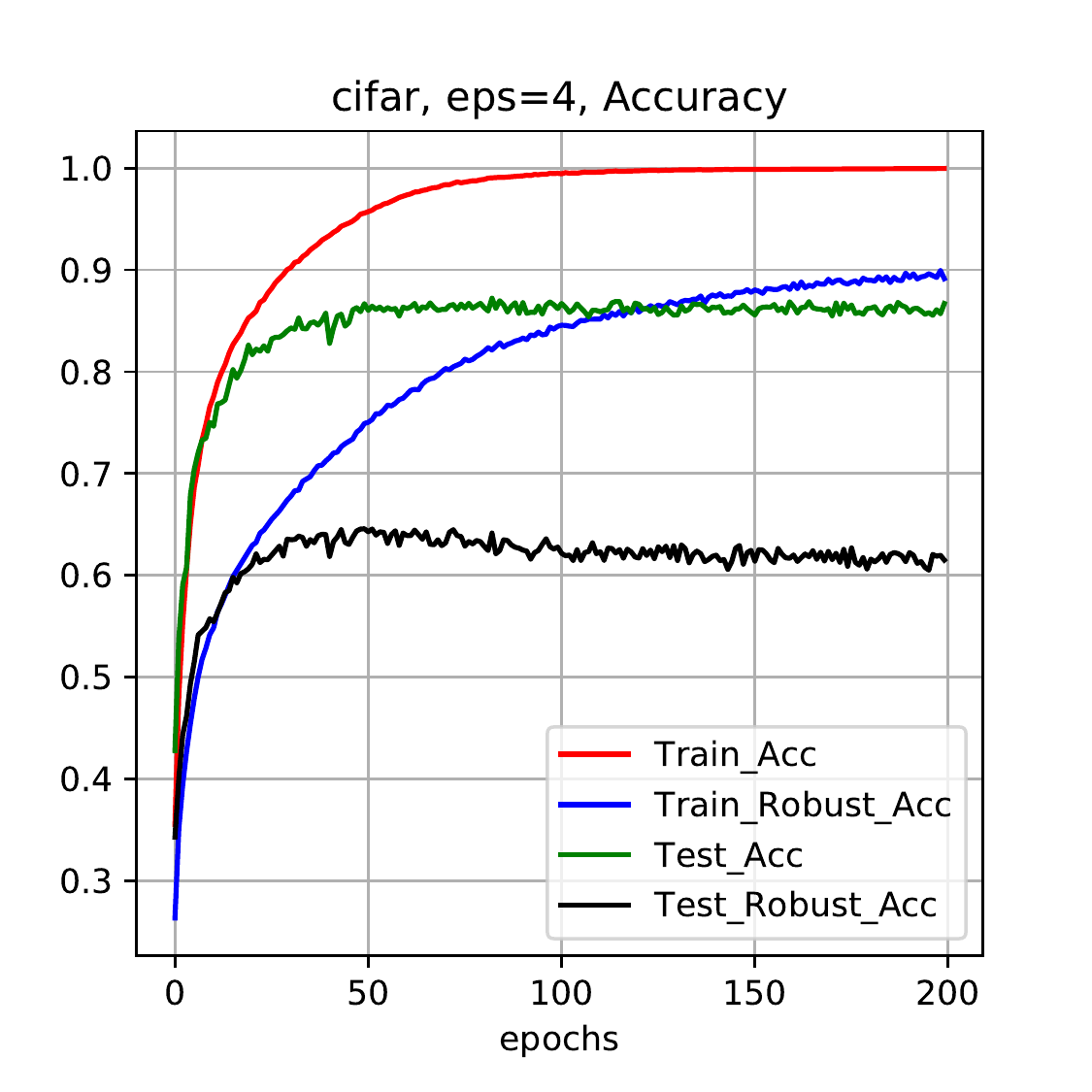}
			\end{minipage}
		}
		\subfigure[]{
			\begin{minipage}[htp]{0.2\linewidth}
				\centering
				\includegraphics[width=1.3in]{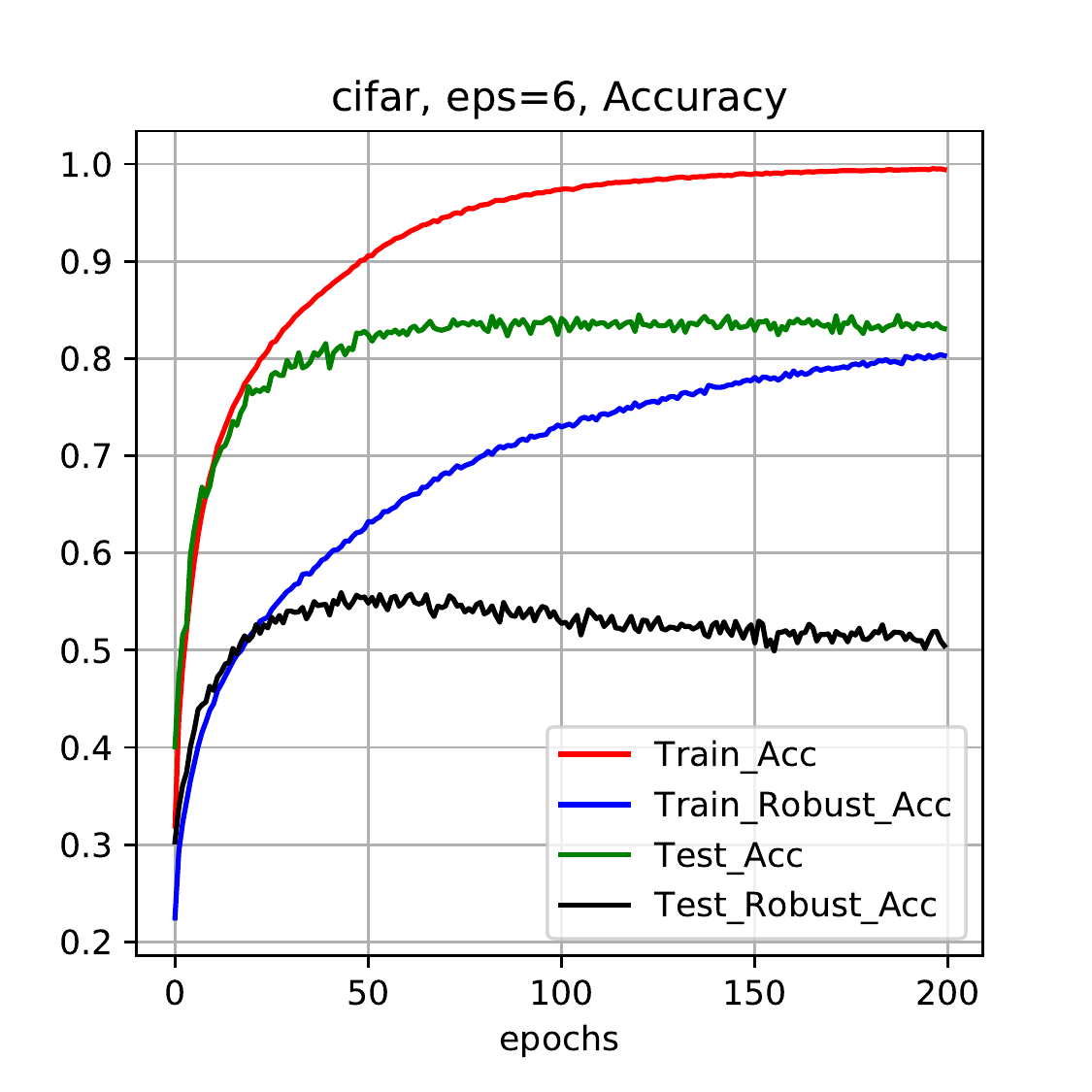}
			\end{minipage}
		}
		\subfigure[]{
			\begin{minipage}[htp]{0.2\linewidth}
				\centering
				\includegraphics[width=1.3in]{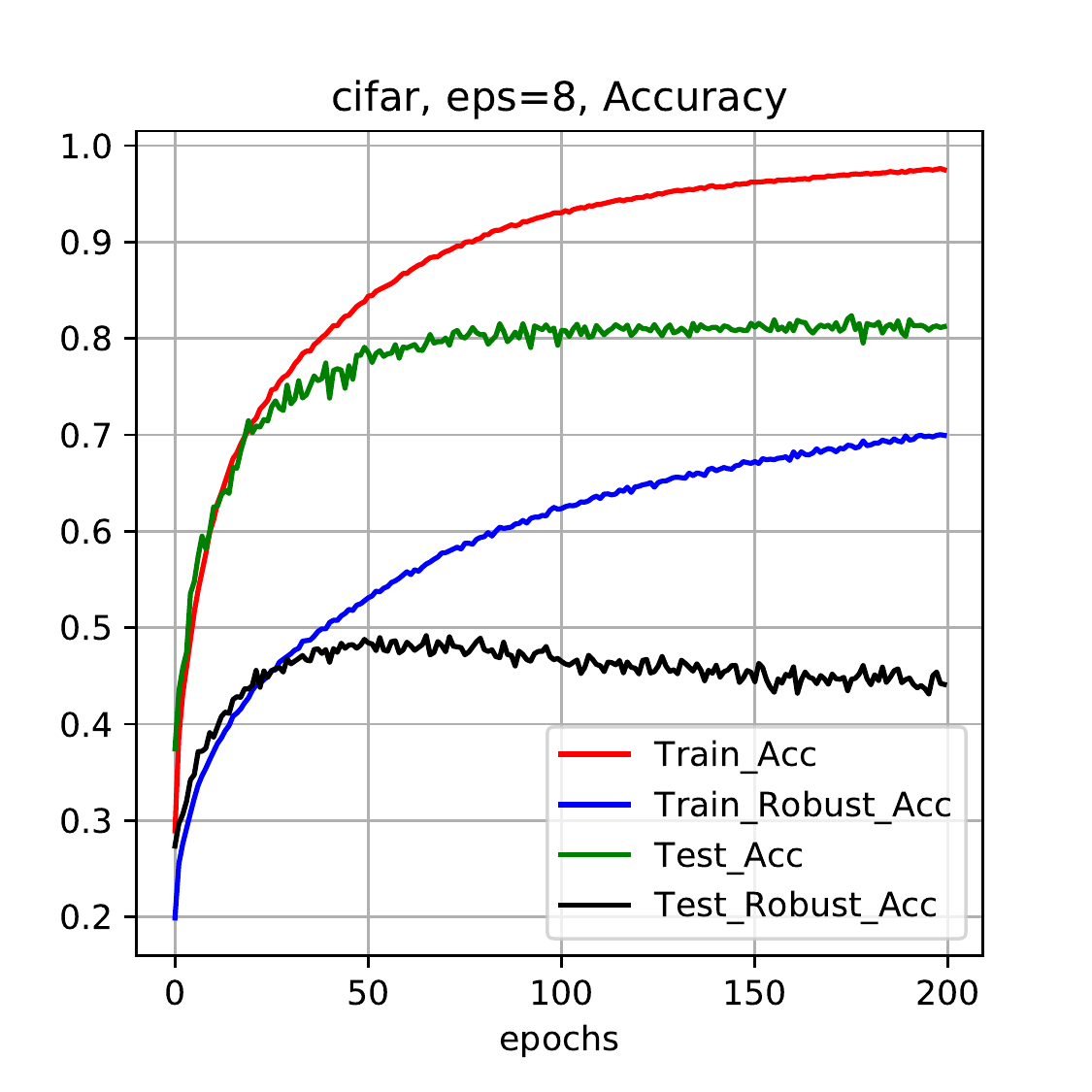}
			\end{minipage}
		}
		\subfigure[]{
			\begin{minipage}[htp]{0.2\linewidth}
				\centering
				\includegraphics[width=1.3in]{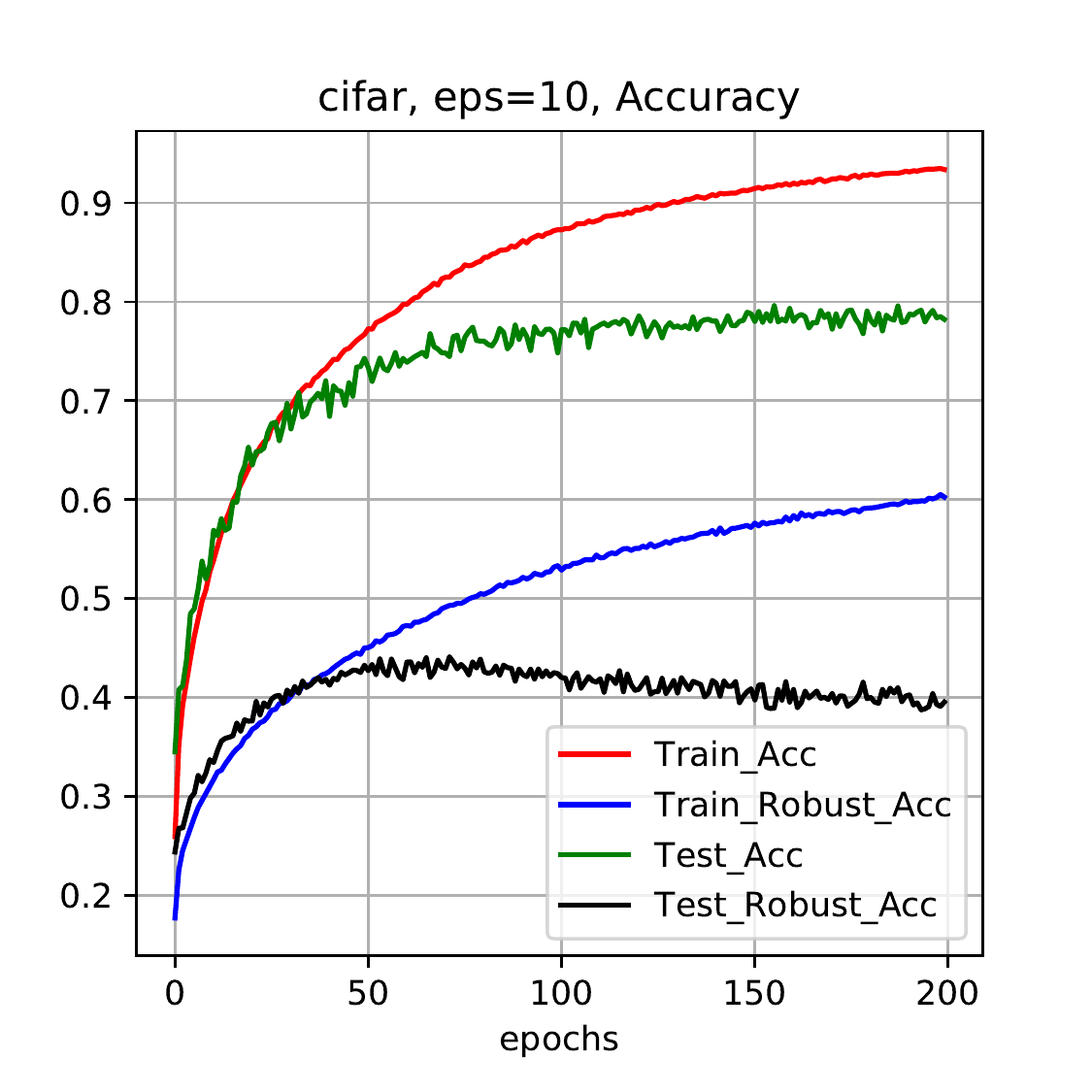}
			\end{minipage}
	}}
	
	\hspace{-0.4in}\scalebox{0.9}{
		\subfigure[]{
			\begin{minipage}[htp]{0.2\linewidth}
				\centering
				\includegraphics[width=1.3in]{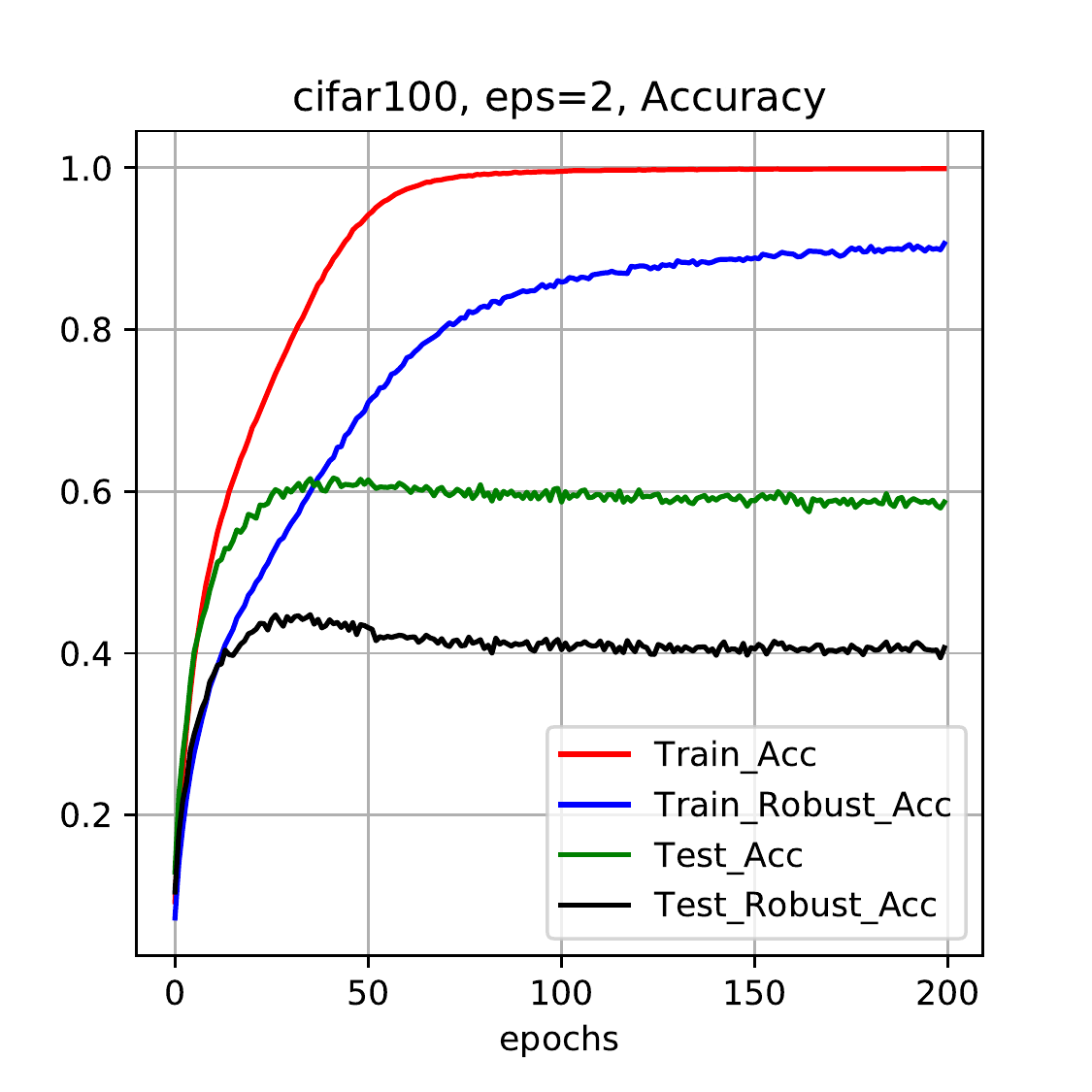}
			\end{minipage}%
		}
		\subfigure[]{
			\begin{minipage}[htp]{0.2\linewidth}
				\centering
				\includegraphics[width=1.3in]{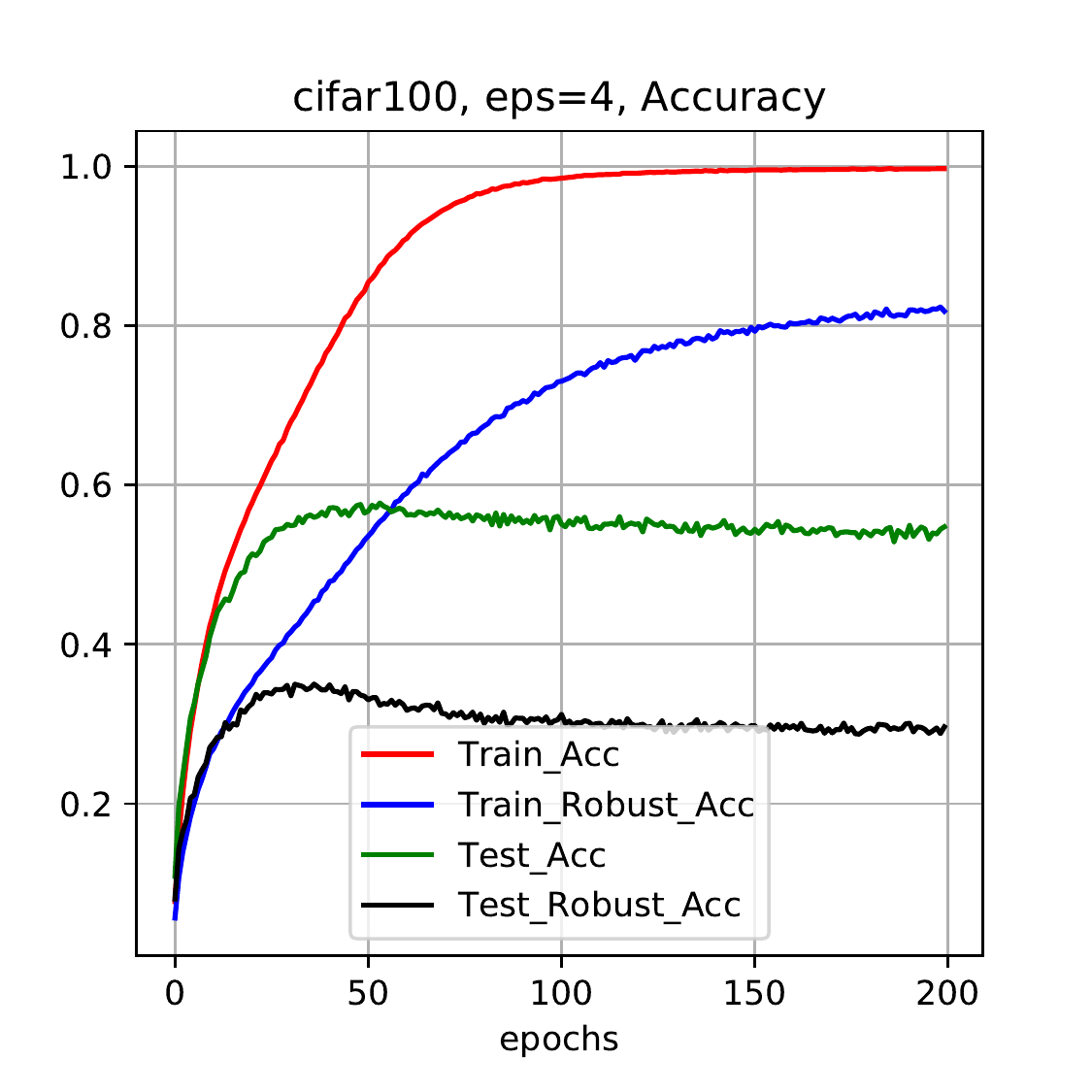}
			\end{minipage}
		}
		\subfigure[]{
			\begin{minipage}[htp]{0.2\linewidth}
				\centering
				\includegraphics[width=1.3in]{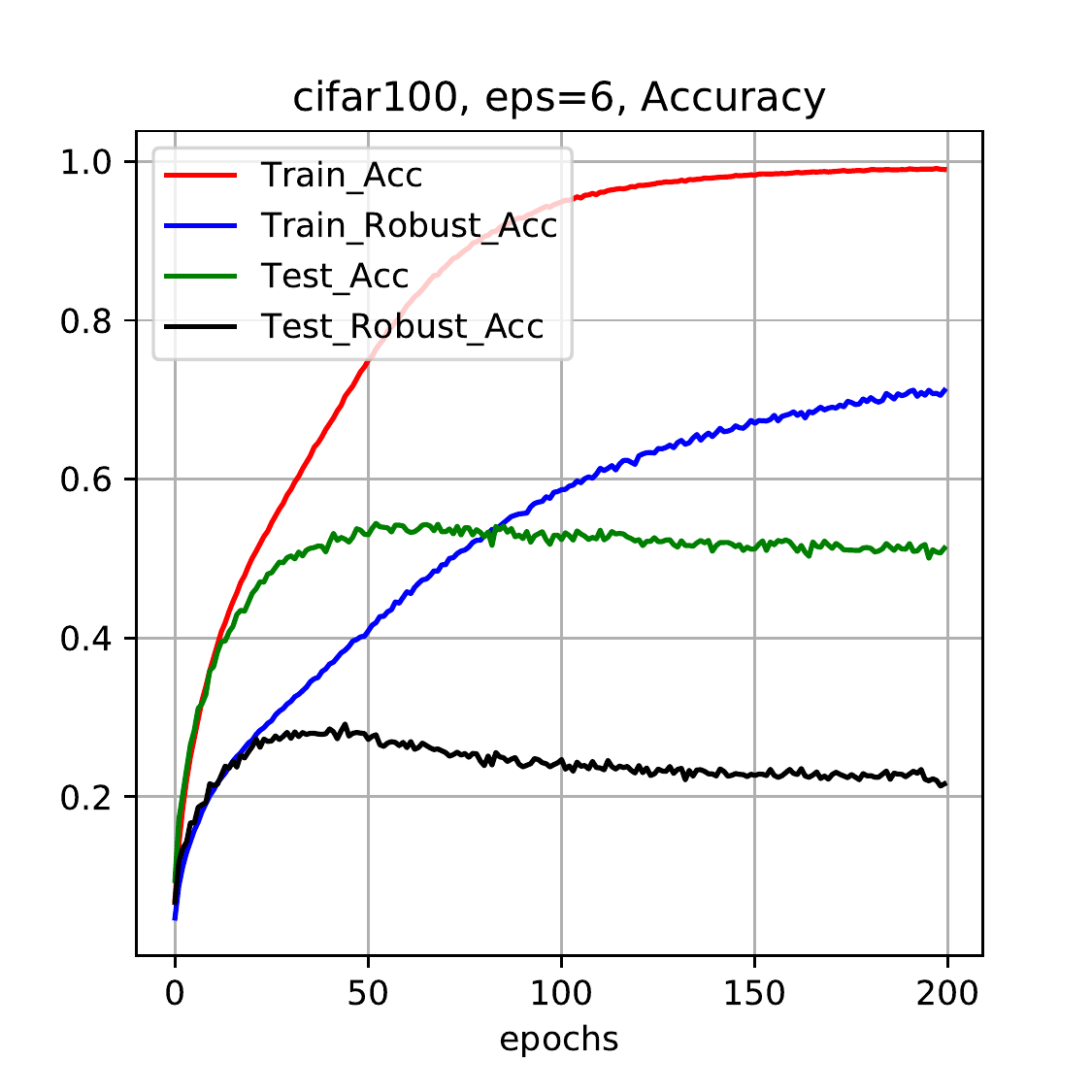}
			\end{minipage}
		}
		\subfigure[]{
			\begin{minipage}[htp]{0.2\linewidth}
				\centering
				\includegraphics[width=1.3in]{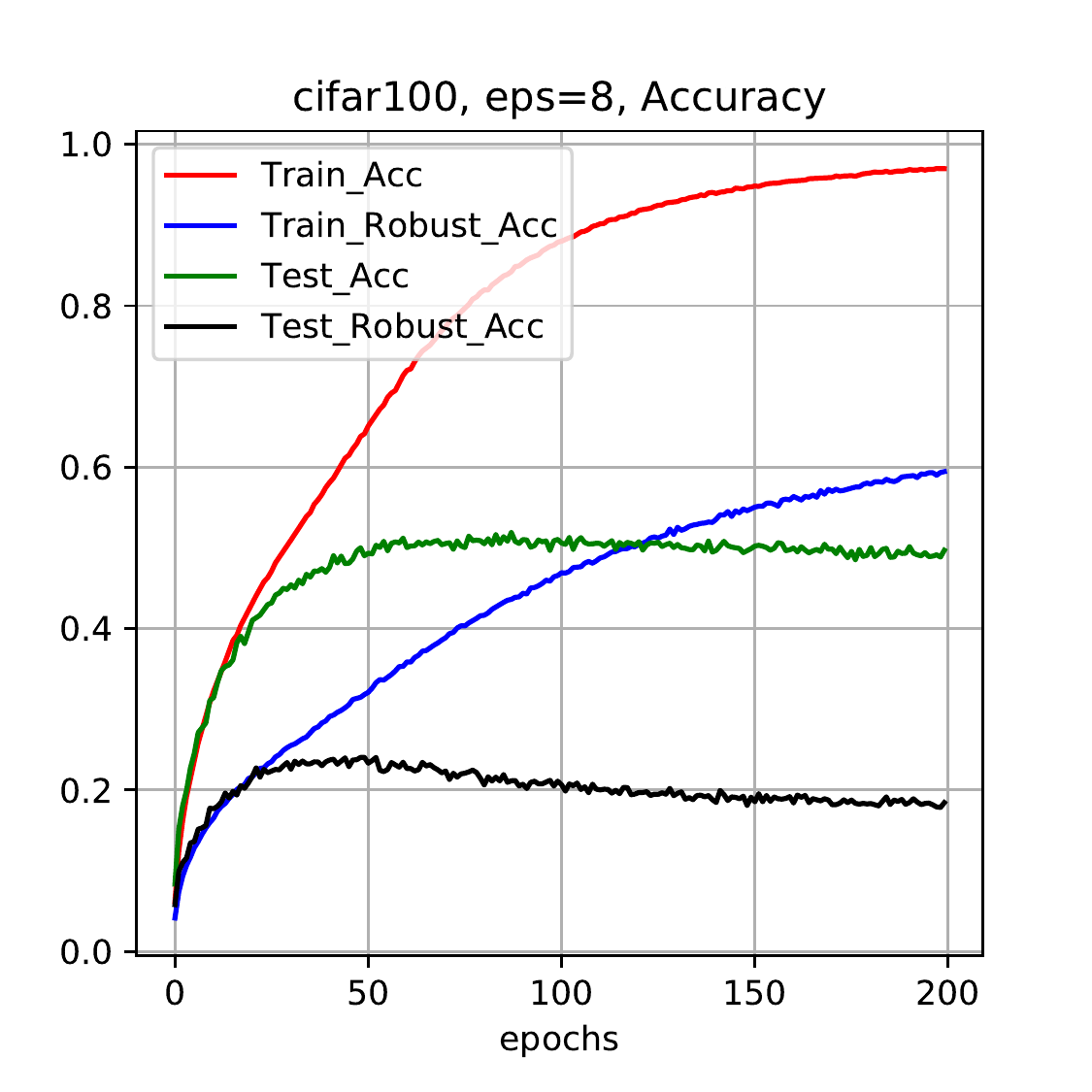}
			\end{minipage}
		}
		\subfigure[]{
			\begin{minipage}[htp]{0.2\linewidth}
				\centering
				\includegraphics[width=1.3in]{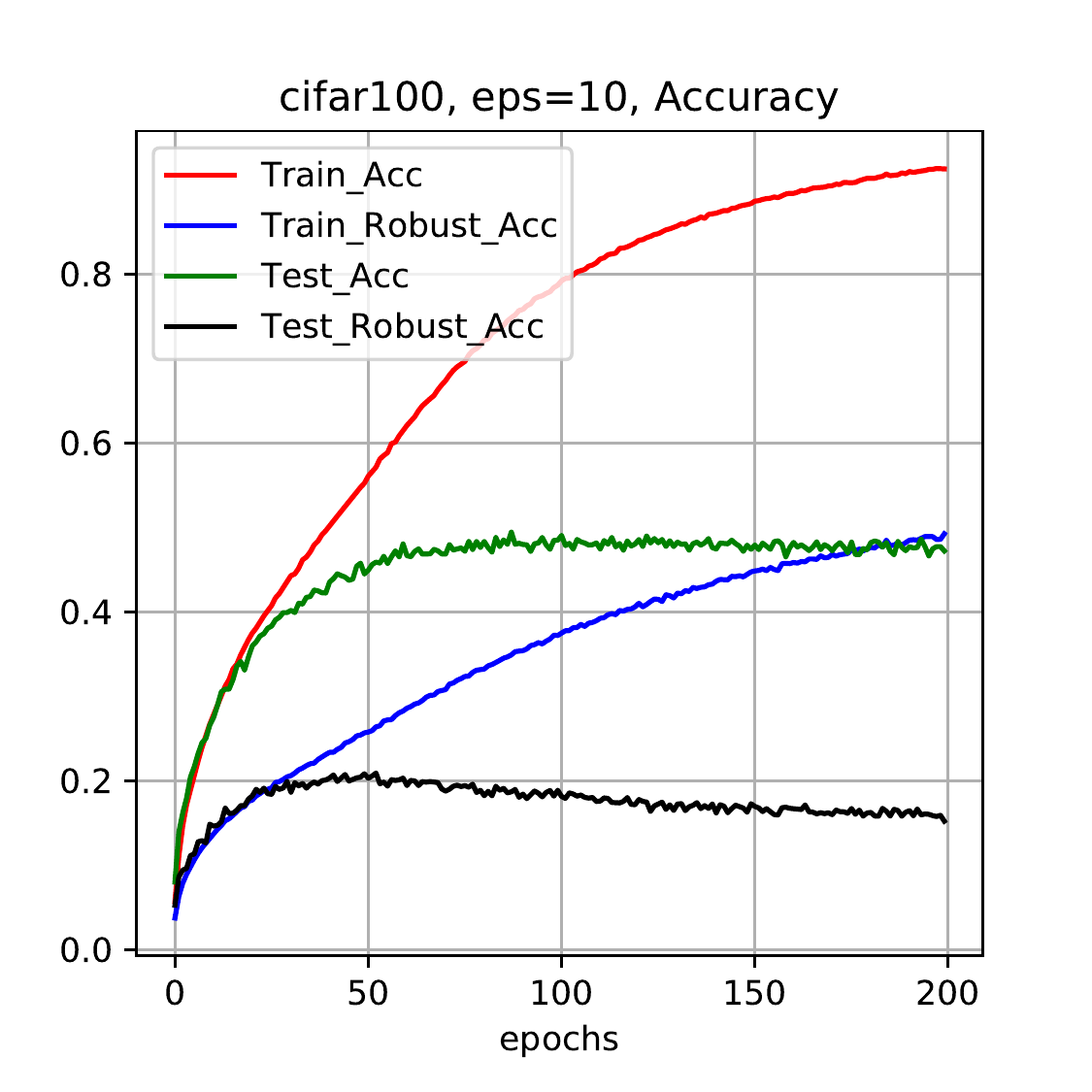}
			\end{minipage}
	}}
	
	\centering
	\vskip -0.1in
	\caption{Accuracy of adversarial training with fixed learning rate = 0.01. The first row is the experiments on SVHN. The second row is the experiments on CIFAR-10. The last row is the experiments on CIFAR-100. The first column to the last column are the experiments of $\epsilon$ equal to 2, 4, 6, 8, and 10, respectively.}
	\label{fig:fixed}
	\vskip -0.1in
\end{figure*}

\begin{figure*}[htbp]
	\centering
	\hspace{-0.4in}\scalebox{0.9}{
		\subfigure[]{
			\begin{minipage}[htp]{0.2\linewidth}
				\centering
				\includegraphics[width=1.3in]{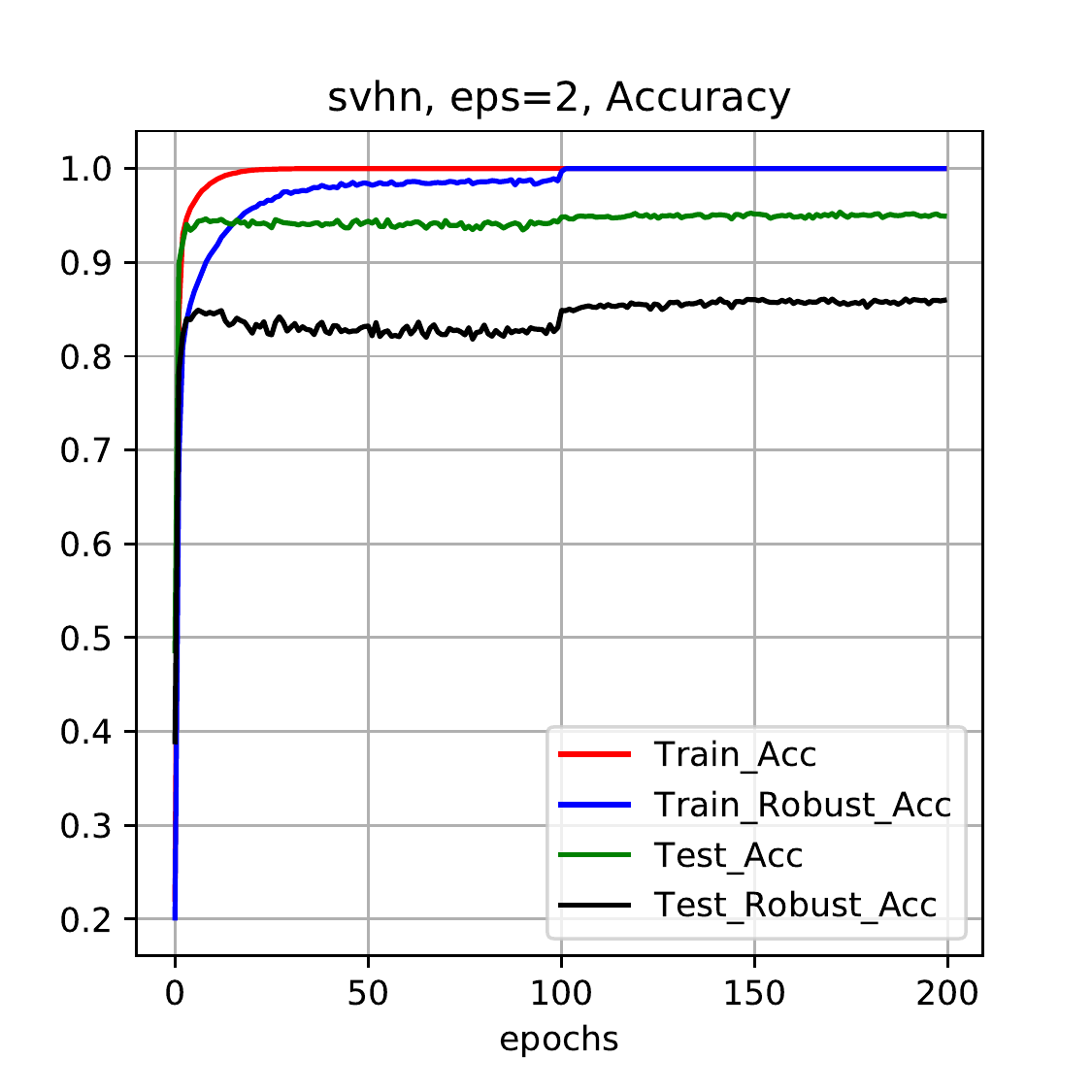}
			\end{minipage}%
		}
		\subfigure[]{
			\begin{minipage}[htp]{0.2\linewidth}
				\centering
				\includegraphics[width=1.3in]{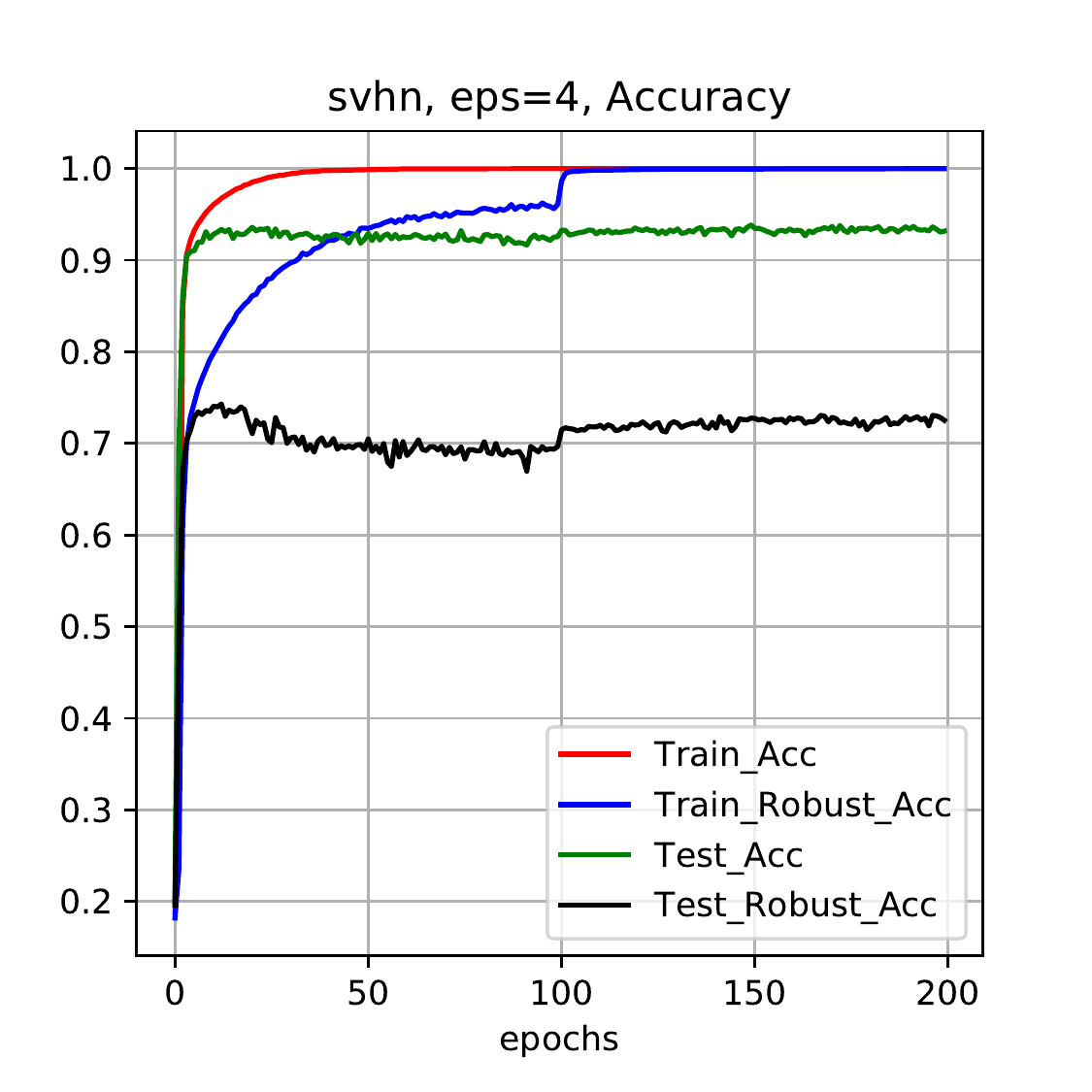}
			\end{minipage}
		}
		\subfigure[]{
			\begin{minipage}[htp]{0.2\linewidth}
				\centering
				\includegraphics[width=1.3in]{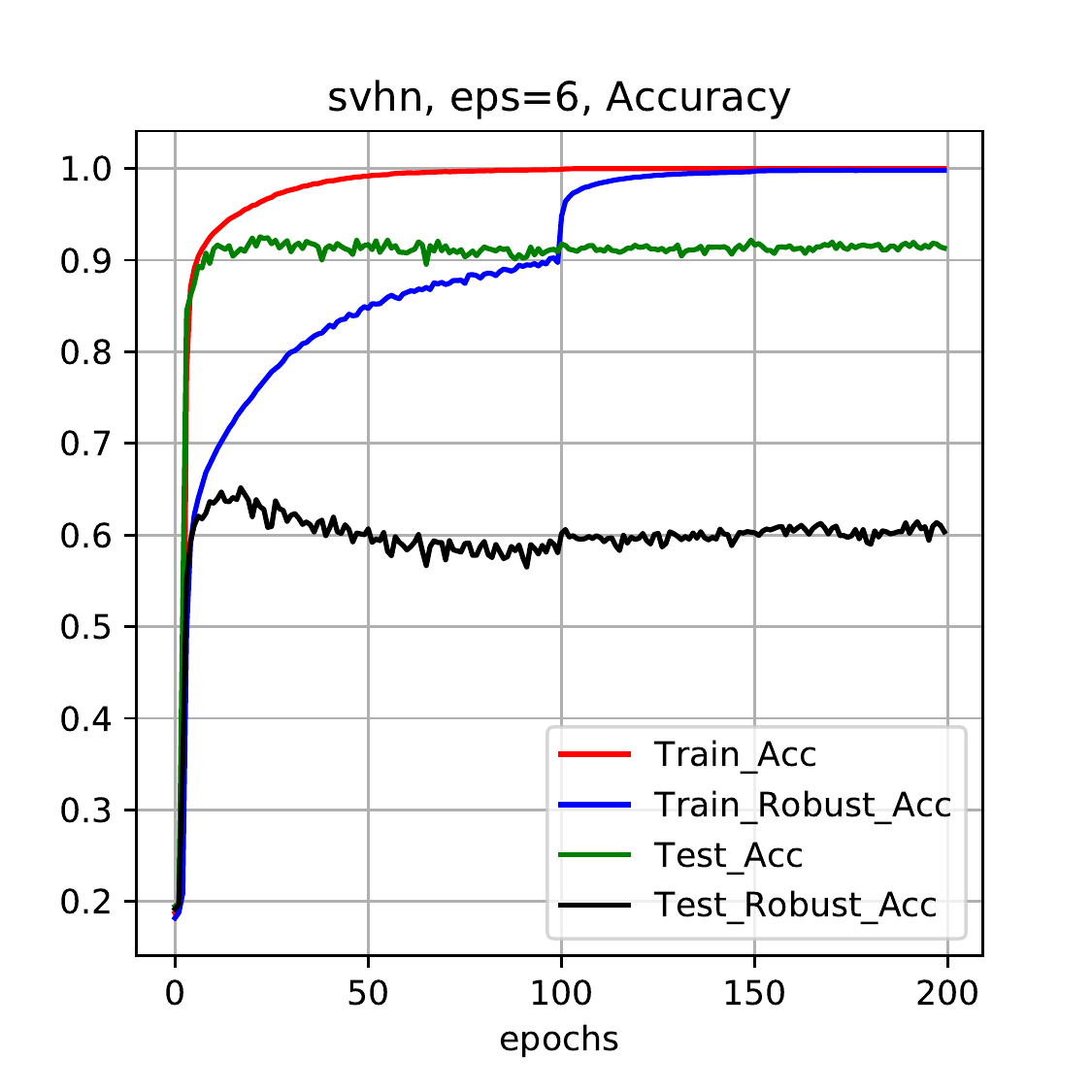}
			\end{minipage}
		}
		\subfigure[]{
			\begin{minipage}[htp]{0.2\linewidth}
				\centering
				\includegraphics[width=1.3in]{./figures/piecewisesvhn_eps8.pdf}
			\end{minipage}
		}
		\subfigure[]{
			\begin{minipage}[htp]{0.2\linewidth}
				\centering
				\includegraphics[width=1.3in]{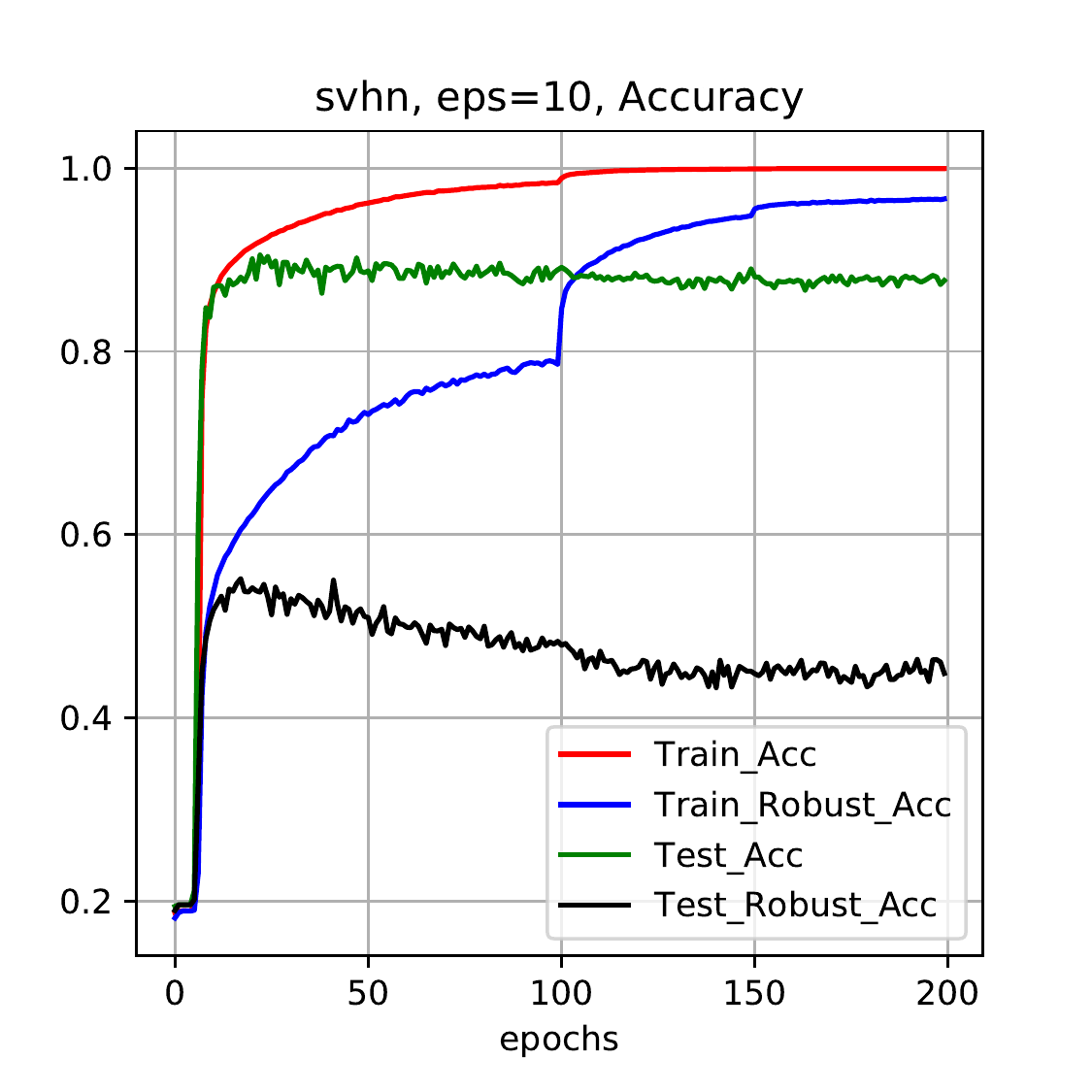}
			\end{minipage}
	}}

	\hspace{-0.4in}\scalebox{0.9}{
		\subfigure[]{
			\begin{minipage}[htp]{0.2\linewidth}
				\centering
				\includegraphics[width=1.3in]{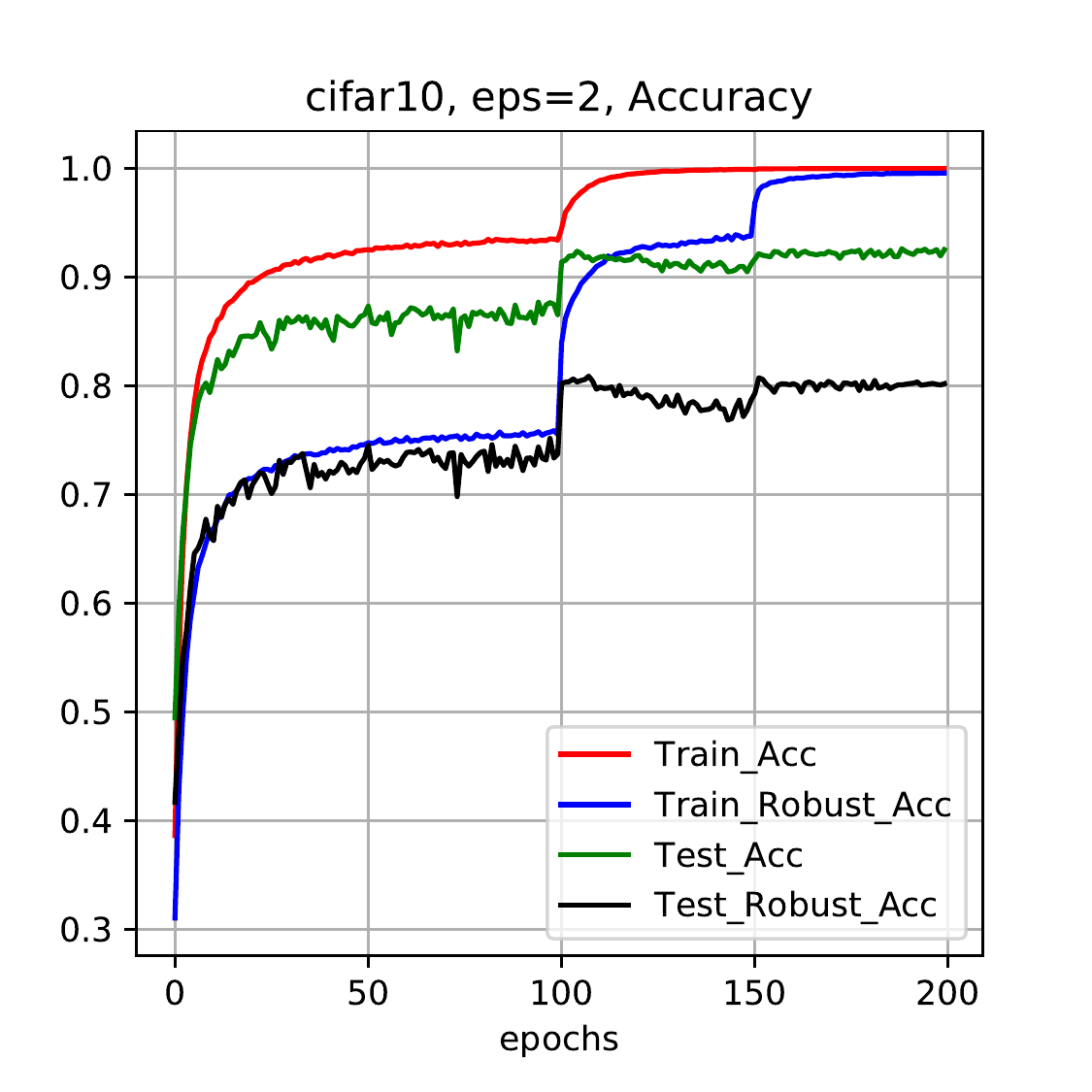}
			\end{minipage}%
		}
		\subfigure[]{
			\begin{minipage}[htp]{0.2\linewidth}
				\centering
				\includegraphics[width=1.3in]{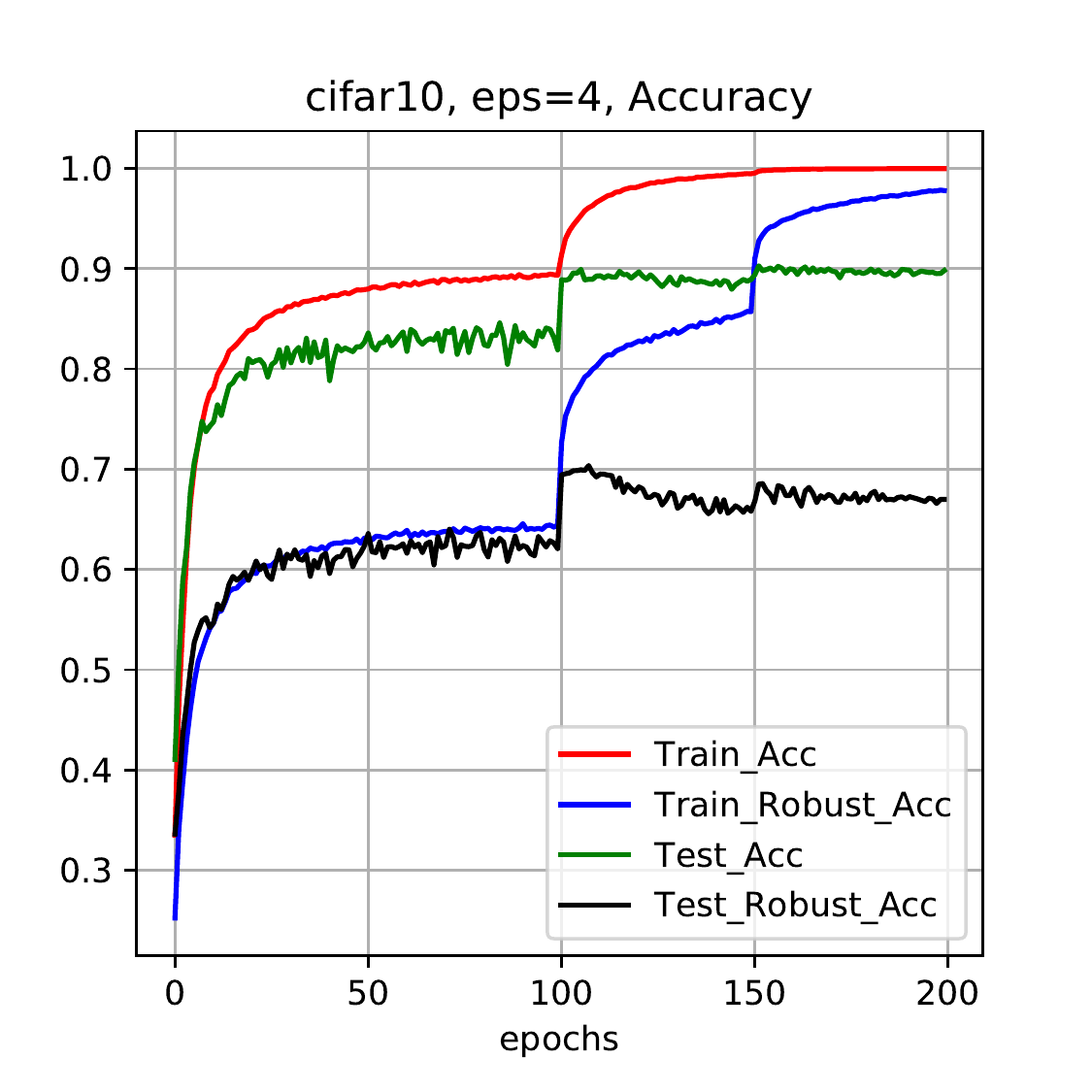}
			\end{minipage}
		}
		\subfigure[]{
			\begin{minipage}[htp]{0.2\linewidth}
				\centering
				\includegraphics[width=1.3in]{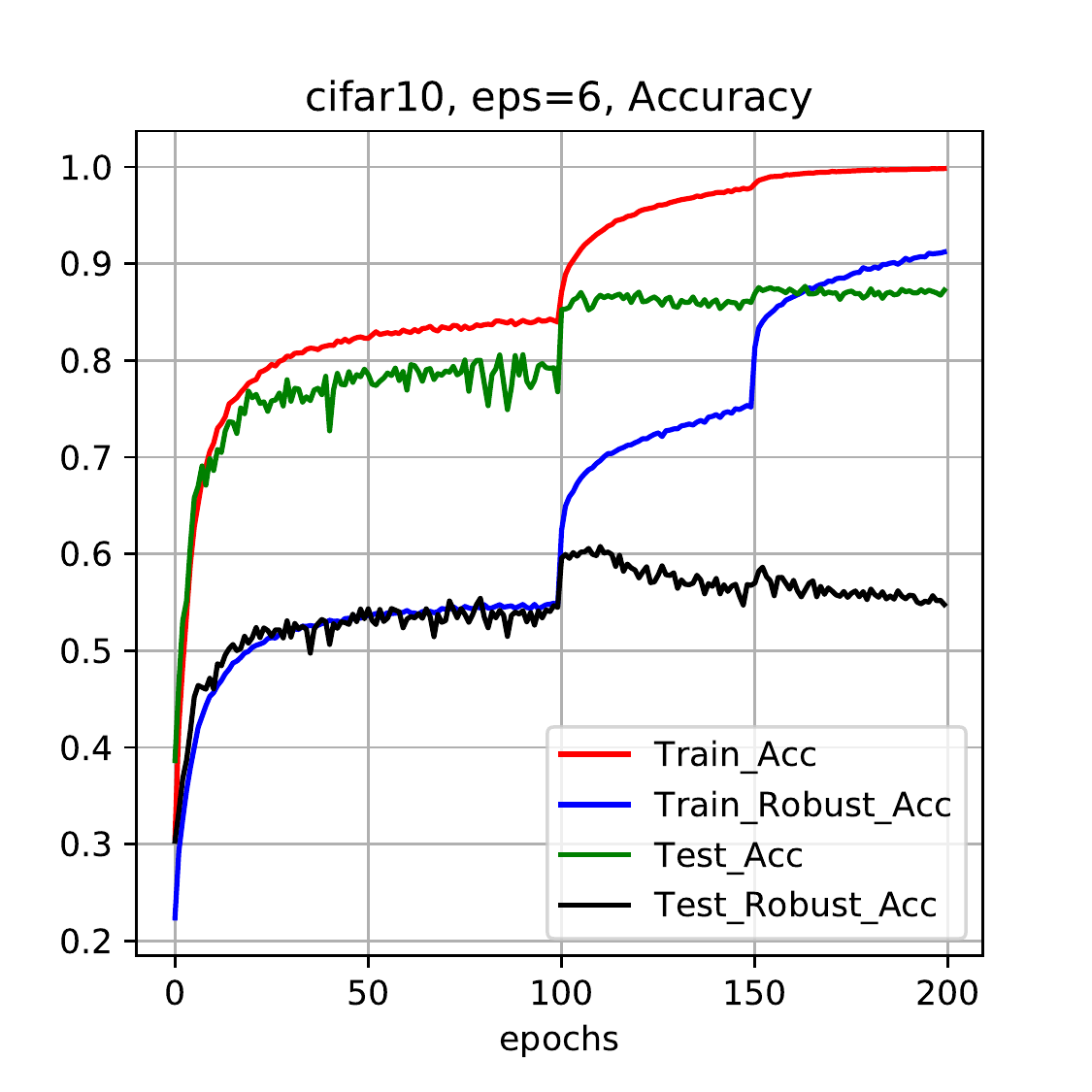}
			\end{minipage}
		}
		\subfigure[]{
			\begin{minipage}[htp]{0.2\linewidth}
				\centering
				\includegraphics[width=1.3in]{./figures/cifar_eps8.pdf}
			\end{minipage}
		}
		\subfigure[]{
			\begin{minipage}[htp]{0.2\linewidth}
				\centering
				\includegraphics[width=1.3in]{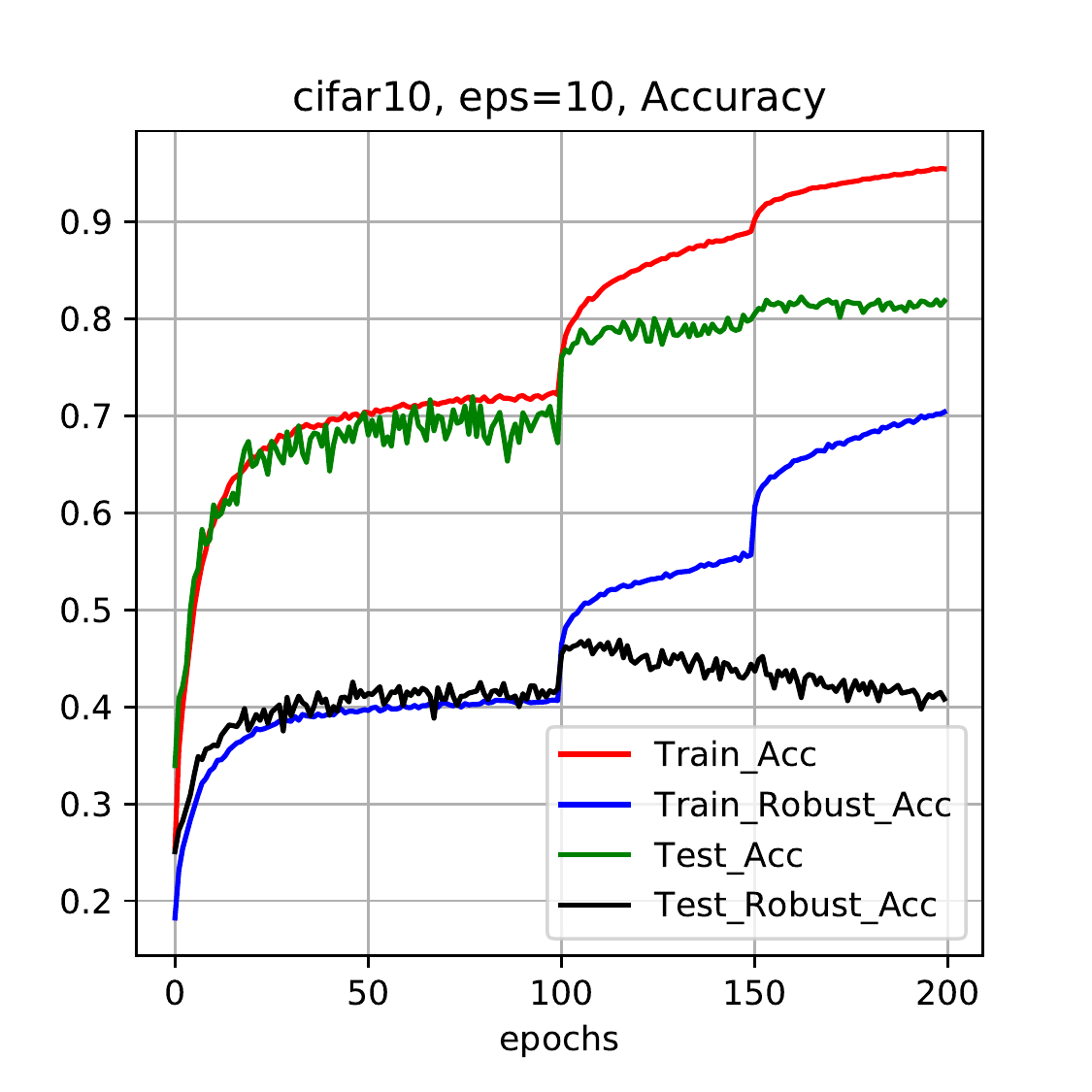}
			\end{minipage}
	}}

	\hspace{-0.4in}\scalebox{0.9}{
		\subfigure[]{
			\begin{minipage}[htp]{0.2\linewidth}
				\centering
				\includegraphics[width=1.3in]{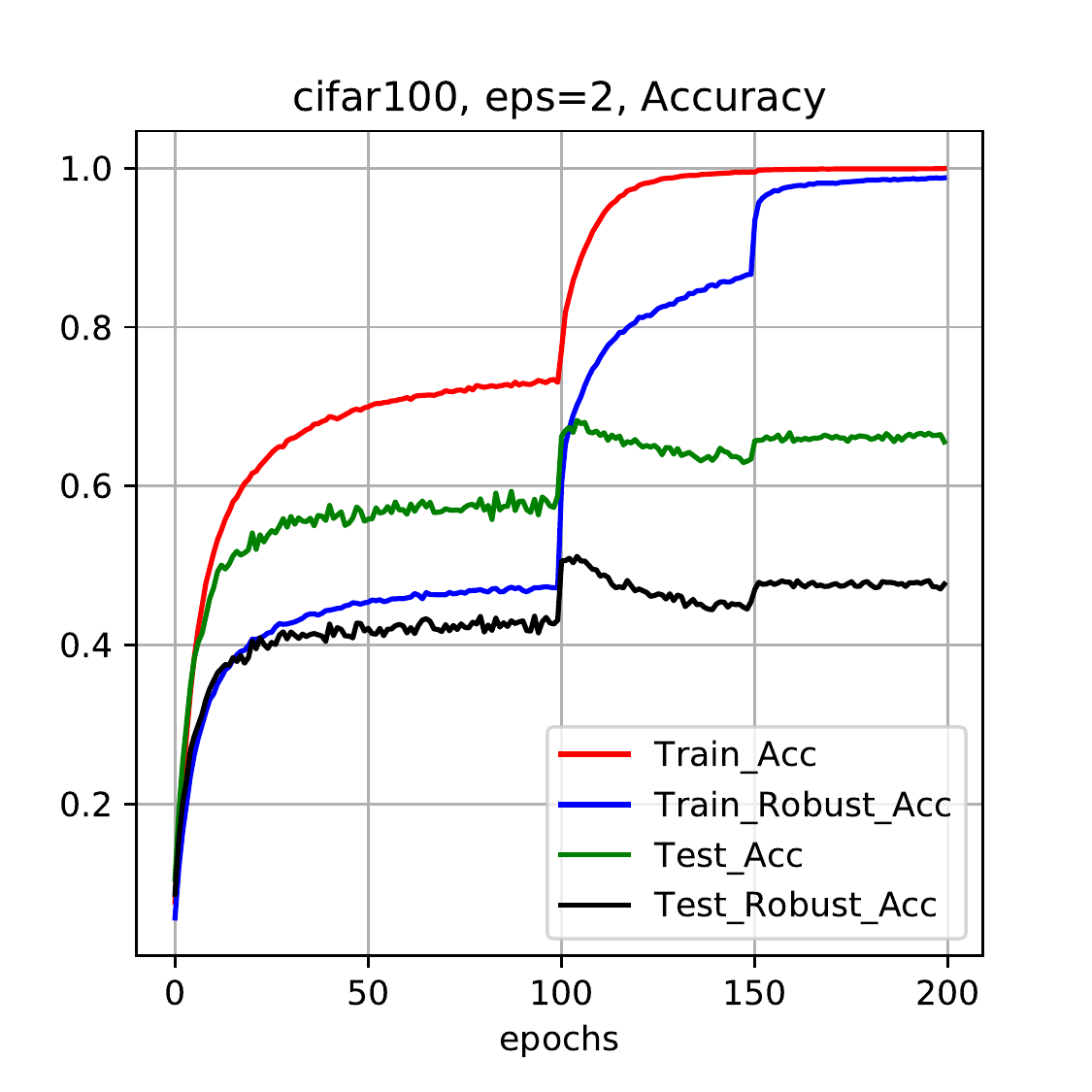}
			\end{minipage}%
		}
		\subfigure[]{
			\begin{minipage}[htp]{0.2\linewidth}
				\centering
				\includegraphics[width=1.3in]{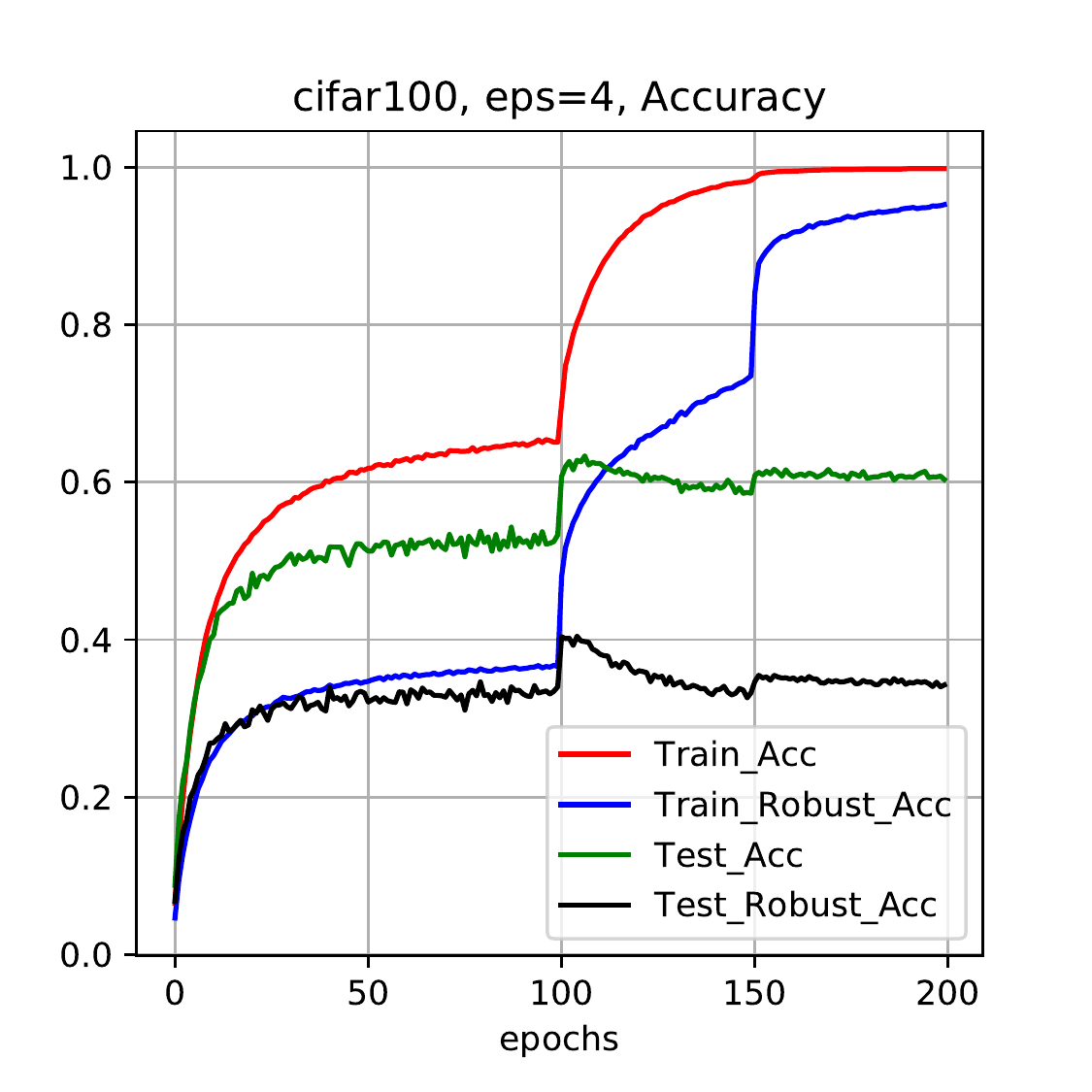}
			\end{minipage}
		}
		\subfigure[]{
			\begin{minipage}[htp]{0.2\linewidth}
				\centering
				\includegraphics[width=1.3in]{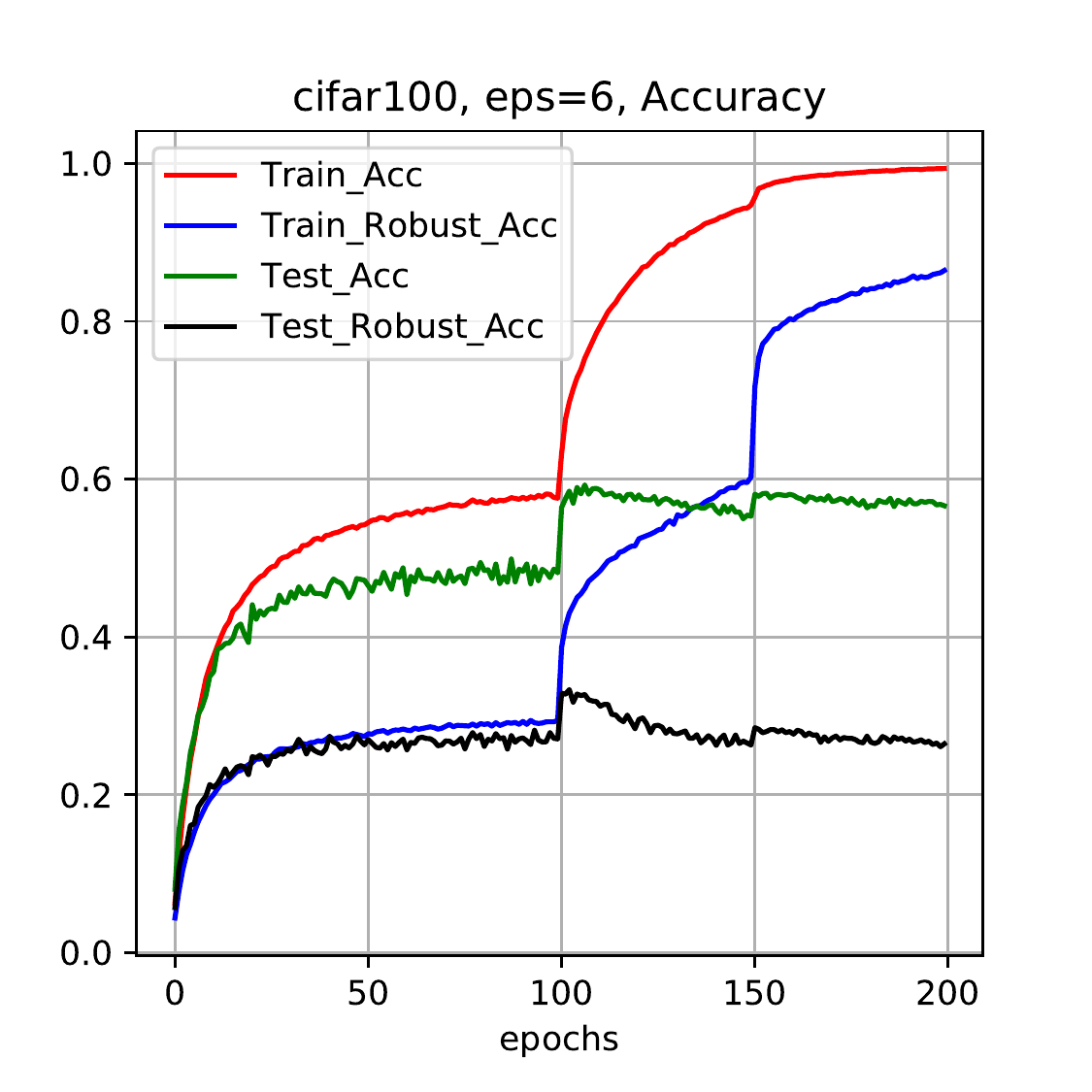}
			\end{minipage}
		}
		\subfigure[]{
			\begin{minipage}[htp]{0.2\linewidth}
				\centering
				\includegraphics[width=1.3in]{./figures/cifar100_eps8.pdf}
			\end{minipage}
		}
		\subfigure[]{
			\begin{minipage}[htp]{0.2\linewidth}
				\centering
				\includegraphics[width=1.3in]{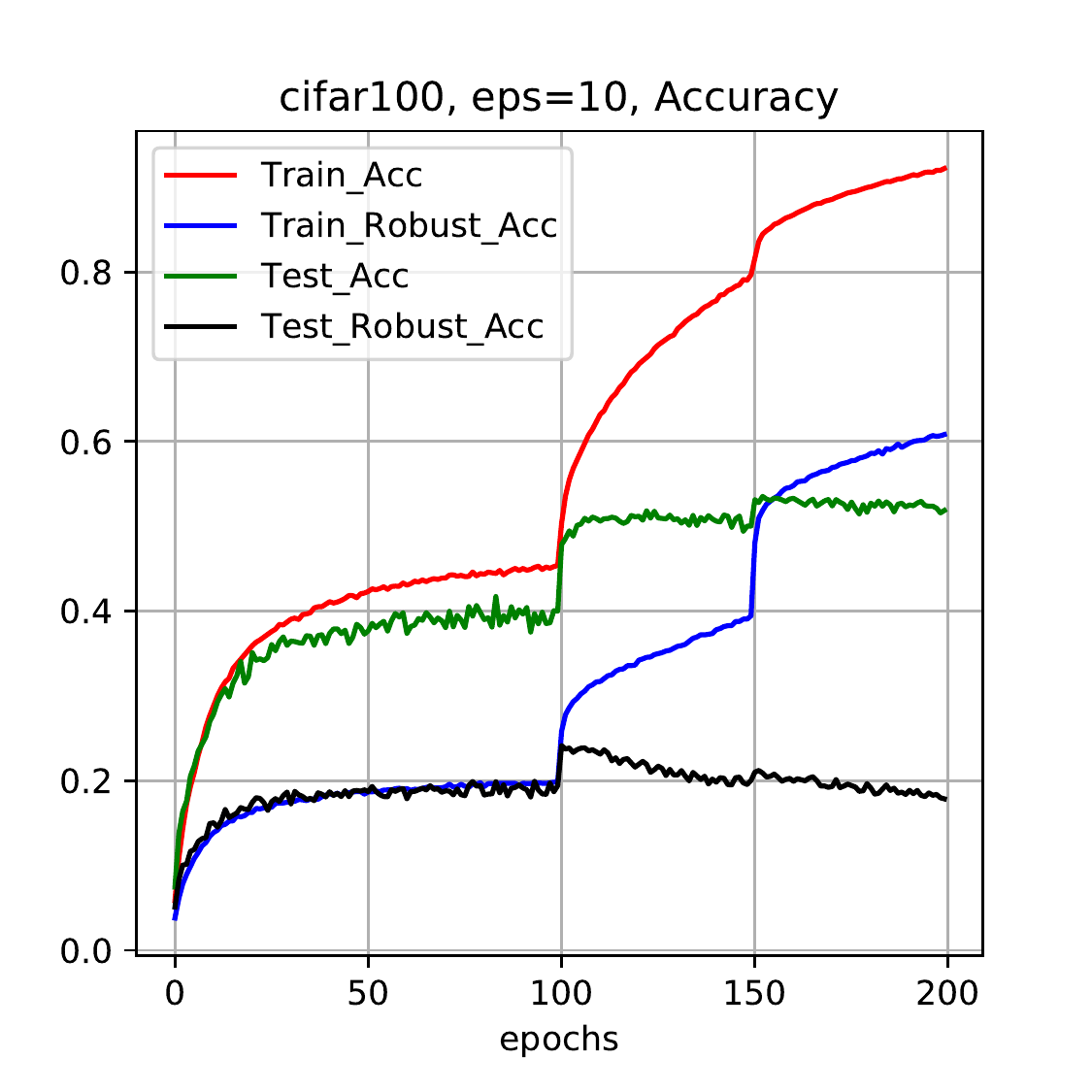}
			\end{minipage}
	}}

	\caption{Accuracy of adversarial training with piece-wise linear learning rate. The first row is the experiments on SVHN. The second row is the experiments on CIFAR-10. The last row is the experiments on CIFAR-100. The first column to the last column are the experiments of $\epsilon$ equal to 2, 4, 6, 8, and 10, respectively.}
	\label{fig:piece}
\end{figure*}

\begin{figure*}[htbp]
\centering
\hspace{-0.4in}\scalebox{0.9}{
\subfigure[]{
\begin{minipage}[htp]{0.2\linewidth}
\centering
\includegraphics[width=1.3in]{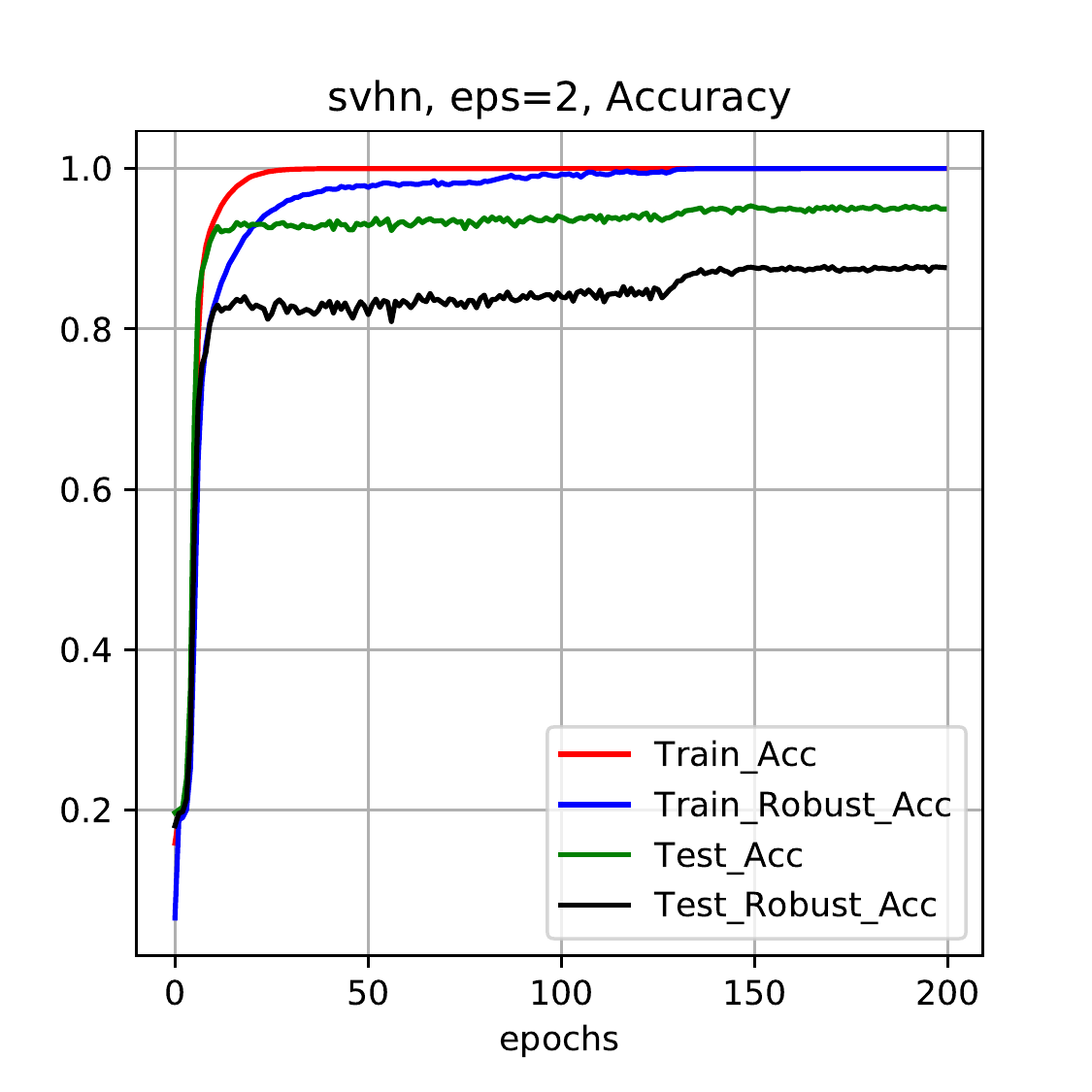}
\end{minipage}%
}
\subfigure[]{
\begin{minipage}[htp]{0.2\linewidth}
\centering
\includegraphics[width=1.3in]{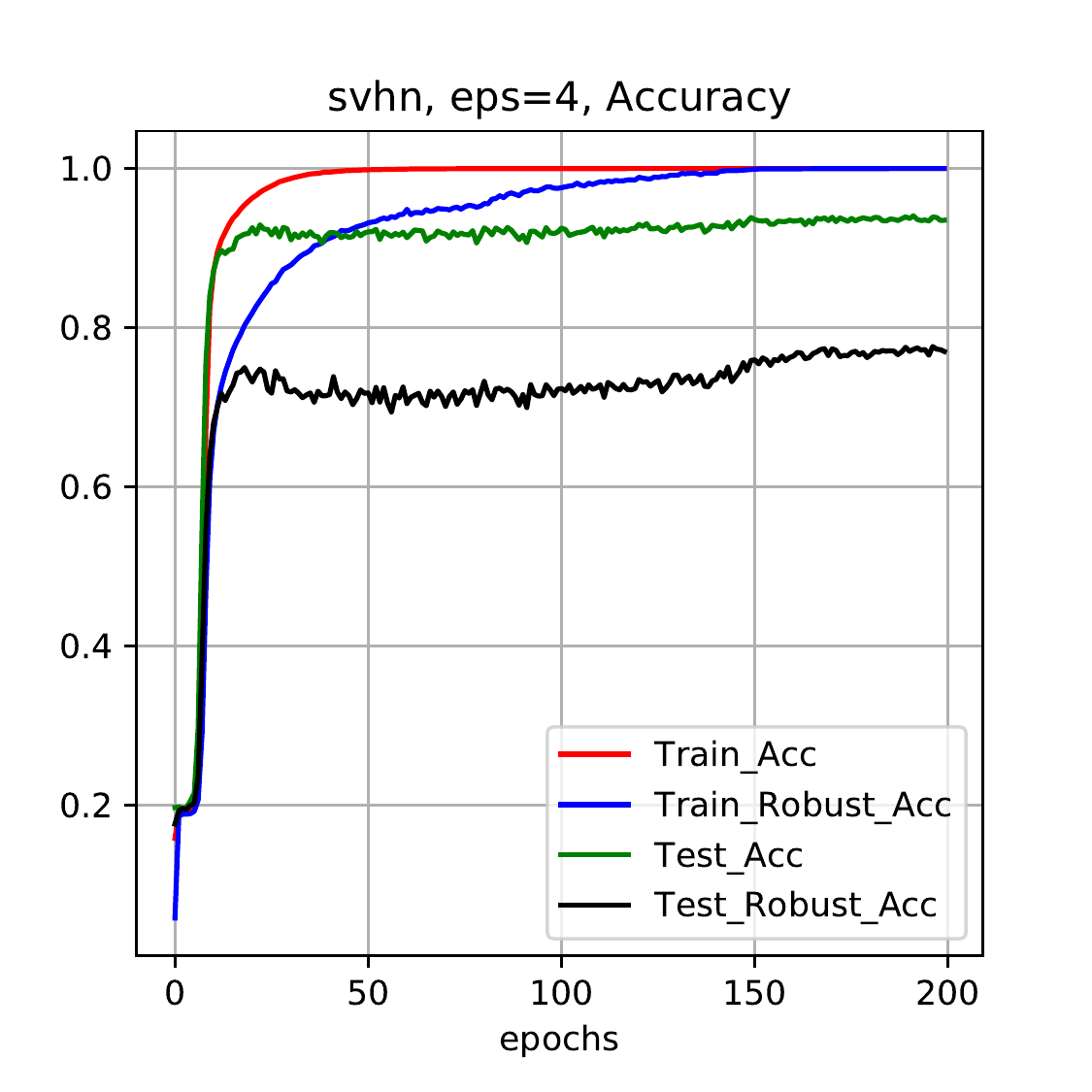}
\end{minipage}
}
\subfigure[]{
\begin{minipage}[htp]{0.2\linewidth}
\centering
\includegraphics[width=1.3in]{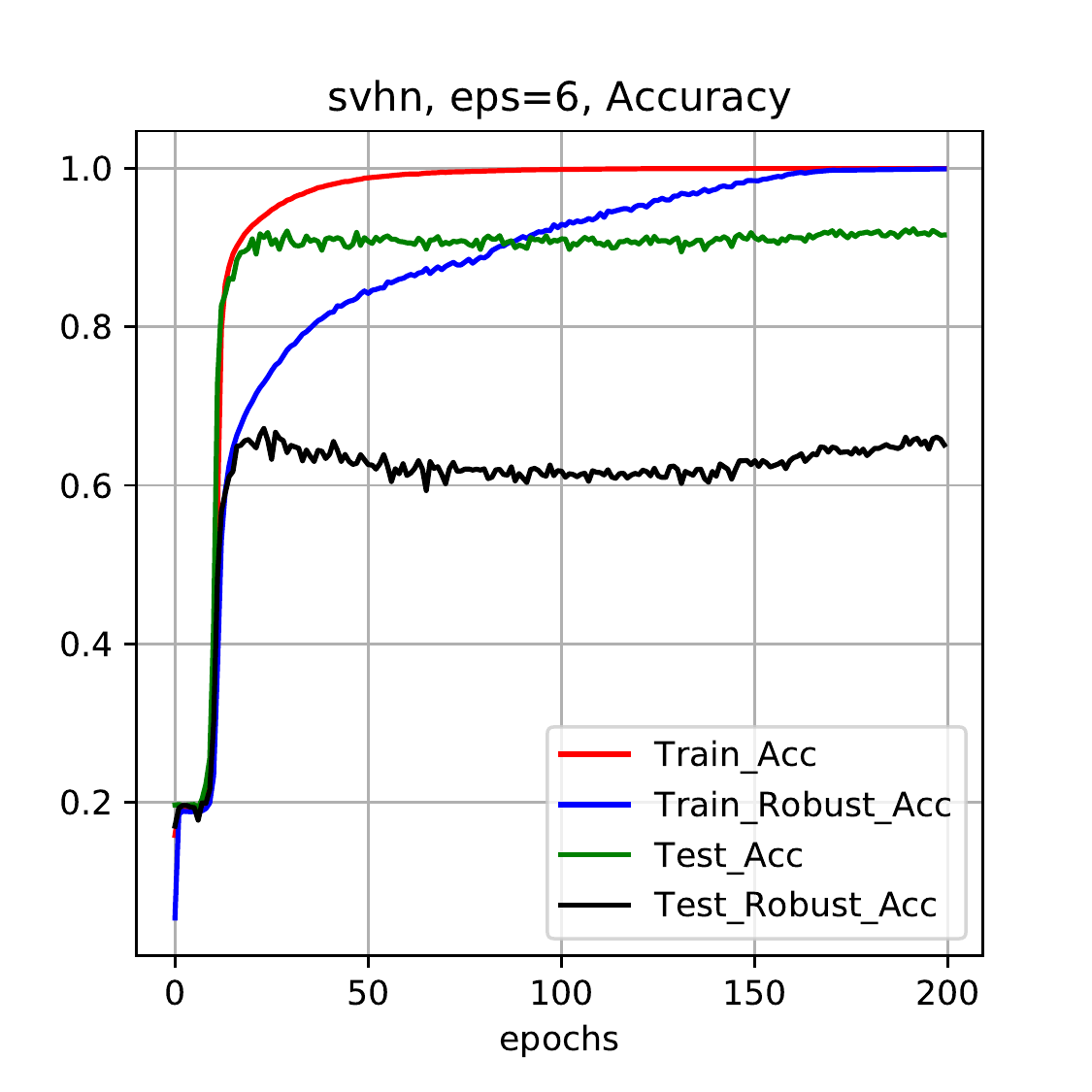}
\end{minipage}
}
\subfigure[]{
\begin{minipage}[htp]{0.2\linewidth}
\centering
\includegraphics[width=1.3in]{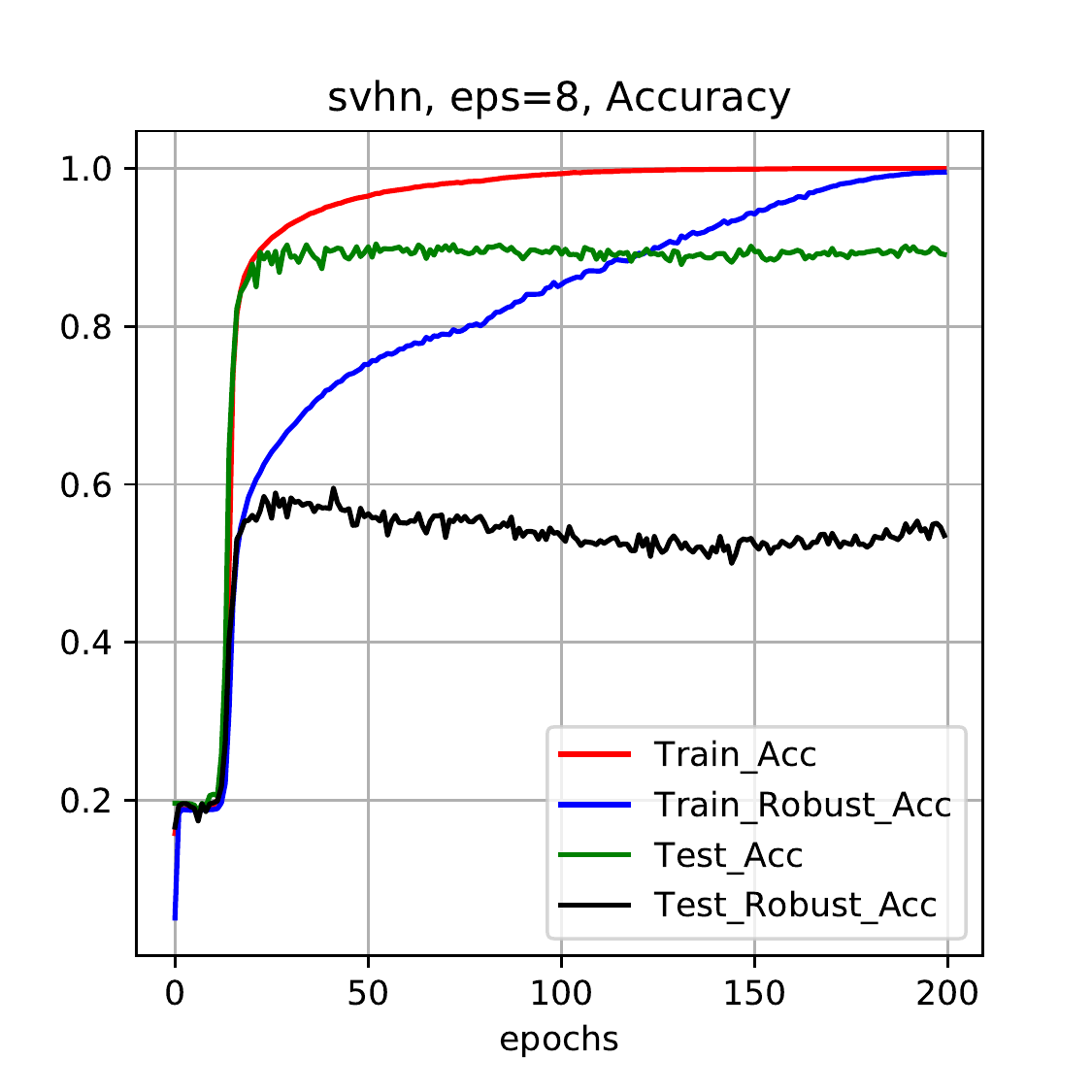}
\end{minipage}
}
\subfigure[]{
\begin{minipage}[htp]{0.2\linewidth}
\centering
\includegraphics[width=1.3in]{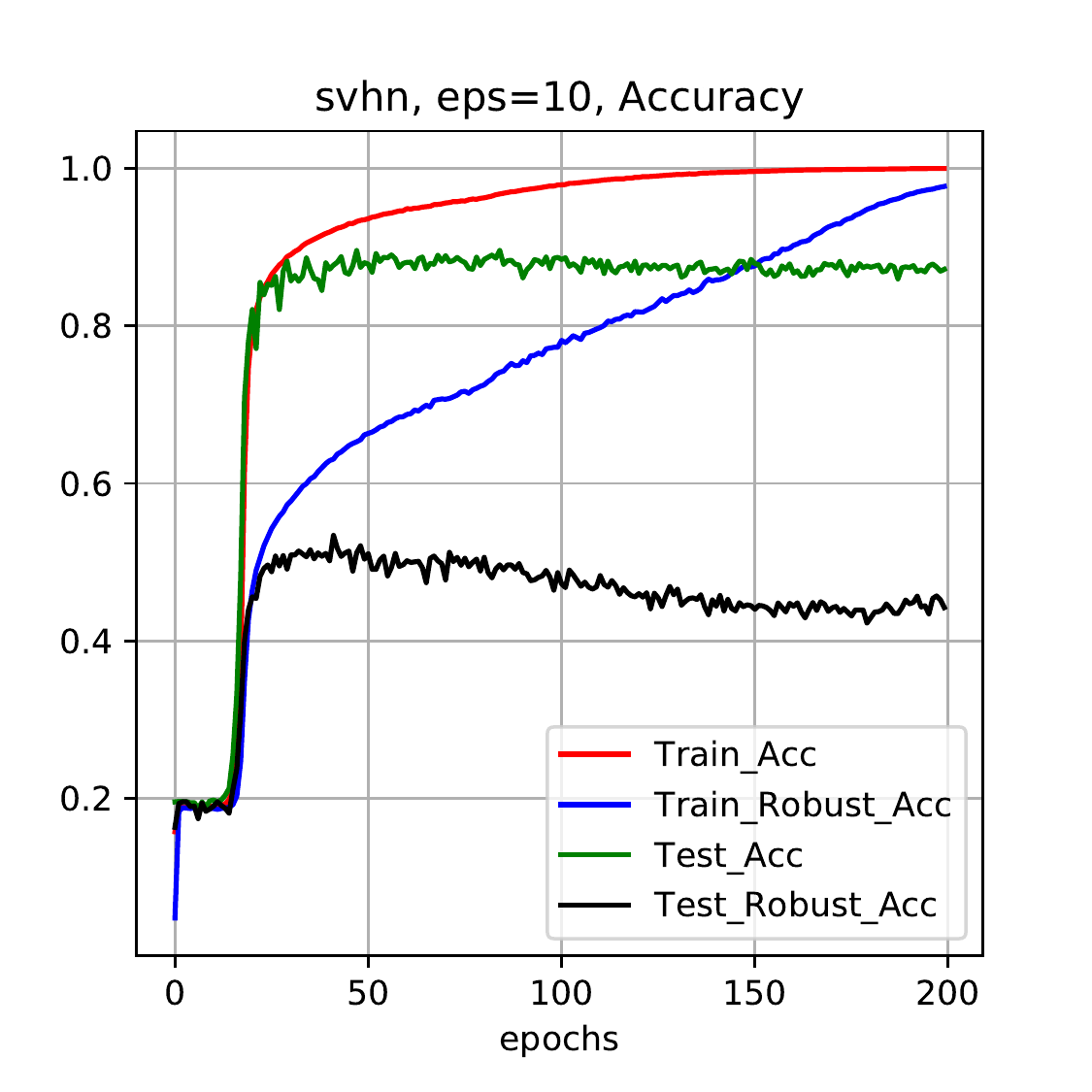}
\end{minipage}
}}

\hspace{-0.4in}\scalebox{0.9}{
\subfigure[]{
\begin{minipage}[htp]{0.2\linewidth}
\centering
\includegraphics[width=1.3in]{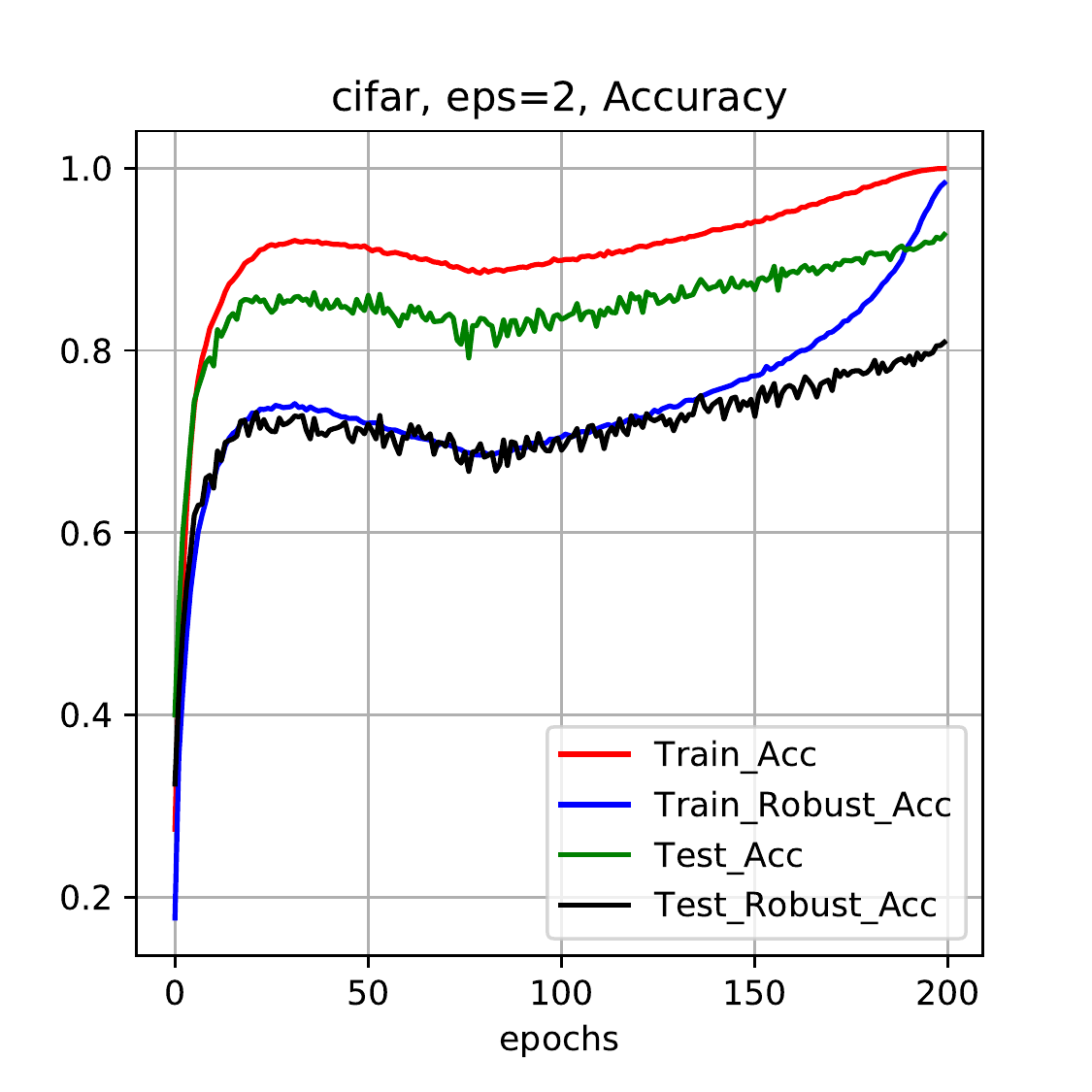}
\end{minipage}%
}
\subfigure[]{
\begin{minipage}[htp]{0.2\linewidth}
\centering
\includegraphics[width=1.3in]{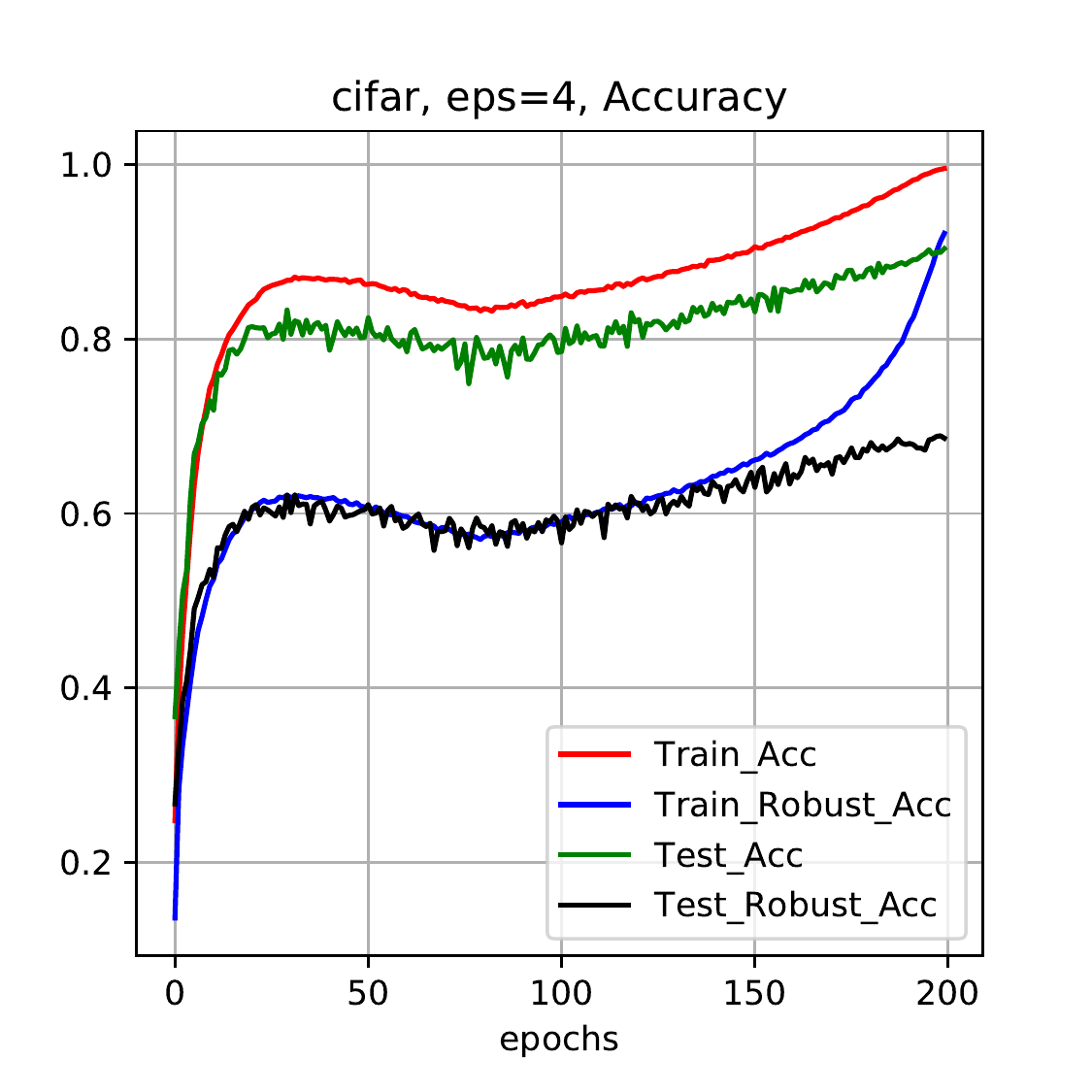}
\end{minipage}
}
\subfigure[]{
\begin{minipage}[htp]{0.2\linewidth}
\centering
\includegraphics[width=1.3in]{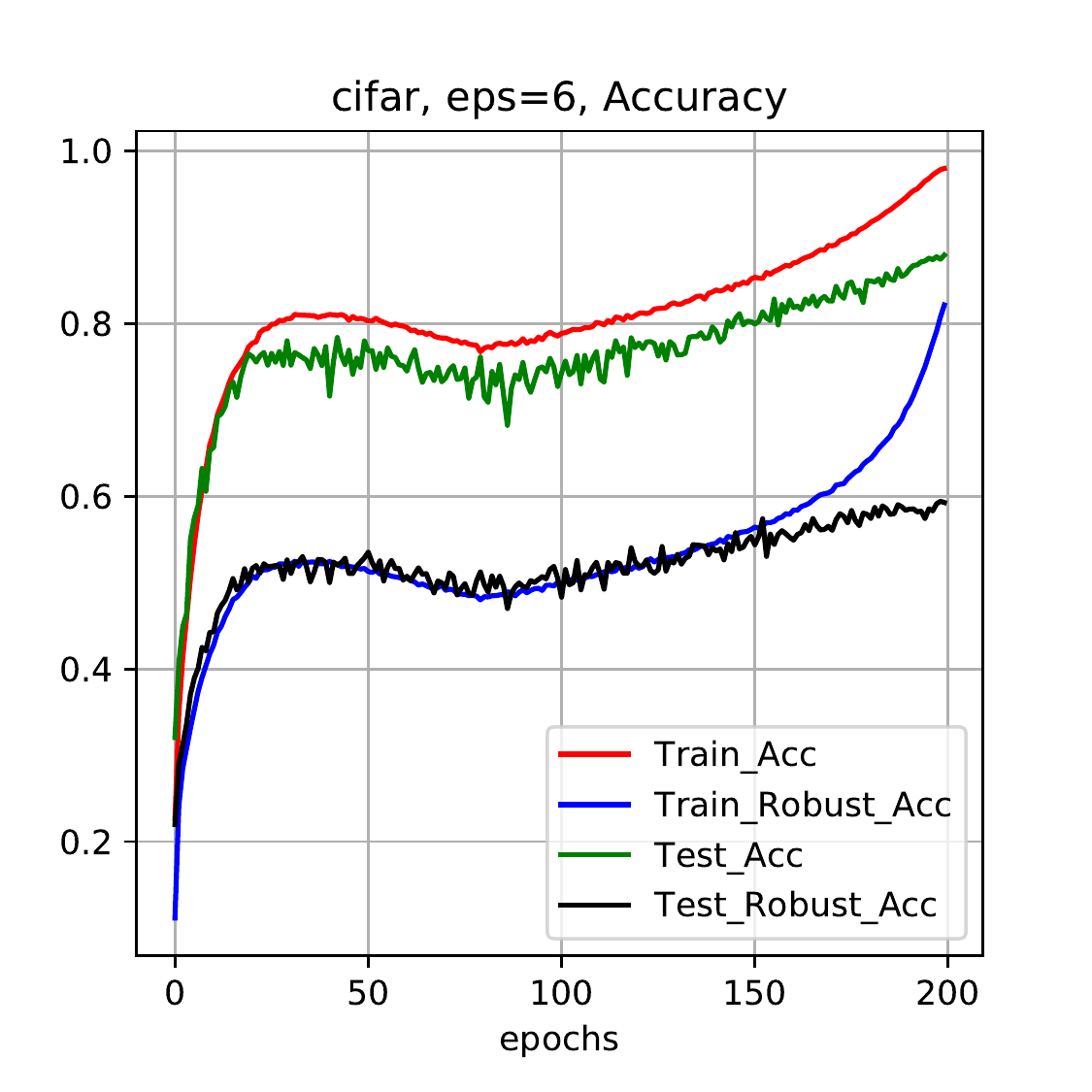}
\end{minipage}
}
\subfigure[]{
\begin{minipage}[htp]{0.2\linewidth}
\centering
\includegraphics[width=1.3in]{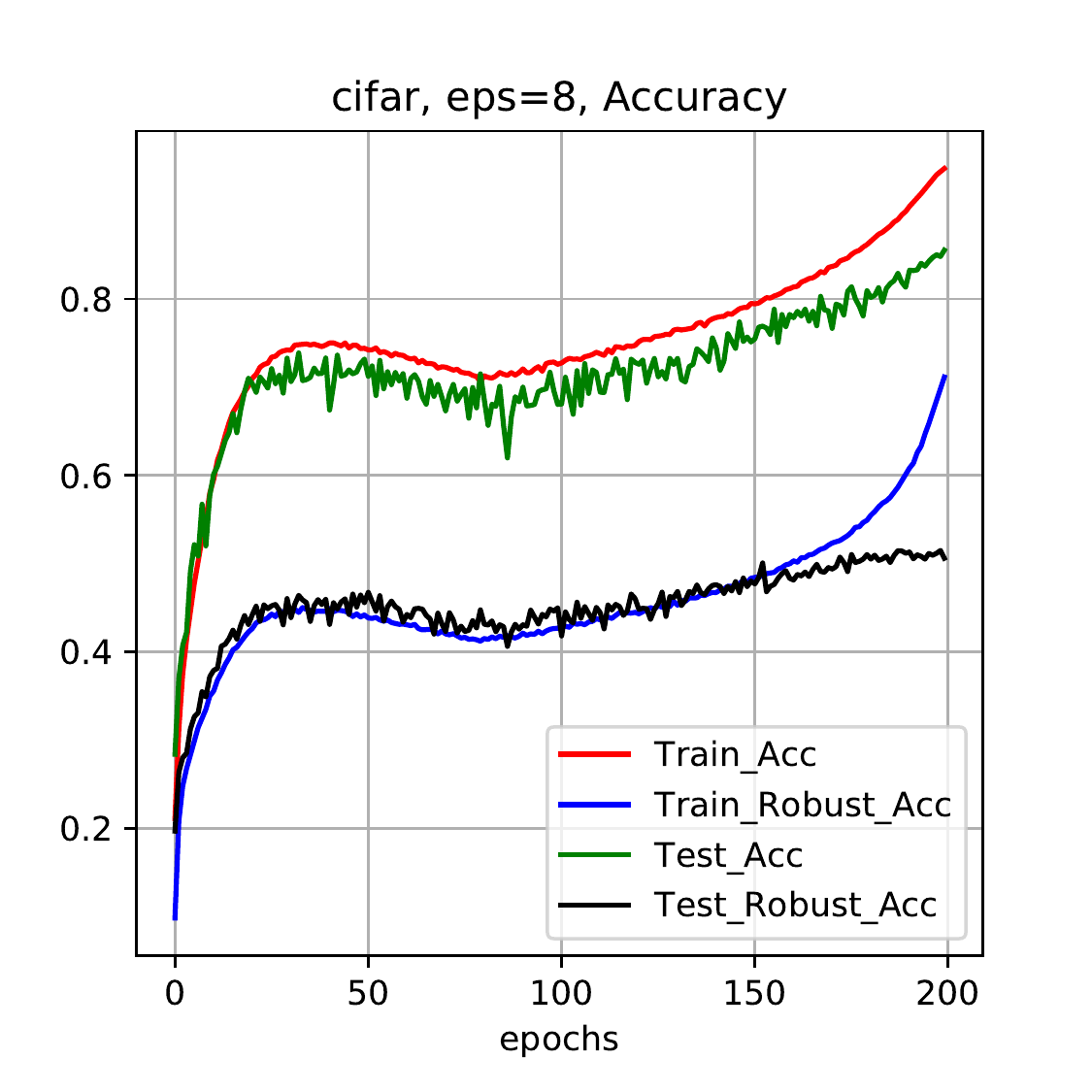}
\end{minipage}
}
\subfigure[]{
\begin{minipage}[htp]{0.2\linewidth}
\centering
\includegraphics[width=1.3in]{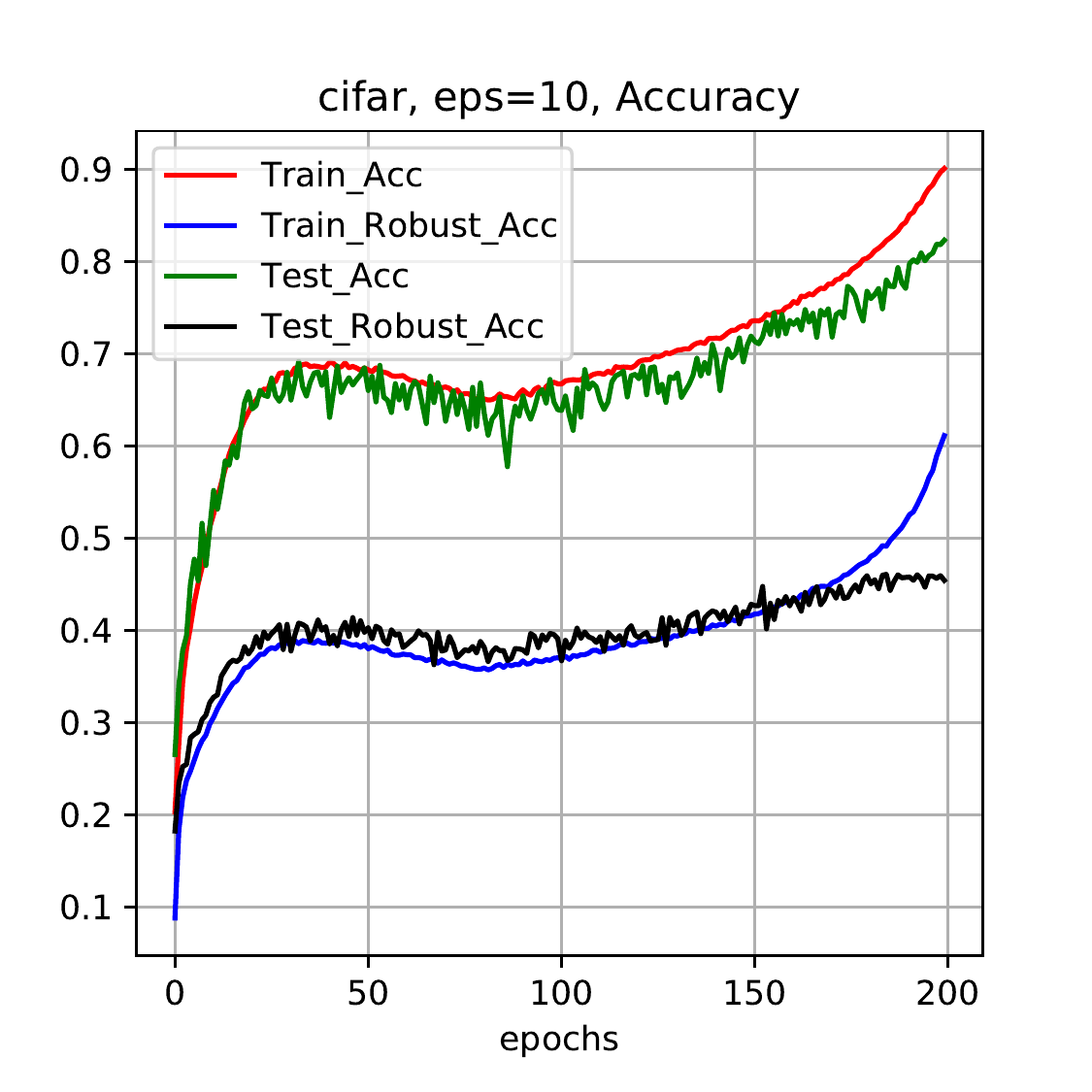}
\end{minipage}
}}

\hspace{-0.4in}\scalebox{0.9}{
\subfigure[]{
\begin{minipage}[htp]{0.2\linewidth}
\centering
\includegraphics[width=1.3in]{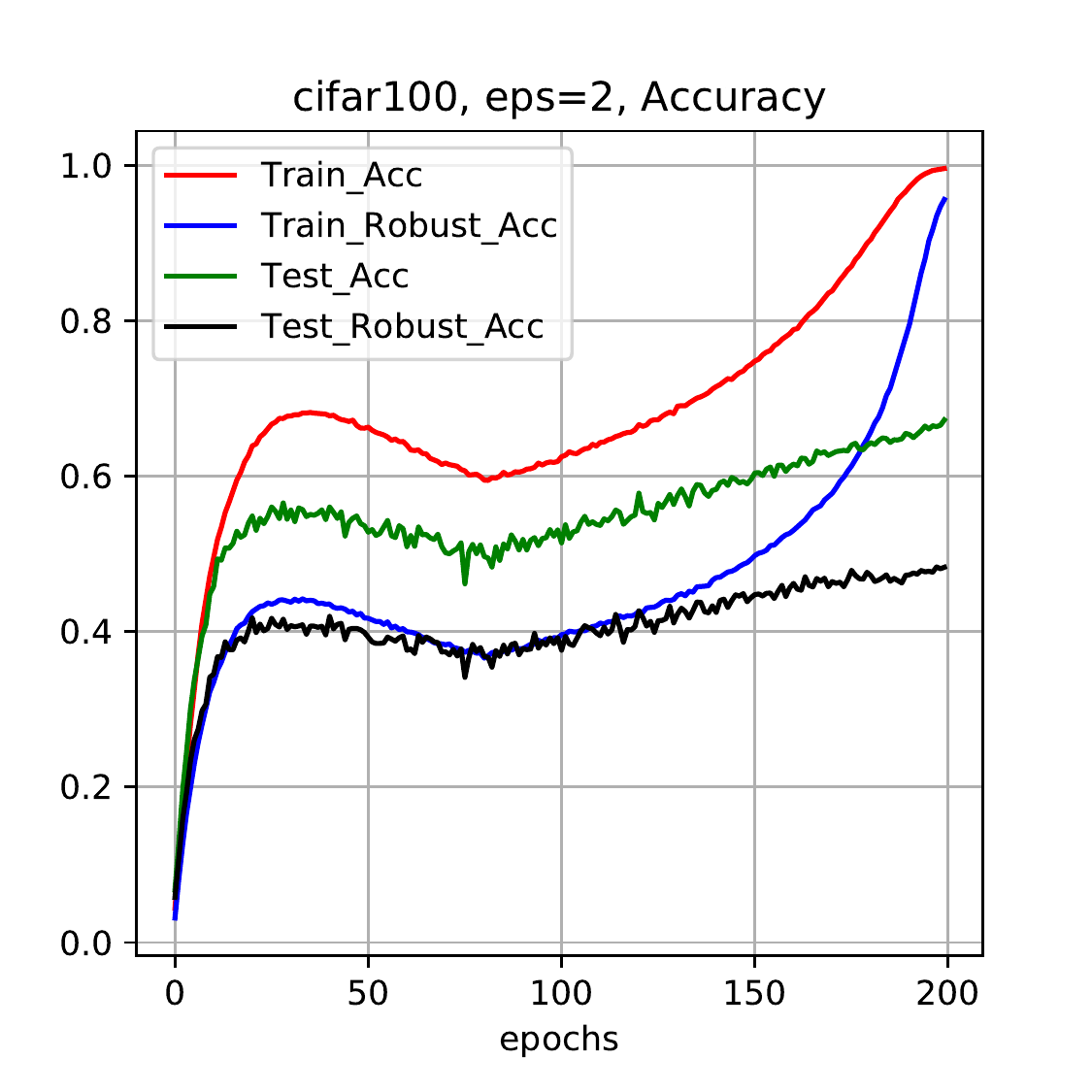}
\end{minipage}%
}
\subfigure[]{
\begin{minipage}[htp]{0.2\linewidth}
\centering
\includegraphics[width=1.3in]{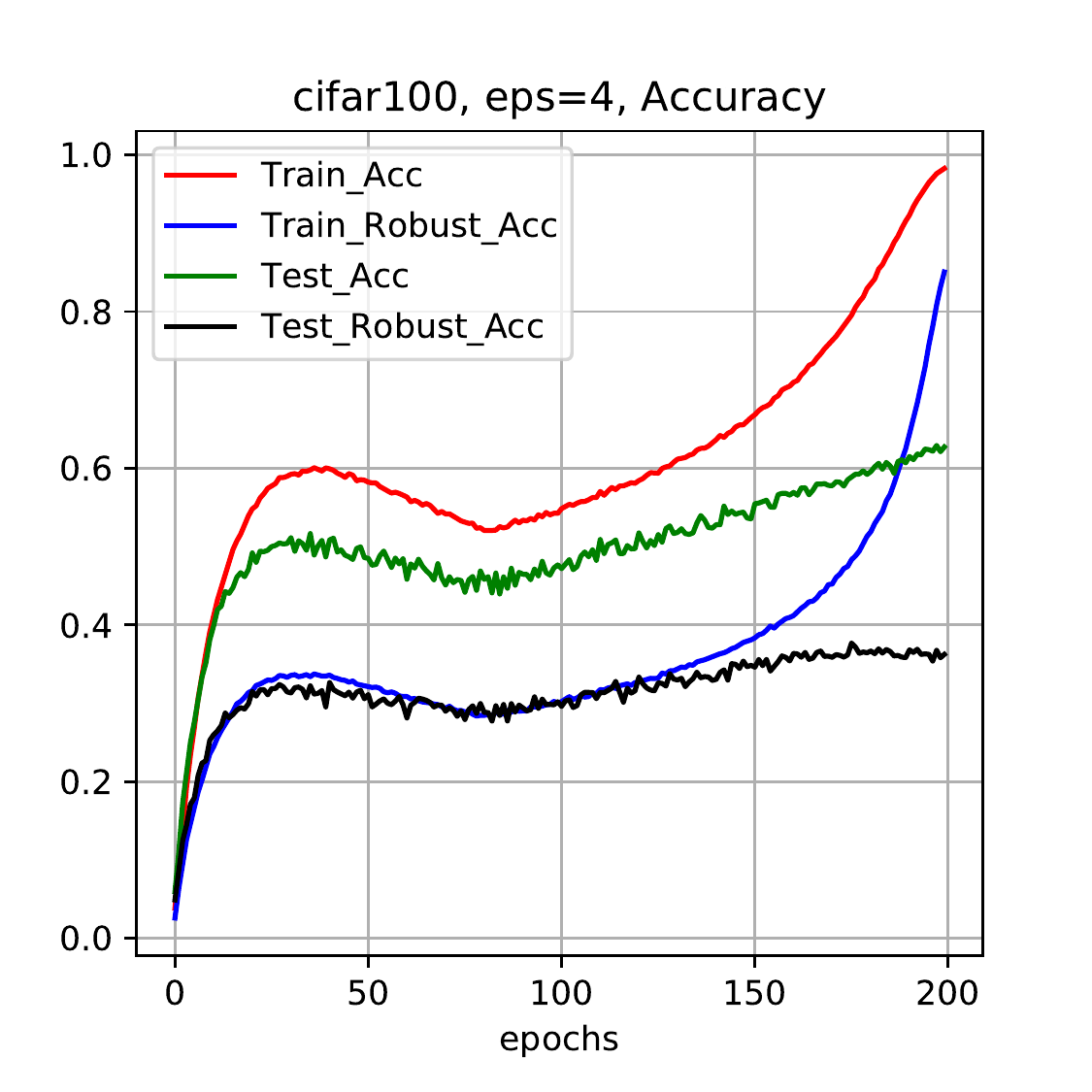}
\end{minipage}
}
\subfigure[]{
\begin{minipage}[htp]{0.2\linewidth}
\centering
\includegraphics[width=1.3in]{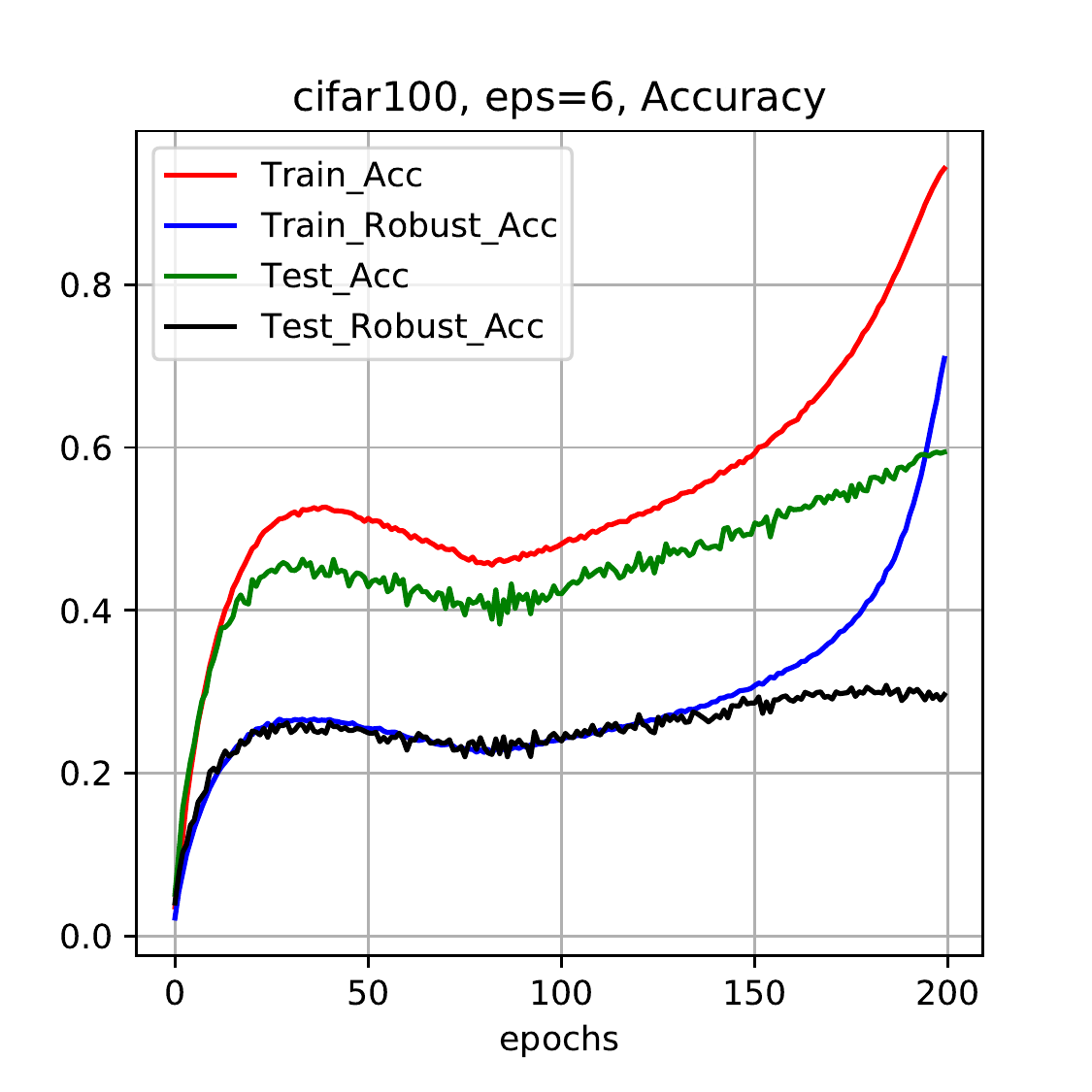}
\end{minipage}
}
\subfigure[]{
\begin{minipage}[htp]{0.2\linewidth}
\centering
\includegraphics[width=1.3in]{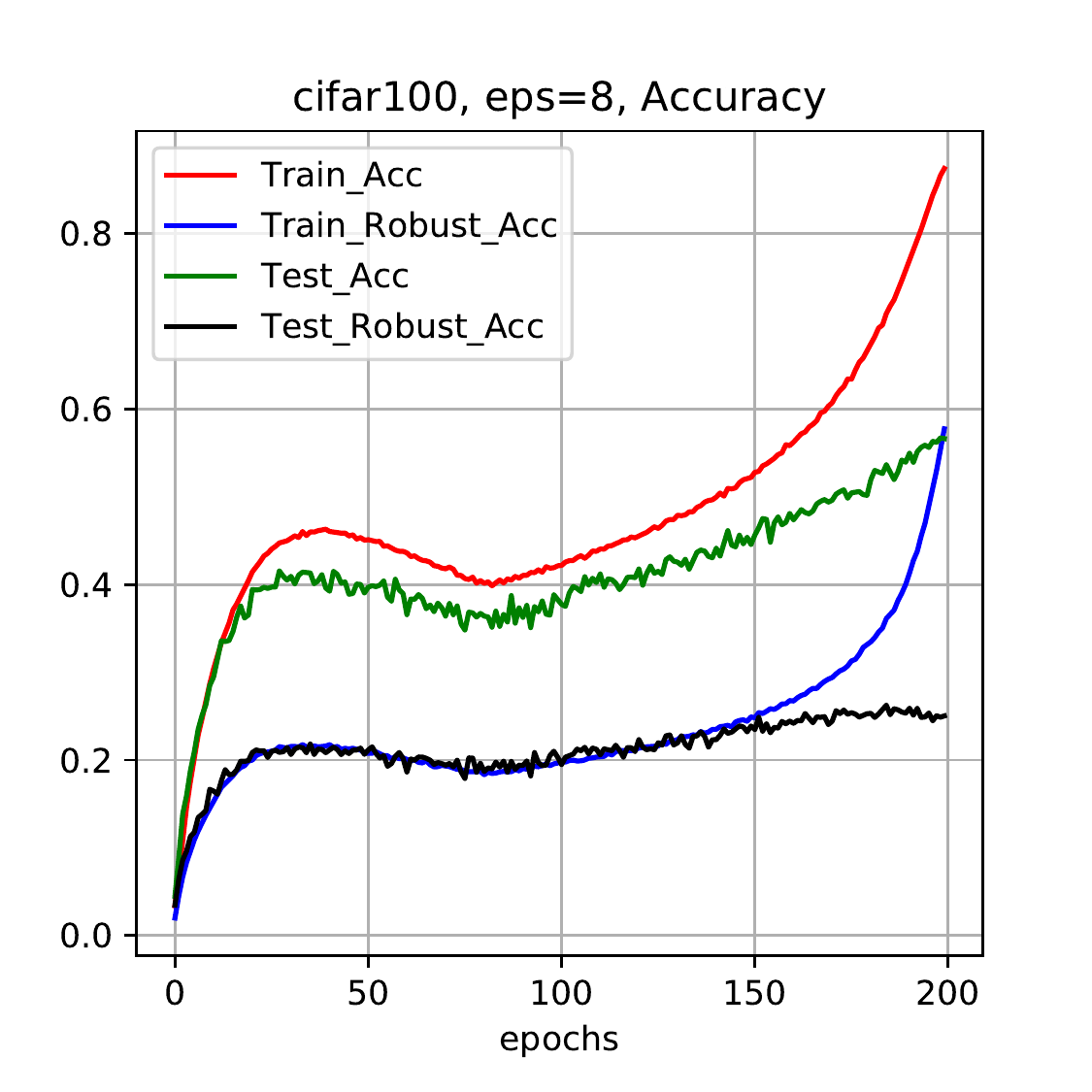}
\end{minipage}
}
\subfigure[]{
\begin{minipage}[htp]{0.2\linewidth}
\centering
\includegraphics[width=1.3in]{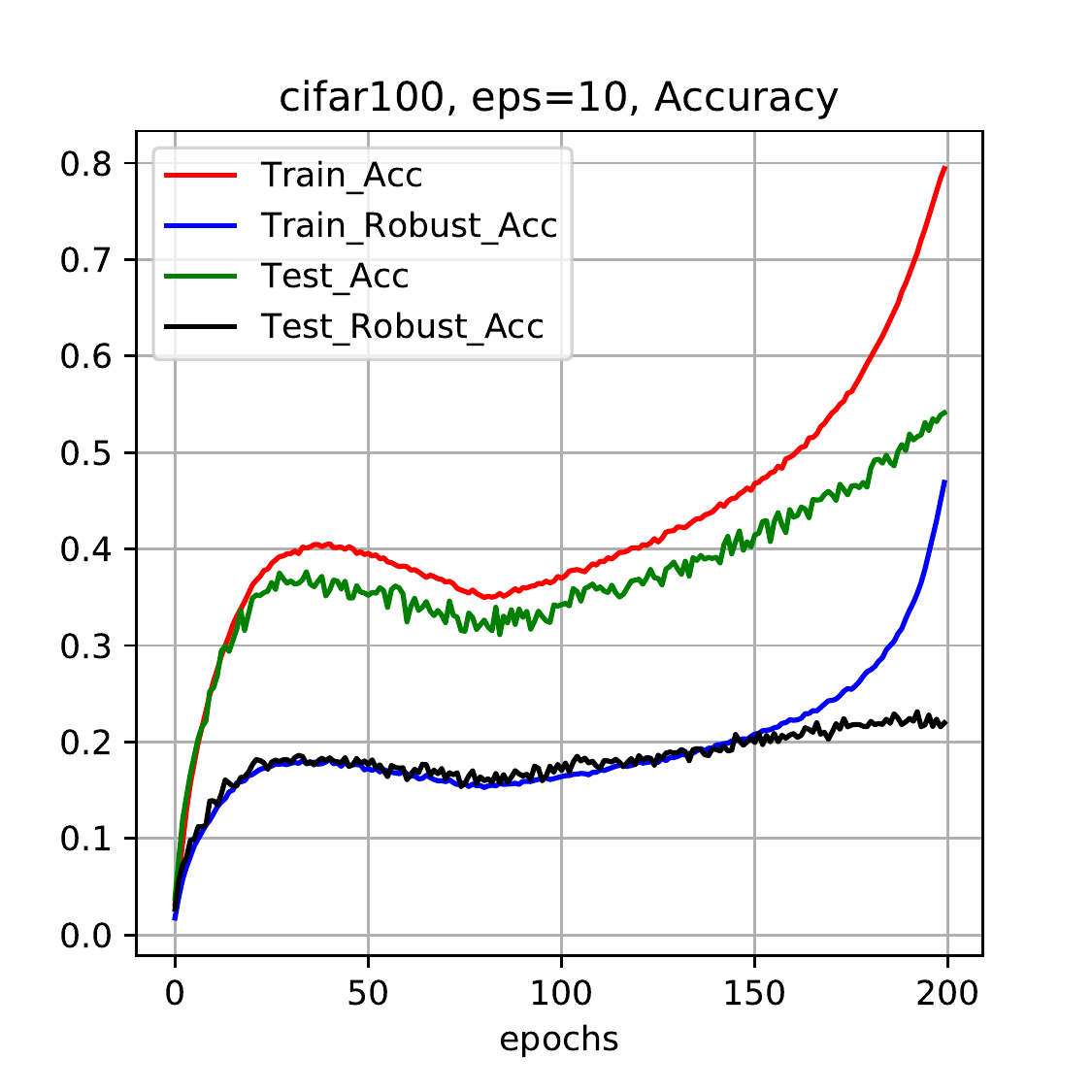}
\end{minipage}
}}
\centering
\caption{Accuracy of adversarial training with super-converge learning rate. The first row is the experiments on SVHN. The second row is the experiments on CIFAR-10. The last row is the experiments on CIFAR-100. The first column to the last column are the experiments of $\epsilon$ equal to 2, 4, 6, 8, and 10, respectively.}
\label{fig:add2}
\vskip -0.1in
\end{figure*}

In this section, we provide additional experiments on SVHN, CIFAR-10, and CIFAR-100. In Fig. \ref{fig:fixed}, we show the experiments of adversarial training using a fixed learning rate. In Fig. \ref{fig:piece}, we show the experiments of adversarial training using a standard piece-wise linear learning rate. In Fig. \ref{fig:add2}, we show the experiments of adversarial training using cyclic learning rate. 

\paragraph{Cyclic Learning Rate.} We illustrate the experiments of adversarial training using cyclic learning rate. The learning rate increases linearly in the first 80 epochs and decreases to zero in the last 120 epochs. This learning rate mainly contributes to optimization. In terms of generalization, as discussed in theoretical settings, the generalization bound is no larger than that of the previous cases.
\label{a3}

\end{document}